\documentclass{article}
\usepackage{arxiv}
\usepackage{graphicx}
\usepackage{tikz}
\usetikzlibrary{arrows,shapes}
\usetikzlibrary{positioning}
\usepackage{amsmath,amssymb,amsfonts}
\usepackage{bm}
\usepackage{bbm}
\usepackage{multirow} 
\usepackage[linesnumbered, ruled]{algorithm2e}
\usepackage{mathrsfs}
\usepackage{tabulary}
\usepackage{stmaryrd}
\usepackage{varwidth}
\usepackage[blocks]{authblk}
\usepackage{longtable}
\usepackage[toc,page]{appendix}
\makeatletter
\newcolumntype{K}[1]{>{\centering\arraybackslash}p{#1}}
\begin{document}

\title{Bayesian Optimization using Deep Gaussian Processes%\thanks{Grants or other notes
}

\author{\textbf{Ali Hebbal} \footnote{Ph.D. student, ONERA, DTIS, Universit\'e Paris Saclay, Universit\'e de Lille, CNRS/CRIStAL, Inria Lille, ali.hebbal@onera.fr}}

\affil{ONERA, DTIS, Université Paris Saclay, Universit\'e de Lille, CNRS/CRIStAL, Inria Lille}
\author{\textbf{Loic Brevault} \footnote{Research Engineer, ONERA, DTIS, Universit\'e Paris Saclay, loic.brevault@onera.fr} \hspace{1pt}  and \textbf{Mathieu Balesdent} \footnote{Research Engineer, ONERA, DTIS, Universit\'e Paris Saclay, mathieu.balesdent@onera.fr}}
\affil{ONERA, DTIS, Universit\'e Paris Saclay, F-91123 Palaiseau Cedex, France}
\author{\textbf{El-Ghazali Talbi} \footnote{Professor at Polytech'Lille - Universit\'e de Lille, el-ghazali.talbi@univ-lille1.fr}  \hspace{1pt} and \textbf{Nouredine Melab} \footnote{Professor at Polytech'Lille - Universit\'e de Lille, nouredine.melab@univ-lille1.fr} }
\affil{ Universit\'e de Lille, CNRS/CRIStAL, Inria Lille, Villeneuve d'Ascq, France }

\maketitle

\begin{abstract}
Bayesian Optimization using Gaussian Processes is a popular approach to deal with the optimization of expensive black-box functions. However, because of the \textit{a priori} on the stationarity of the covariance matrix of classic Gaussian Processes, this method may not be adapted for non-stationary functions involved in the optimization problem. To overcome this issue, a new Bayesian Optimization approach is proposed. It is based on Deep Gaussian Processes as surrogate models instead of classic Gaussian Processes. This modeling technique increases the power of representation to capture the non-stationarity by simply considering a functional composition of stationary Gaussian Processes, providing a multiple layer structure. This paper proposes a new algorithm for Global Optimization by coupling Deep Gaussian Processes and Bayesian Optimization. The specificities of this optimization method are discussed and highlighted with academic test cases. The performance of the proposed algorithm is assessed on analytical test cases and an aerospace design optimization problem and compared to the state-of-the-art stationary and non-stationary Bayesian Optimization approaches.
\end{abstract}
\clearpage
\noindent\textbf{Notations:} \\
\begin{itemize}
\item A scalar is represented by a lower case character: $y\in \mathbb{R}$\\
\item A vector is represented by a bold character:  $\textbf{x}\in \mathbb{R}^d, \textbf{x}=[x_1,\ldots,x_d]^\top$\\
\item A matrix is represented by upper case character: $X=\begin{bmatrix} x_{1,1} &\cdots&x_{1,j} &\cdots & x_{1,d} \\\vdots &\ddots &\vdots &\ddots& \vdots \\x_{i,1} &\cdots&x_{i,j} &\cdots & x_{i,d}\\\vdots &\ddots &\vdots &\ddots& \vdots\\x_{n,1} &\cdots&x_{n,j} &\cdots & x_{n,d} \end{bmatrix} \in \mathcal{M}_{n,d} $ \\
\item The $i^{th}$ row of a matrix X is noted $\textbf{x}^{(i)\top}$ \\
\item The $j^{th}$ column of a matrix X is noted $\textbf{x}_j$ \\
\end{itemize}

%$n$ & number of objectives\\
%$n_c$ & number of constraints \\
%$D$ & Dimension of the input space \\
%$L$ & Number of layers in a Deep Gaussian Process \\
%$\prec$  & Pareto dominance relation \\
%$\textbf{x}^{(i)}$  & $i$-th element of the DoE \\
%$x_i$  & $i$-th component of vector \textbf{x}\\
%$BO$ & Bayesian Optimization \\
%$MO$ & Multi-Objective\\
%$EHVI$   & Expected HyperVolume Improvement\\
%$GP$ & Gaussian Process \\
%$DGP$& Deep Gaussian Process \\
%$\textbf{h}_l$ & $l^{th}$ hidden unit \\
%$\mathcal{Z},\textbf{u}$ & Input-output induced variables\\
%$M$ & Number of induced inputs\\
%$p(\cdot)$ & Distribution of a variable\\
%$q(\cdot)$ & Approximated variational distribution \\

%\end{longtable}}

\section{Introduction}
\label{intro}

The design of complex systems often involves computationally intensive simulation codes that provide black-box functions as objectives and/or constraints. For example, for multidisciplinary design optimization problems, disciplinary codes are often modeled as black-box functions and the optimization process requires an iterative loop between these disciplines, inducing a computational burden. Within the context of black-box functions relying on legacy codes that do not provide analytical forms of the functions or the gradients, the use of exact optimization approaches is often not tractable. Furthermore, the high computational cost makes the use of algorithms that needs a consequent number of evaluations (gradient approximation, evolutionary algorithms, \textit{etc.}) not suitable. Moreover, the objective and constraints functions involved often have non-linear landscape with multiple local optima, hence, complicating more the optimization problem. 

One popular way to deal with expensive black-box function optimization is Bayesian Optimization (BO). BO algorithms are iterative sampling procedures based on surrogate models. To avoid running excessively the expensive functions, surrogate models allow the emulation of the statistical relationship between the design variables and the responses (objective function and constraints) given a dataset also called Design of Experiments (DoE). Different surrogate models can be used in design optimization \cite{wang2007review}. One of the most popular Bayesian Optimization algorithms is ``Efficient Global Optimization'' (EGO) developed by Jones \textit{et al.} \cite{jones1998efficient}. It is based on Gaussian Process (GP) regression \cite{rasmussen2006gaussian} (also called Kriging). The main advantage of GP is that in addition to the prediction, it provides an uncertainty estimation of the surrogate model response. Based on these two outputs, infill criteria are constructed to iteratively add the most promising candidates to the dataset. These points are then evaluated on the exact functions and the surrogate models are updated and so on, until a stopping criterion is satisfied.

Classical GP regression is based on stationary covariance functions $k(\cdot,\cdot)$, \textit{i.e.} the covariance function is invariant by translation: $ \forall \textbf{x},\textbf{x}',\textbf{h} \in \mathbb{R}^d, k(\textbf{x}+\textbf{h},\textbf{x'}+\textbf{h})=k(\textbf{x},\textbf{x'})$ \cite{papoulis1991stochastic}. Hence, the covariance depends only on the shift $ \textbf{x}-\textbf{x'}$. This induces a uniform level of smoothness in the prediction over the design space. This stationary prior covariance is effective to deal with stationary functions, however, it induces a major issue when the functions to be approximated are not stationary \textit{i.e.} the level of variability of the response changes dramatically from one region of the input space to another. 

The question of non-stationarity is discussed in different fields of research. In climate science due to dramatic changes in precipitation, the stationarity assumption is dropped for modeling climate phenomena \cite{milly2008stationarity} \cite{cordery1993non} \cite{garg2012learning}. In signal processing and finance among other fields, non-stationary models are often used to fit time series over a long period of time \cite{konda2006fitting}. Also in geostatistics non-stationarity occurs when dealing with a region with different landscapes and topographic features \cite{atkinson2007non}.  

In design optimization, due to the abrupt change of a physical property, the objective functions or the constraints may vary with different degrees of smoothness from one region of the design space to another. Specifically, aerospace design engineering involves different disciplines which can induce non-stationary processes. For example in aerodynamics, Computational Fluid Dynamics (CFD) problems often have different specific flow regimes due to separation zones, circulating flows, vortex bursts, transitions from subsonic to transonic, supersonic and hypersonic flow regimes. In the propulsion discipline, the combustion involves irreversible thermodynamics transformations, that are characterized by sudden and rapid changes (\textit{e.g.}, sudden state change of the matter, spontaneous chemical reactions, spontaneous mixing of matter of different states). There can also be non-stationarities in the structure discipline, for example in the stress\text{-}strain curve of a material, the elastic region, the strain hardening region and the necking region present different behaviors. 

To overcome this issue, different approaches have been developed to deal with non-stationarity in regression such as regression trees \cite{breiman2017classification}, neural networks \cite{specht1991general}, wavelet process modeling \cite{vidakovic2009statistical}. GP regression has also been adapted to non-stationary cases. Indeed, one can summarize the different approaches for non stationary GP regression in three main categories: direct formulation of a non-stationary covariance function \cite{higdon1999non} \cite{paciorek2006spatial}, local stationary covariance functions \cite{rasmussen2002infinite} \cite{haas1990kriging} and input-space warping approaches \cite{sampson1992nonparametric} \cite{xiong2007non}. However, these approaches may have some limitations when dealing with BO problems where data is scarse, or in high dimensional problems (Section~\ref{sec:1}).

Another approach which is not among these classic methods to handle non-stationarity consists in using Deep Gaussian Processes (DGPs) \cite{damianou2013deep}. DGPs correspond to a functionnal composition of GPs, which may allow the description of more complex functions than standard GPs. The key contribution of this paper is to define a BO \& DGP algorithm making the coupling of DGPs and BO possible to optimize problems involving non-stationary functions. The performance of this algorithm is assessed on different analytical test functions and on an aerospace optimization design problem.

%The coupling of DGPs and BO has been briefly introduced previously \cite{hebbal2018efficient}, with preliminary promising results on 1D and 2D analytical problems. In this paper, a deeper investigation on the combination of DGPs and BO is presented in order to define a BO \& DGP algorithm.  

The following paper is organized in three main sections. Section~\ref{sec:1} provides a review of literature on the different concepts used along the paper. BO with a focus on GP and infill criteria is described. Then, the different non-stationary approaches for GP are presented. Finally, the DGP modeling approach is introduced, with a review on some inference approaches used for its training. In Section~\ref{sec:2}, the coupling of BO and DGPs is discussed. This discussion covers several aspects, such as the training approach of DGP in the context of BO, uncertainty quantification, architecture of the DGP and infill criteria, in order to define the BO \& DGP algorithm, Deep Efficient Global Optimization (DEGO). Section~\ref{sec:3}, presents experimentations on analytical optimization test problems and on an aerospace optimization test problem, to assess the performance of DEGO compared to traditional existing approaches.
\section{State of the art}
\label{sec:1}
\subsection{Bayesian Optimization}
\label{ssec:11}
%
%\subsubsection{General framework}
%\label{sssec:111}
The context of expensive black-box optimization problems is considered throughout this paper meaning that the objective function $f^{exact}:\mathbb{X} \subseteq \mathbb{R}^d \rightarrow \mathbb{R}$ and the $n_c$ constraint functions $g_i^{exact}:\mathbb{X} \subseteq \mathbb{R}^{d} \rightarrow \mathbb{R}, 1\leq i \leq n_c$ are computationally expensive and black-box functions. Without loss of generality the following minimization problem ($\mathscr{P}$) is considered:
\begin{equation}
(\mathscr{P})\left|\begin{array}{rcl} \underset{\textbf{x}}{\text{Minimize}} &y&=f^{exact}(\textbf{x})\\ \mbox{subject to } & g^{exact}_i(\textbf{x}) & \leq 0, \mbox{     } \forall i \in \{ 1,\ldots,n_c \} \end{array}\right.
\end{equation}

The expensive and black-box aspects of the considered functions limit their evaluations to a sparse data-set (Design of Experiments):\\
$(DoE)\left\{\begin{array}{rcl} X &=&\left[\textbf{x}^{(1)},\ldots,\textbf{x}^{(n)}\right]^\top, \textbf{x}^{(i)} \in \mathbb{R}^d,  \forall i \in \{ 1,\ldots,n \} \\ \textbf{y}&=&\left[y^{(1)}=f^{exact}\left(\textbf{x}^{(1)}\right),\ldots,y^{(n)}=f^{exact}\left(\textbf{x}^{(n)}\right)\right]^\top\\ \textbf{c}_i&=&\left[c^{(1)}_i=g^{exact}_i\left(\textbf{x}^{(1)}\right),\ldots,c^{(n)}_i=g^{exact}_i\left(\textbf{x}^{(n)}\right)\right]^\top, \forall i \in \{ 1,\ldots,n_c \} \end{array} \right.$\\
where $n$ is the size of the dataset.

BO algorithms are sequential design algorithms. The design space is filled sequentially with new candidates with the objective to improve the current minimum in the DoE: 
\begin{equation}
y_{\min}=\min\left\{f^{exact}(\textbf{x}^{(i)})  | i \in \{ 1,\ldots,n \} \text{ and }  \forall j \in \{ 1,\ldots,n_c \}, g^{exact}_j(\textbf{x}^{(i)})\leq 0\right\}
\label{ymin}
\end{equation} 

This sequential aspect of the BO algorithms consists of two iterative operations. The first one is the modeling of the expensive black-box functions ($f^{exact},g^{exact}_1,\ldots,g^{exact}_{n_c}$) involved in the optimization problem using the DoE $X$ and the corresponding exact evaluations $\textbf{y},\textbf{c}_1,\ldots,\textbf{c}_{n_c}$ together with a surrogate modeling approach (Random Forest, Polynomial models, Gaussian Process, Neural Networks, \textit{etc}.) to obtain approximations ($\hat{f},\hat{g}_1,\ldots,\hat{g}_{n_c}$). These latter are cheaper to evaluate, which allows a greater number of evaluations than the exact functions. 

The second procedure consists of the determination of the most promising candidate to add to the current DoE in order to improve the current minimum $y_{min}$ using the information given by the surrogate models. This is achieved by optimizing an acquisition function (also called infill sampling criterion) on the design space, which is a performance measure of a candidate's potential from a minimization point of view. Once the most promising point is added to the dataset, it is evaluated on the exact expensive functions and the surrogate models are updated, and so on until a stopping criterion is reached (Fig.~\ref{EGO_framework}). Hence the two key aspects in BO algorithms are the surrogate model and the infill sampling criterion. 

One of the most popular BO algorithms is the Efficient Global Optimization (EGO) \cite{jones1998efficient}. It uses GP as surrogate model and the Expected Improvement as the infill sampling criterion. This work is focused on BO algorithms using GP and its variants as surrogate models.

\tikzstyle{format} = [fill=blue!50!cyan!70,draw,align=center,font=\small,text=white,draw, thick,rounded corners=3pt,text width=3cm,minimum size=1.5cm]
\tikzstyle{medium} = [diamond, draw, thin, fill=blue!50!cyan!70,minimum size=1cm,text=white,text width=1cm]
%\line/
%,below right =0.5cm and 0.8cm of h21

\begin{figure}[h]
\begin{center}
\begin{tikzpicture}[node distance=0.7cm, auto,>=latex', thick, bend angle=89]
%\path[use as bounding box] (-5,-5) rectangle (5,5);
\path[->] node[format] (doe) {\textbf{Initial DoE}};

\path[->] node[format, below = of doe] (eva) {\textbf{Evaluation of the dataset on the objective function and constraints}}
                  (doe) edge node {\large $X$} (eva);

\path[->] node[medium, below =of eva] (stop) {\begin{center} \textbf{Stopping Criterion} \end{center}}
				(eva) edge node {\large $\textbf{y}, \textbf{c}_i$} (stop);
    
                  %(eva.south) edge  [bend right]  node {\TeX} (krig.west);
%\node[y width=5cm] (leyend) at (5,-2.5)
%\draw[->] (stop.south) -- node {south of stop} (5,5);
%									%\draw[->] (eva.south) |- (krig.east);			
\path[->] node[format,right=of stop] (krig) {\textbf{Training the GP models for the objective function and constraints}}
					(stop) edge node {\normalsize No} (krig);

\path[->] node[format, right =of krig] (infill) {\textbf{Optimization of the infill criterion}}
							(krig) edge node {\large $f,g$} (infill);

\path[->] node[format, above = 1.455cm of infill] (add) {\textbf{Addition of the most promising point to the dataset}}
							(infill) edge node {\large $\textbf{x}_{new}$} (add);
	
\draw[->] (add.west) -- node {\large $X$} (eva.east);

\path[->] node[below =of stop] (yes) {\large \textbf{End}}
			(stop) edge node {\normalsize Yes} (yes);;

\end{tikzpicture}
\caption{Bayesian Optimization with Gaussian Process framework}
\label{EGO_framework}
\end{center}
\end{figure}
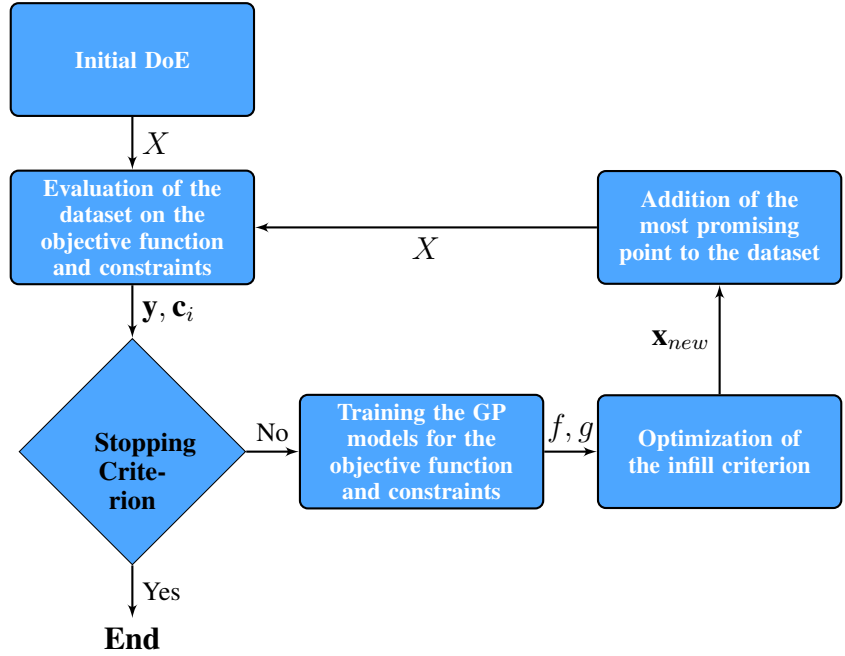

\subsubsection{Gaussian Process}
\label{sssec:112}
%A surrogate model is  built from a set of points called training set or design of experiment (DoE) $\mathcal{X}=\{\textbf{x}^{(1)},...,\textbf{x}^{(N)}\}$ and their associated response values $\mathcal{Y}=\{y^{(1)},...,y^{(N)}\}$ where $N$ is the number of observations and $\textbf{x}^{(i)} \in \mathbb{R}^d$. Then, it is possible to use this model to predict the output at a new point. The interest of a surrogate model is its cheap computational evaluation cost, thus instead of evaluating the expensive black-box function for a large number of points, it is possible to build a surrogate-model using fewer evaluations of the exact function, and predict its value in the other points using this model. 
A Gaussian Process \cite{rasmussen2006gaussian} $f$ is a stochastic process indexed by a set $\mathbb{X} \subseteq \mathbb{R}^d$: $\{f(\textbf{x}) :\textbf{x} \in \mathbb{X} \}$ such as any finite number of random variables of the process has a joint Gaussian distribution:
\begin{equation}
\forall n'\in \mathbb{N}^*, \forall X' =\left[\textbf{x}'^{(1)},\ldots,\textbf{x}'^{(n')} \right]^\top, f(X') \sim \mathcal{N} \left(\mu \left(X'\right),k\left({X',X'}\right) \right)
\label{GPprop}
\end{equation}
with $f(X') = \left [f\left(\textbf{x}'^{(1)}\right),\ldots,f\left(\textbf{x}'^{(n')}\right)  \right]^\top$. A GP, is completely defined by its mean and covariance functions and is noted $f(\cdot) \sim \mathcal{GP} \left(\mu(\cdot),k(\cdot,\cdot)\right)$, with $\mu(\cdot)$ the mean function and $k(\cdot,\cdot)$ the covariance function.

\begin{figure}
\begin{minipage}[c]{0.5\linewidth}
\center
\includegraphics[width=0.8\linewidth]{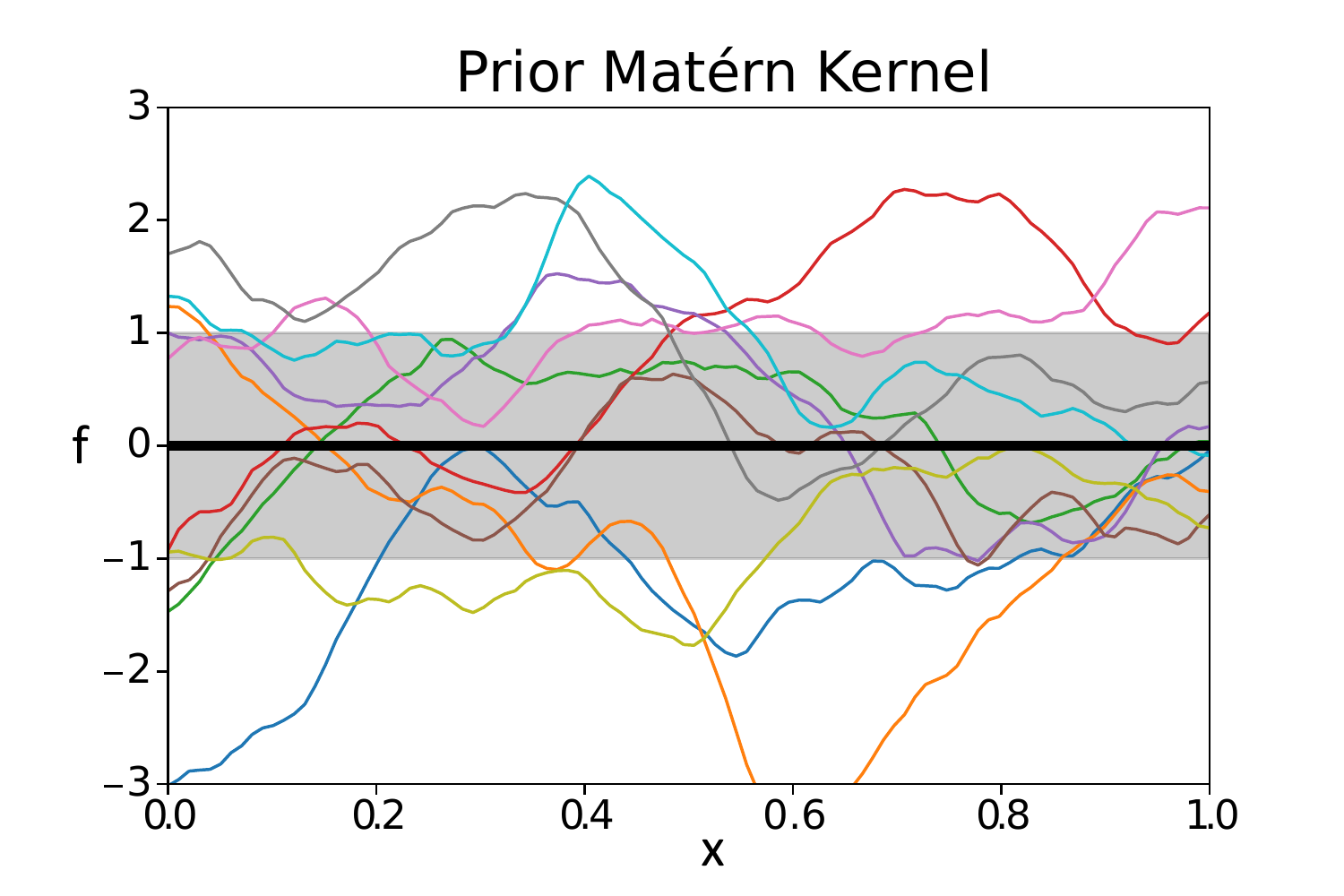}\\
%\tiny $   \left(1+\sqrt{3}\sum\limits_{i=1}^d \textcolor{red}{\theta_i}.|x_i-x'_i| \right ) \exp \left(-\sqrt{3}\sum\limits_{i=1}^d \textcolor{red}{\theta_i}.|x_i-x'_i| \right )$
\end{minipage}
\hfill
\begin{minipage}[c]{0.5\linewidth}
\center
\includegraphics[width=0.8\linewidth]{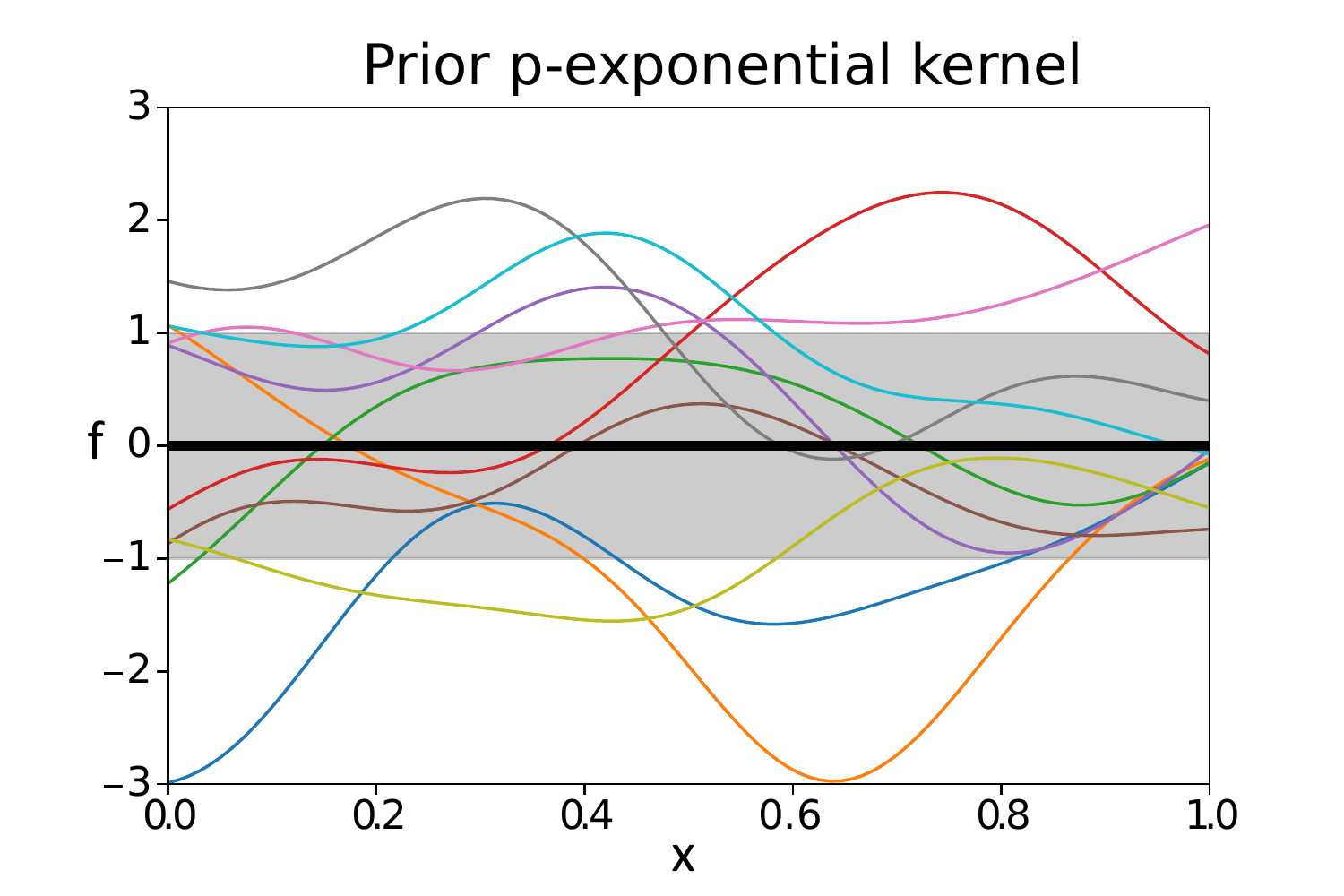}\\
% \tiny $  \exp \left(-\sum\limits_{i=1}^d \textcolor{red}{\theta_i}.|x_i-x'_i|^{\textcolor{red}{p_i}}\right)$
\end{minipage}

\caption{One-dimensional samples from the prior of a $3/2$ Mat\'ern Kernel (left) and a p-exponential kernel (right)}
\label{covariance_function}
\end{figure}

These properties are used to build GP regression models. Given a DoE $X$ and its associated response values $\textbf{y}$, this surrogate model may be used to predict in new locations $X^*$ the output values $\textbf{y}^*$. To do so, a GP prior is considered $f(\cdot) \sim \mathcal{GP} \left(\mu(\cdot),k^{\boldsymbol{\Theta}} (\cdot,\cdot)\right)$. The prior mean function $\mu(\cdot)$ can take a form that describes the trend of the approximated function if it is known (universal Kriging) otherwise a constant mean function $\mu$ may be considered (ordinary Kriging). The prior covariance function $k^{\boldsymbol{\Theta}}$ represents the prior belief of the unknown function to be modeled (\textit{e.g.} smoothness, periodicity, stationarity, separability) (see Figure~\ref{covariance_function}). The prior covariance function depends on a number of hyper-parameters $\boldsymbol{\Theta}$ allowing a better fit on the data. 
When dealing with noisy observations, $\textbf{y}$ are not the exact values of the unknown function in $X$ but a noisy version. One way to deal with this case is to introduce an identically distributed Gaussian noise with variance $\sigma^2$ such as the relationship between the latent (non-observed) function values $\textbf{f}=f(X)$ and the observed response $\textbf{y}$ is given by: $p(\textbf{y}|\textbf{f})= \mathcal{N}(\textbf{f},\sigma^2 I_n)$ where $I_n$ is the identity matrix of size $n$. Hence, the prior covariance on the noisy observations becomes: 
$k_{noisy}^{\boldsymbol{\Theta}}(X,X)=k^{\boldsymbol{\Theta}}(X,X) + \sigma^2 I_n$.

Training the GP consists in finding the optimal values of the hyper-parameters $\boldsymbol{\Theta}$, $\sigma$ and $\mu$ (for ordinary kriging). These values are obtained by a standard maximum likelihood procedure:

%$\{ \text{Max } : p(\textbf{y}|\mathcal{X}) = \mathcal{N}(\textbf{y}|\pmb{\mu}, \textbf{K}^\Theta_{NN}+\sigma^2 I_N)
% \\ \text{w.r.t} : \boldsymbol{\Theta}, \sigma, \mu  $
 \begin{equation}
\left\{\begin{array}{rcl} \text{Maximize } &:& p(\textbf{y}|X) = \mathcal{N}(\textbf{1}\mu , k^{\boldsymbol{\Theta}}(X,X)+\sigma^2 I_n)
 \\ \text{According to} &:&  \boldsymbol{\Theta}, \sigma, \mu  \end{array}\right.
 \end{equation}

Once the hyper-parameters optimized $\hat{\boldsymbol{\Theta}}, \hat{\sigma}, \hat{\mu} $, the prediction in new locations $X^*= \left[{\textbf{x}^*}^{(1)},\ldots,{\textbf{x}^*}^{(n^*)}\right]^\top$ is done in two steps. Firstly, using the property of a GP, Eq.(\ref{GPprop}), the joint distribution of the predicted outputs $\textbf{f}^*=f(X^*)$ and the observed outputs $\textbf{y}$ is given by:
\begin{equation}
\left(\begin{array}{c}\textbf{y}\\ \textbf{f}^*\end{array}\right) \sim \mathcal{N} \left(\textbf{1}\hat{\mu},\begin{array}{cc}k^{\hat{\boldsymbol{\Theta}}}(X,X) + \hat{\sigma}^2 I_n &,k^{\hat{\boldsymbol{\Theta}}}(X,X^*)\\k^{\hat{\boldsymbol{\Theta}}}(X^*,X)  &,k^{\hat{\boldsymbol{\Theta}}}(X^*,X^*)\end{array}\right)
\end{equation}
Then, by using the conditional distribution of a joint Gaussian distribution, which is equivalent here to conditioning the prior distribution on the observations \textbf{y}, the posterior distribution is obtained:
\begin{equation}
\textbf{f}^*|X^*,\textbf{y},X \sim \mathcal{N}\left( \hat{f}(X^*), \hat{\Sigma}(X^*) \right)
\end{equation}
where $\hat{f}(X^*)$ and $\hat{\Sigma}(X^*)$ are respectively the mean and the covariance of the posterior distribution and are defined as:
\begin{equation}
\hat{f}(X^*)=\textbf{1} \hat{\mu}+k^{\hat{\boldsymbol{\Theta}}}(X^*,X){\left(k^{\hat{\boldsymbol{\Theta}}}(X,X)+\hat{\sigma}^2I_n\right)}^{-1}(\textbf{y}-\textbf{1}\hat{{\mu}})
\end{equation}
\begin{equation}
 \hat{\Sigma}(X^*)= k^{\hat{\boldsymbol{\Theta}}}(X^*,X^*)-k^{\hat{\boldsymbol{\Theta}}}(X^*,X){\left(k^{\hat{\boldsymbol{\Theta}}}(X,X)+\hat{\sigma}^2I_n \right)}^{-1}k^{\hat{\boldsymbol{\Theta}}}(X,X^*)
\end{equation}

%The drawback of standard GP is its algorithmic complexity $\mathcal{O}(N^3)$ induced by the inversion of the covariance matrix $K_{\mathcal{X},\mathcal{X}}$. Hence, making standard GP not suitable for large dataset. To overcome this limit of GPs, Sparse Gaussian Process consisting of low rank approximation of the covariance matrix $K_{\mathcal{X},\mathcal{X}}$ using induced variables have been developed \cite{snelson2006sparse} \cite{titsias2009variational}. The idea is to consider a set of $M$ inducing pair of input-output variables $\mathcal{Z}=\left\{\textbf{z}^{(1)},\ldots,\textbf{z}^{(M)}\right\}$ and $\textbf{u}=\mathbbm{f}(\mathcal{Z})=\left\{u^{(N)},\ldots,u^{(M)}\right\}$

\begin{center}

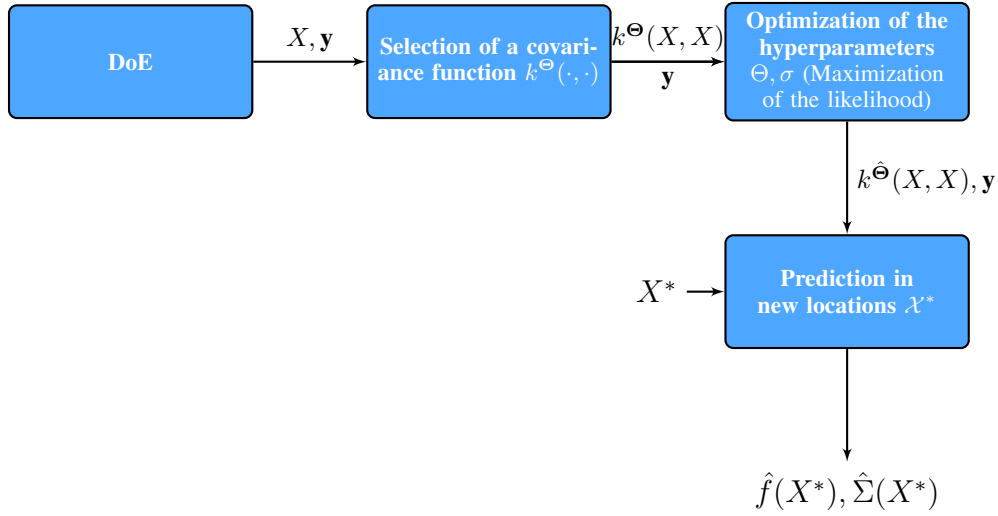
\begin{figure}[h]
\center
\begin{tikzpicture}[node distance=1.5cm, auto,>=latex', thick, bend angle=89]
%\path[use as bounding box] (-5,-5) rectangle (5,5);
\path[->] node[format] (doe) {\textbf{DoE}};

\path[->] node[format, right = of doe] (cov) {\textbf{Selection of a covariance function $k^{\boldsymbol{\Theta}}(\cdot,\cdot)$}}

                  (doe) edge node {\normalsize $X, \textbf{y}$} (cov);

\path[->] node[format, right =of cov] (like) {\textbf{Optimization of the hyperparameters} $\Theta, \sigma$ (Maximization of the likelihood) }
				(cov) edge node[below] {\normalsize $\textbf{y}$} (like)
				(cov) edge node {\normalsize $k^{\boldsymbol{\Theta}}{(X,X)}$} (like);
    		
\path[->] node[format,below=of like] (pred) {\textbf{Prediction in new locations} $\mathcal{X}^*$}
					(like) edge node {\normalsize $k^{\hat{\boldsymbol{\Theta}}}{(X,X)},\textbf{y} $} (pred);
\path[->] node[left =0.5cm of pred] (Xstar) {\large $X^*$}
			(Xstar) edge node {} (pred);
\path[->] node[below =of pred] (fin) {\large $\hat{f}(X^*),\hat{\Sigma}(X^*)$}
			(pred) edge node { } (fin);

\end{tikzpicture}
\caption{Gaussian Process Regression framework}

\label{GPR_framework}

\end{figure}
\end{center}
Notice that in the special case of a single test point $\textbf{x}^*$, the posterior distribution comes back to a univariate Gaussian distribution: $f^*|\textbf{x}^*,\textbf{y},X \sim \mathcal{N}\left( \hat{f}(\textbf{x}^*), \hat{s}^2(\textbf{x}^*) \right)$, where $\hat{f}(\textbf{x}^*)$ is the mean prediction on $\textbf{x}^*$ and  $\hat{s}$ its associated standard deviation. 
Obtaining a Gaussian posterior distribution gives along with the prediction, an uncertainty measure as a Gaussian error which is useful in the construction of infill criteria for Bayesian Optimization.

The steps of GP regression are summarized in (Figure~\ref{GPR_framework}). Henceforth, for  notation simplifications, the dependence of the prior covariance function on $\boldsymbol{\Theta}$ is dropped, and $k(X,Z)$ is written $K_{X,Z}$. Moreover, the prior GP is considered with a zero mean function $\mu=0$.

\subsubsection{Infill Criteria}
\label{sssec:113}

For selecting infill sample candidates, a variety of criteria has been developed \cite{picheny2013benchmark}. Each criterion performs a trade-off between exploration i.e. investigating regions where the variance is large and exploitation i.e. investigating regions where the prediction is minimized. The Probability of Improvement (PI) criterion samples the location where the probability of improving the current minimum $y_{\min}$ (\textit{cf.} Eq.(\ref{ymin})) is maximized.
\begin{equation}
PI(\textbf{x})=\mathbb{P}\left[f(\textbf{x})\leq y_{\min}\right]=\Phi\left(\frac{y_{\min}-\hat{f}(\textbf{x})}{{\hat{s}(\textbf{x})}}\right)
\end{equation} 
where $\Phi(\cdot)$ is the Cumulative Distribution Function (CDF)of the univariate Gaussian probability distribution. The higher values of $PI(\textbf{x})$, the higher chances that $\hat{f}(\textbf{x})$ is better than $y_{\min}$. The drawback of this criterion is that only the probability is taken into account and not how much a point may improve the current minimum value. This will add a large number of points around the current best point. 
%It can be suitable for local refinement, but not for a global exploration.
The Expected Improvement (EI) overcomes this issue by taking into account the improvement induced by a candidate that is defined as: $I(\textbf{x})=\text{max}\{0,y_{\text{min}}-f(\textbf{x})\}$. EI is then computed as the expectation taken with respect to the posterior distribution:
\par
\begin{eqnarray}
EI(\textbf{x})&=&\mathbb{E}\left[I(\textbf{x})\right]\\
&=& \int_{\mathbb{R}} \text{max}\{0,y_{\text{min}}-t\}p(t|\textbf{x},X,\textbf{y}) \text{d}t\\
\label{eq:EI1} 
&=& (y_{\text{min}}-\hat{f}(\textbf{x}))\Phi\left(\frac{y_{\text{min}}-\hat{f}(\textbf{x})}{{\hat{s}(\textbf{x})}}\right)+{\hat{s}(\textbf{x})}\phi\left(\frac{y_{\text{min}}-\hat{f}(\textbf{x})}{{\hat{s}(\textbf{x})}}\right)
\label{eq:EI2} 
\end{eqnarray}%
\normalsize
where $\phi(\cdot)$ denotes the Probability Density function (PDF) of the univariate Gaussian probability distribution. Two important terms constitute the EI formula. The first part is the same as in PI, but multiplied by a factor that scales the EI value on the supposed improvement value. The second part expresses the uncertainty. It tends to be large when the uncertainty on the prediction is high. So, the EI is large for regions of improvement and also for regions of high uncertainty, allowing global refinement properties. The maximization of the EI can be performed using multi-start of gradient-based optimization algorithms, Monte-Carlo simulations or evolutionary algorithms \cite{frazier2018tutorial}.

Thompson sampling has also been adopted as an infill criterion \cite{basu2017analysis}. It consists in drawing a sample from the posterior distribution and choosing the index of the minimum of this sample as an infill candidate. Other methods can also be mentioned as confidence bound criteria \cite{cox1992statistical} or information theory based infill criteria \cite{hernandez2014predictive}. Recently, portfolio methods combining between these different infill criteria have been developed \cite{shahriari2014entropy} \cite{hoffman2011portfolio}. This multitude of methods shows that there is no single infill criterion that performs better over all problem instances.

To handle constraints in BO, different techniques are used \cite{sasena2002flexibility} \cite{parr2012infill}. The direct method \cite{sasena2001use} which consists in the optimization of the unconstrained infill criterion under approximated constraints. The Expected Violation strategy \cite{audet2000surrogate} which considers the optimization of the unconstrained infill criterion under the constraint of an expected violation inferior to a threshold. The Probability of Feasibility approach \cite{schonlau1996global} which optimizes the product of an unconstrained infill criterion with the probability of feasibility of the constraints.

\subsection{Non-stationary approaches}
\label{ssec:12}
Standard GP regression is based on the \textit{a priori} that the variation in the output depends only on the variation in the design space and not in the region considered. This is induced by the use of stationary covariance functions as \textit{a priori} $ \forall \textbf{x},\textbf{x}',\textbf{h} \in \mathbb{R}^d,  k(\textbf{x}+\textbf{h},\textbf{x'}+\textbf{h})=k(\textbf{x},\textbf{x'})=k_*(\textbf{x}-\textbf{x}')$ where $k_*(\cdot)$ is a scalar function defined on $\mathbb{R}^d$. This \textit{a priori} is generally valid for functions where there is no change in the smoothness of the function considered in the design space. However, this is not suitable for functions with drastic variations. For example, the modified Xiong function (\textit{cf.} Eq.(\ref{xiong}) in Appendix~\ref{appendixA}, Figure~\ref{ns1DGP}),
%\begin{equation}
%h(x)=-0.5\left(\sin\left(40(x-0.85)^4\right)\cos\left(2(x-0.95)\right)+0.5(x-0.9)+1\right), \text{  } x \in[0,1]
%\label{xiong}
%\end{equation}
has two regions with different level of variations. It presents one region where the function varies with a high frequency $x\in [0,0.3]$ and the other where the function varies slowly $x \in [0.3,1]$. This makes the GP regression not suitable for this function.
\begin{figure}[h]
\center
\includegraphics[scale=0.5]{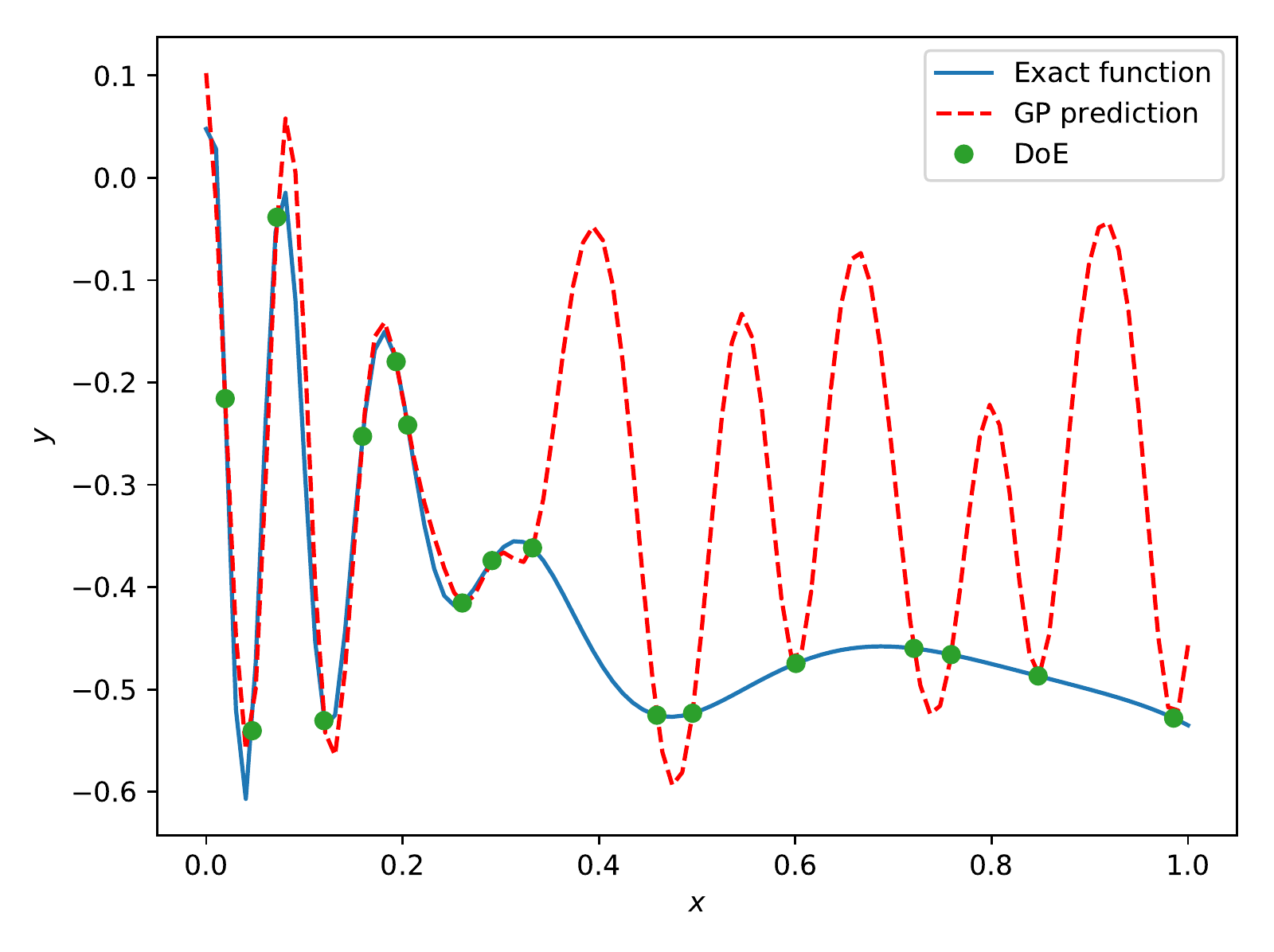}
\caption{Approximation of the modified Xiong-function, a non-stationary 1-dimensional function by a standard GP model. The model can not capture the stability of the region $[0.4,1]$ and continues to oscillate.}
\label{ns1DGP}
\end{figure}

To overcome this issue different GP adaptations to non-stationarity have been proposed. These adaptations are presented in the present Section.

\subsubsection{Direct formulation of non-stationary kernels}

\label{sssec:122}

Most of the methods in literature that use a direct formulation of a non-stationary covariance function, follow the work of Higdon \textit{et al.} \cite{higdon1999non}. The main idea is to use a convolution product of  spatially-varying kernel functions to define a new class of kernels that is non-stationary:
\begin{equation}
k^{NS}(\textbf{x}_i,\textbf{x}_j)= \int_{\mathbb{R}^d} k^S(\textbf{x}_i,\textbf{u}) k^S(\textbf{x}_j,\textbf{u})\text{d}\textbf{u}
\end{equation}
where $k^S$ is a stationary covariance function, $\textbf{x}_i,\textbf{x}_j$ are locations in $\mathbb{R}^d$. 
Higdon \textit{et al.} \cite{higdon1999non} give an analytical form of the non-stationary covariance resulting from convolving Gaussian kernels. Paciorek \cite{paciorek2006spatial} extends this approach by giving the analytical form of the non-stationary covariance function resulting from convolving any stationary kernels:
\begin{equation}
k^{NS}(\textbf{x}_i,\textbf{x}_j)=|\Sigma_i|^{\frac{1}{4}}|\Sigma_j|^{\frac{1}{4}}\left|\frac{\Sigma_i+\Sigma_j}{2}\right|^{-\frac{1}{2}} k_*^S(\sqrt{Q(\textbf{x}_i,\textbf{x}_j)})
\
\label{NSpaciorek}                                                                                                                                                                                                                                                                                                                                                                                                                                                                                                                                                                                                                                                                                                                                                                                                                                                                                                                                                                                                                                                                                                                                                                                                                                                                              
\end{equation}
where:
\begin{equation}
Q(\textbf{x}_i,\textbf{x}_j)=(\textbf{x}_i-\textbf{x}_j)^T {\left(\frac{\Sigma_i+\Sigma_j}{2}\right)}^{-1}(\textbf{x}_i-\textbf{x}_j)
\end{equation}
and $\Sigma_i=\Sigma(\textbf{x}_i)$ is a $d \times d$ matrix-valued function which is positive definite for all $\textbf{x}$ in $\mathbb{R}^d$. In the stationary case $\Sigma(\cdot)$ is a constant arbitrary matrix (often a diagonal matrix containing the lengthscales of each dimension). The interesting observation is that in the resulting non-stationary covariance function $k^{NS}(\cdot,\cdot)$, the Mahalanobis distance $\sqrt{(\textbf{x}_i-\textbf{x}_j)^T \Sigma^{-1} (\textbf{x}_i-\textbf{x}_j)}$ is not used in the stationary covariance function $k_*^S(\cdot)$. Instead, the square root of $Q(\textbf{x}_i,\textbf{x}_j)$ is used, that is a quadratic form with the average of the kernel matrices $\Sigma(\cdot)$ in the two locations. Paciorek \cite{paciorek2006spatial} gives the special case of a non-stationary Mat\'ern covariance function using Eq.(\ref{NSpaciorek}). The construction of the kernel matrix $\Sigma(\cdot)$ for each $\textbf{x}$ in the domain is performed \textit{via} an eigendecomposition parametrization, which can be difficult when increasing the space dimension. Gibbs \cite{gibbs1998bayesian} proposes a simpler parametrization by choosing the matrix $\Sigma(\textbf{x})$ as a diagonal matrix of lengthscales, hence, obtaining lengthscales depending on the location of \textbf{x}. This class of approaches due to its high parametrization requirements (defining a kernel matrix for each location) may not be suitable for high-dimensional problems.

\subsubsection{Local stationary covariance functions}
\label{sssec:123}
The local stationary approaches are based on the idea that non-stationary functions have a locally stationary behavior. Haas \cite{haas1990kriging} proposes a moving window approach where the training and prediction regions move along the input space using a stationary covariance function. This window has to be restrained enough, so that along this window the function is stationary. Other methods consist in dividing the input space into different subsets and using a different model for each subset, this is also known as mixture of experts. Rasmussen and Gharmani \cite{rasmussen2002infinite} propose a mixture of GP experts, that is different stationary GPs in different subspaces of the input space. The division of the input space is performed by a gating network. In this approach the learning dataset is also partitioned, meaning that each model is trained using the dataset in its own region. Using the same concept of local GPs, the Tree Gaussian Process approach of Gramacy \cite{gramacy2008bayesian} can be cited.

However, these approaches present some limitations. Indeed, in computationally expensive problems, data are sparse and using a local surrogate model with sparser data may be problematic, due to the poor prediction capability especially for high dimensional problems.

\subsubsection{Warped GPs}
\label{sssec:124}
These approaches first introduced by Sampson \textit{et al.} \cite{sampson1992nonparametric}, also called non-linear mapping, consist of a deformation of the input space in order to express the non-stationarity using a stationary covariance function. Specifically, a stationary covariance function $k^S(\cdot,\cdot)$, and a function $w(\cdot):\mathbb{R}^d \rightarrow \mathbb{R}^d$ are considered, then the non-stationary covariance function is obtained by simply combining $w(\cdot)$ and $k^S(\cdot,\cdot)$: 
\begin{equation}
k^{NS}(\textbf{x}_i,\textbf{x}_j)=k^S(w(\textbf{x}_i),w(\textbf{x}_j))
\end{equation}
The difficult task in this class of approaches is the estimation of $w(\cdot)$. Gibbs approach that was cited in the direct formulation methods can also be obtained \textit{via} non linear mapping. It consists in considering $w(\cdot)$ as a multidimensional integral of non-negative density functions. These density functions are defined as a weighted sum of $l$ positive Gaussian radial basis functions. The drawback of this approach is the number of parameters equal to $l \times d $. Moreover, the number of radial basis functions $l$ needed to capture the non-stationarity increases with the dimension of the space $d$, inducing an over-parametrized structure of the covariance function in high-dimensional situations. To overcome this issue, the non-linear mapping approach proposed by Xiong \textit{et al.} uses a piece-wise linear density function with parametrized knots. Hence, reducing the number of parameters. However, the deformation in this approach is done only along canonical axes. Marmin \textit{et al.} \cite{marmin2018warped} adress this issue by introducing a parametrized matrix $A$ allowing a linear mapping of the input space before undergoing the non-linear mapping of $w(\cdot)$. The non-linear mapping approach was studied in the context of BO in \cite{toal2012non} where it was compared to regular GP. This allowed the authors to set up a new approach mixing regular GP with non-linear mapping when dealing with BO called Adaptive Partial Non-Stationary (APNS).   

These approaches have some limitations when dealing with sparse data or relatively high-dimensional problems. To overcome these issues, another approach may be the use of Deep Gaussian Processes to handle non-stationarity. 
 
\subsection{Deep Gaussian Processes}
\label{ssec:13}
\subsubsection{Definition}
\label{sssec:131}

The intuition behind using the concept of Deep Gaussian Process (DGP) for non-stationary functions, comes from the Deep Learning theory. The basic idea is to capture multiple variations through the composition of multiple functions. Hence, learning highly varying functions using composition of simpler ones \cite{lecun2015deep}.\\
Following this intuition, Damianou and Lawrence \cite{damianou2013deep} developed DGP, a nested structure of GPs considering the relationship between the inputs and the final output as a functional composition of GPs (Figure~\ref{DGP}):
\begin{equation}
y=f_{L-1}(\ldots\textbf{f}_l(\ldots(\textbf{f}_1(\textbf{f}_0(\textbf{x})))))+\epsilon
\label{eq:}
\end{equation}
where $L$ is the number of layers and $\textbf{f}_l(\cdot)$ is an intermediate GP. Each layer $l$ is composed of an input node $H_{l}$, an output node $H_{l+1}$ and a multi-output GP $\textbf{f}_l(\cdot)$ mapping between the two nodes, getting the recursive equation: $H_{l+1}=\textbf{f}_l(H_{l})$. A Gaussian noise $\epsilon\sim\mathcal{N}(0,\sigma^2)$ is introduced such as $\textbf{y}=f_{L-1}(H_{L-1})+\epsilon$. The one column matrix $H_L=f_{L-1}(H_{L-1})$ refers to the noise free version of $\textbf{y}$. An exploded view showing the multidimensional aspect of DGPs is illustrated in Figure~\ref{DGP2}.

\tikzstyle{medium} = [circle, draw, thin, fill=blue!20, minimum height=4em]
\tikzstyle{textt} = [circle, thin, fill=white,align=center, minimum height=2.5em]
\begin{figure}[!h]
\center
\begin{tikzpicture}[node distance=2.5cm, auto,>=latex', thick]
\path[use as bounding box] (2,-4) rectangle (10,1);
\node[medium,fill=white,rectangle,text width=0.8cm,align=center] (x) {$X$ };
\path[->] node[medium,fill=gray!20, right of= x] (h1) {$H_1$}
                  (x) edge node [above=0.7cm] {\tiny $\textbf{f}_0\sim \mathcal{GP}(0,K_{XX})$} (h1);
\path[->] node[medium,fill=gray!20, right of = h1] (h2) {$H_2$}
                  (h1) edge node [above=0.7cm] {\tiny $\textbf{f}_1\sim \mathcal{GP}(0,K_{H_1H_1})$} (h2);                 
\path[->] node[textt,right of=  h2] (tt) {...};
            
%\path[->] node[medium, above of=h2] (h1) {$\textbf{h}_1$}
%                  (x) edge node {$\textbf{v}_1$} (h1);
\path[->] node[medium,fill=gray!20, right of= tt] (hN) {$H_{L-1}$};
\path[->] node[medium,fill=white, right of= hN] (y) {$\textbf{y}$}
                  (hN) edge node [above=0.7cm]  {\tiny $f_{L-1} \sim \mathcal{GP}(0,K_{H_{L-1} H_{L-1}})+\epsilon$}  (y);
                  
\node[medium,fill=white, rectangle,text width=0.8cm,align=center,below=0.5cm of x] (x) {${X}$ };
\node[ right =1cm of x] (pp) {A deterministic observed variable};
\path[->] node[medium,fill=gray!20, below =0.3cm of x] (h1) {$H_i$};
\node[ right =1cm of h1] (pp) {A distribution with \textbf{non}-observed realizations};
\path[->] node[medium,fill=white, below =0.3cm of h1] (y) {$\textbf{y}$};
\node[ right =1cm of y] (pp) {A distribution with observed realizations};
                  
\end{tikzpicture}
\vspace{2cm}
\caption{A representation of the structure of a DGP}
\label{DGP}
\end{figure}
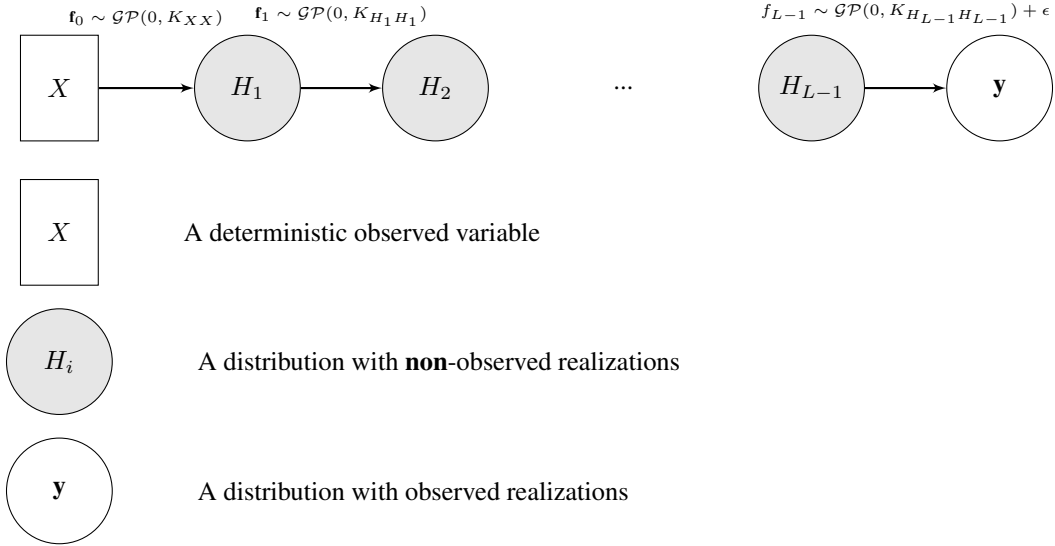
\tikzset{
    role/.style = {rounded rectangle, minimum size=6mm, very thick, draw=blue!50,top color=white,bottom color=blue!20, font=\ttfamily    \scriptsize,align=center,text=black},
    hv/.style = {to path={-| (\tikztotarget)}},
    vh/.style = {to path={|- (\tikztotarget)}},
    comp/.style = {Diamond-},
    ass/.style = {-{>[width=2pt 2]}},
    imp/.style = {-{Latex[open]}},
    proh/.style = {{Bar[sep=1pt]}-{Bar[sep=1pt]}}
}

\begin{figure}[!h]
\center
\begin{tikzpicture}[node distance=2.5cm, auto,>=latex', thick]
\path[use as bounding box] (2,2.5) rectangle (10,-3.5);
\node[medium,fill=white,rectangle,text width=0.8cm,align=center] (x) {$X$ };
\path[->] node[medium,fill=gray!20, above right   =1 cm and 1.6 cm of  x] (h11) {$H_{1,1}$}
                  (x) edge node  { } (h11);
%\path[->] node[medium,fill=gray!20,right of= x] (h12) {$h_{12}$}
%                     
                  
\path[->] node[right of= x] (h12) {$\vdots$}
	(x) edge node  {$\textbf{f}_0$ } (h12) ;
                     
\path[->] node[medium,fill=gray!20, below right =1cm and 1.6 cm of  x] (h13) {$H_{1,d_1}$}
                  (x) edge node  { } (h13) ;

\path[->] node[medium,fill=gray!20,right of= h11 ] (h21) {$H_{2,1}$}
                  (h11) edge node { } (h21)
                  (h12) edge node { } (h21)  
                  (h13) edge node {} (h21);   
\path[->] node[right of= h12] (dots) {$\vdots$}
				
	(h12) edge node  { } (dots)
	 (h11) edge node  { } (dots)
	 (h13) edge node  { } (dots);  
	                
\path[->] node[medium,fill=gray!20,  right of=   h13] (h22) {$H_{2,d_2}$}
                  (h11) edge node {} (h22)
                  (h12) edge node { } (h22)  
                  (h13) edge node {$\textbf{f}_1$ } (h22);
                  
\path[->] node[textt,right of = dots] (tt) {...}; 
\path[->] node[right of = tt] (dots) {$\vdots$}; 
\path[->] node[medium,fill=gray!20, above  = 1cm of dots ] (hL1) {$H_{L-1,1}$};
\path[->] node[medium,fill=gray!20, below =  1 cm of dots,] (hL4) {\hspace{-0.5cm}\begin{varwidth}{0.8cm} \begin{tabular}{c}
$H_{L-1},$\\
$_{d_{L-1}}$ \\
\end{tabular} \end{varwidth}}; 
\path[->] node[medium,fill=white,right of= dots] (y) {$\textbf{y}$}
                  (hL1) edge node {} (y)
                  (dots) edge node {$f_{L-1}+\epsilon$} (y)
                  (hL4) edge node {} (y);
                  
\end{tikzpicture}
\caption{An exploded view of the structure of a DGP}
\label{DGP2}
\end{figure}
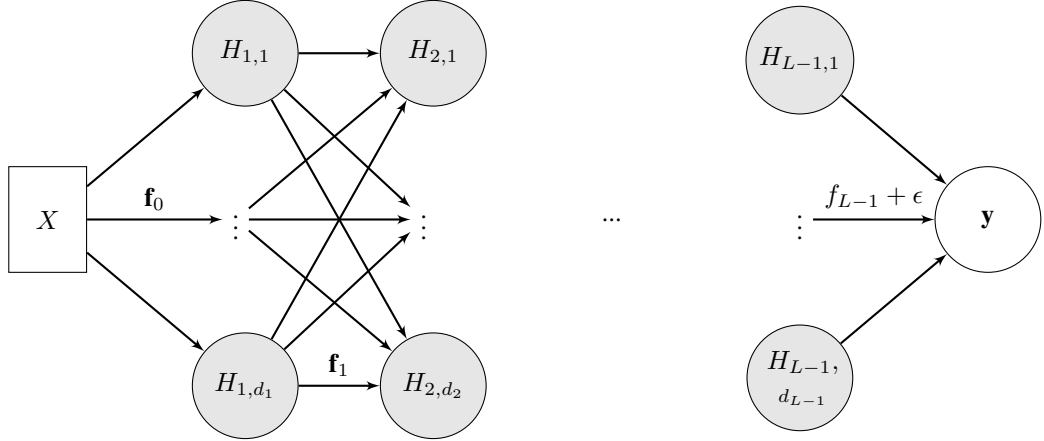

\begin{figure}[h]
\center
\includegraphics[scale=0.5]{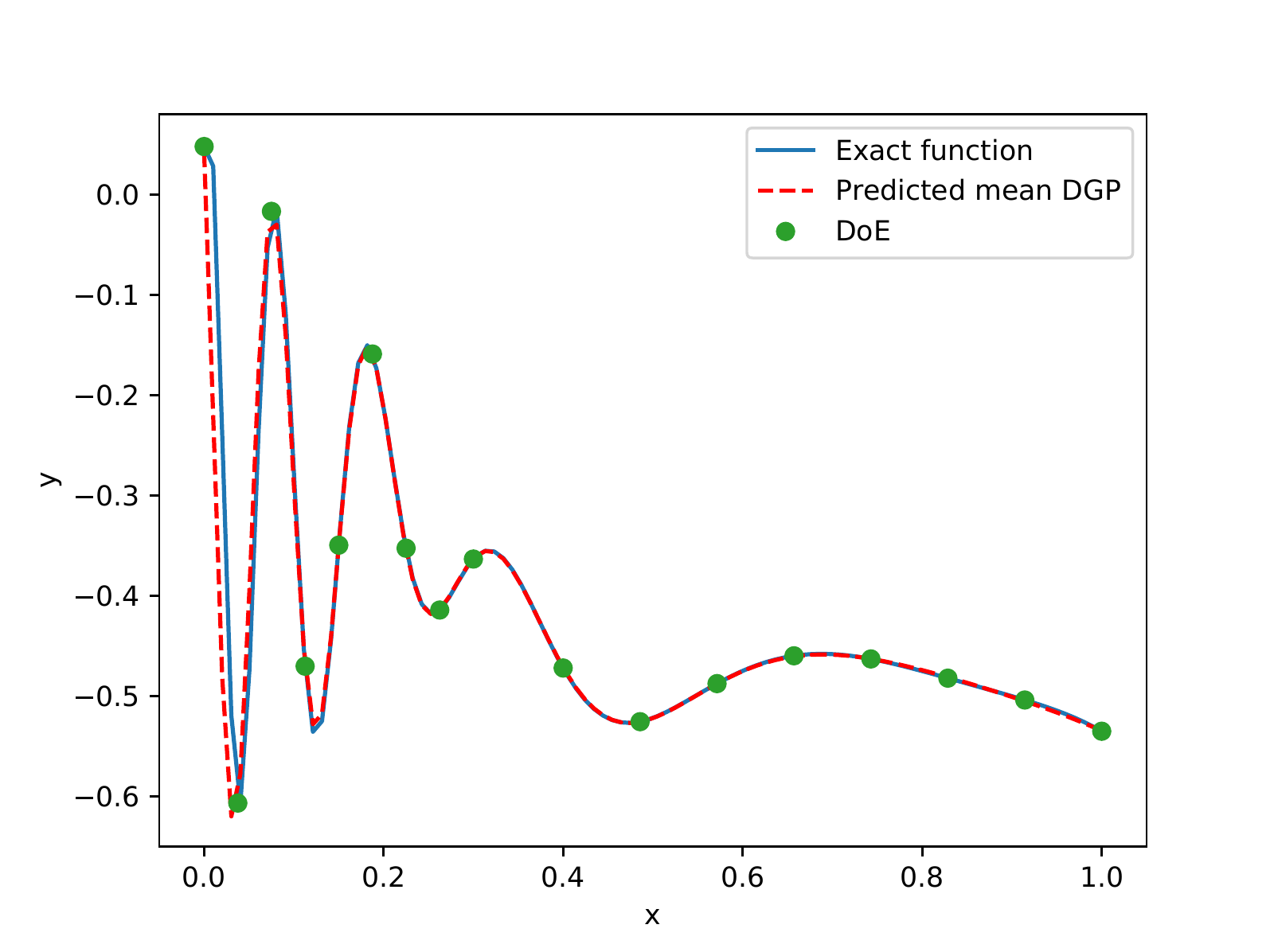}
\caption{Approximation of the modified Xiong-function, a non-stationary 1-dimensional function by a 2 layers DGP model. The model captures the non-stationarity of the real function.}
\label{approx_xiong_DGP}
\end{figure}

This hierarchical composition of GPs presents better results than regular GPs in the approximation of complex functions \cite{damianou2013deep} \cite{dai2015variational} \cite{salimbeni2017doubly} (see Figure~\ref{approx_xiong_DGP}). In fact, DGP allows a flexible way of kernel construction through input warping and dimensionality expansion to better fit data. 

In GP regression models, the hyper-parameters involved are the kernel parameters, the mean function parameters and the noise variance. The optimization of these hyper-parameters in the training of GPs is analytically tractable for a Gaussian noise variance. In DGPs, in addition to these parameters considered for each layer, non-observable variables $H_1,\ldots, H_l, \ldots, H_L$ are involved. Hence, the marginal likelihood for DGP can be written as: 

\begin{eqnarray}
p\left(\textbf{y}|X\right)&=&\int_{H_1}\ldots\int_{H_l}\ldots\int_{H_L} p\left(\textbf{y},H_1,\ldots,H_l,\ldots,H_L|X\right)\text{d}H_1 \ldots \text{d}H_l \ldots \text{d}H_L \nonumber\\
&=&\int_{\{H_l\}_1^L }p\left(\textbf{y},\{\textbf{h}_l\}_1^L|X\right)\text{d}\{H_l\}_1^L \nonumber \\ 
&=&\int_{\{H_l\}_1^L} p(\textbf{y}|H_L)p(H_L|H_{L-1})\ldots p(H_1|X) \text{d}\{H_l\}_1^L 
\end{eqnarray}
where $\{H_l\}_1^L$ is the set of non-observable (latent) variables $\{H_1,\ldots,H_L\}$.\\
%\begin{equation}
%p\left(\textbf{y}|\mathcal{X}\right)=\int_{\{\textbf{h}_l\}_1^L }p\left(\textbf{y},\{\textbf{h}_l\}_1^L|\mathcal{X}\right)\text{d}\{\textbf{h}_l\}_1^L=\int_{\{\textbf{h}_l\}_1^L} p(\textbf{y}|\textbf{h}_L)p(\textbf{h}_L|\textbf{h}_{L-1})...p(\textbf{h}_1|\mathcal{X}) \text{d}\{\textbf{h}_l\}_1^L
%\label{eq:dgp1}
%\end{equation}
The computation of this marginal likelihood is not analytically tractable. Indeed, $p(H_{l+1}|H_{l})$ non-linearly involves the inverse of the covariance matrix $K_{H_lH_l}$, which makes the integration of the conditional probability $p(H_{l+1}|H_{l})$ with respect to $H_{l}$ analytically not tractable.

%The analytical computation of this likelihood is intractable. Indeed, the conditional probability $p(\textbf{h}_{l+1}|\textbf{h}_{l})$ contains the latent variable $\textbf{h}_{l}$ non-linearly inside the inverse of the covariance matrix $K_{\textbf{h}_l \textbf{h}_l}$.

To overcome this issue the marginal likelihood is approximated using an approximate inference technique. Several approaches such as variational inference \cite{damianou2013deep}, Expectation propagation \cite{bui2016deep}, Markov chain Monte Carlo techniques  \cite{salimbeni2017doubly} \cite{NIPS2018_7979} have been developed and are briefly discussed in this Section.
\subsubsection{Direct variational inference approach}
\paragraph{A Variational Bayes approach.}
Let Y and Z be respectively a set of observed and latent variables. Variational Bayes methods consist in approximating the posterior distributions of the set of latent variables $p(Z|Y)$ by a variational distribution $q(Z)$ belonging to a family of distributions which has a simple form (usually the exponential family). By marginalizing over the latent variables, $p(Y)$ can be rewritten as:
\begin{eqnarray}
p(Y)&=& \int_{Z} p(Y,Z) \text{d}Z \nonumber\\
&=&\int_{Z} p(Y,Z) \frac{q(Z)}{q(Z)}\text{d}Z \nonumber\\
&=& \mathbb{E}_q \left[\frac{p(Y,Z)}{q(Z)}    \right]
\end{eqnarray}	
Since the log function is a concave function using Jensen's inequality on the expectation gives: 
\begin{eqnarray}
\log p(Y)&=& \log \left( \mathbb{E}_q \left[\frac{p(Y,Z)}{q(Z)}    \right]\right) \nonumber\\
&\geq & \mathbb{E}_q \left[\log\left(\frac{p(Y,Z)}{q(Z)} \right)    \right]
\label{ELBO}
\end{eqnarray}
This bound obtained is called the Evidence Lower Bound (ELBO). Then, the ELBO is maximized in order to obtain a tight bound on the likelihood $p(\textbf{Y})$. It can be shown that maximizing the ELBO is equivalent to minimizing the Kulback\text{-}Liebler (KL) divergence between $q(Z)$ and $p(Z|Y)$ \cite{damianou:thesis15}. 

Therefore, by introducing a variational distribution $q\left(\left\{H_l\right\}_1^L\right)$ in the case of DGPs and by directly applying the previous inequality Eq.(\ref{ELBO}), the following result is obtained:
\begin{eqnarray}
\log p(\textbf{y}|X)&\geq & \mathbb{E}_q \left[\log\left(\frac{p(\textbf{y},\left\{H_l\right\}_1^L| X )}{q\left(\left\{H_l\right\}_1^L\right)} \right)    \right] \nonumber\\
&\geq & \mathbb{E}_q \left[\log\left(p\left(\textbf{y}|\left\{H_l\right\}_1^L, X \right) \right)\right] + \mathbb{E}_q  \left[ \log \left (\frac{p\left(\left\{H_l\right\}_1^L| X \right)}{q\left(\left\{H_l\right\}_1^L\right)} \right)  \right] \nonumber\\
&\geq & \mathbb{E}_q \left[\log\left(p\left(\textbf{y}|\left\{H_l\right\}_1^L, X \right) \right)\right] - KL \left (q\left(\left\{H_l\right\}_1^L\right)|| p\left(\left\{H_l\right\}_1^L| X \right) \right)\label{ELBO1} 
\end{eqnarray}
The second term in Eq.(\ref{ELBO1}) is the KL divergence between the variational distribution and the prior distribution on the latent variables. The KL divergence is analytically tractable if the prior and the variational distributions on the latent variables are restrained to Gaussian distributions. However, the first term is still analytically intractable since it still involves the integration of the inverse of the covariance matrices with respect to the latent variables. To overcome this issue, Damianou \textit{et al.} \cite{damianou:thesis15} followed the work of Titsias and Lawrence \cite{titsias2010bayesian} in the context of Bayesian Gaussian Process Latent Variable Model by introducing a set of inducing variables to obtain an analytical tractable lower bound.
\paragraph{Introduction of inducing variables.} Inducing variables were first introduced in the context of sparse GP \cite{snelson2006sparse} \cite{titsias2009variational}. To overcome the drawback of regular GP that involves the inversion of the covariance matrix of the whole dataset $K_{XX} \in \mathcal{M}_{nn}$, sparse GPs introduce a set of latent variables consisting of an input-output pairs $Z$ and $\textbf{u}$: 

$(\text{Induced variables})\left\{\begin{array}{rcl} Z &=&\left[\textbf{z}^{(1)},\ldots,\textbf{z}^{(m)}\right]^\top, \textbf{z}^{(i)} \in \mathbb{R}^d,  \forall i \in \{ 1,\ldots,m \} \\ \textbf{u}&=&\left[u^{(1)}=f\left(\textbf{z}^{(1)}\right),\ldots,u^{(m)}=f\left(\textbf{z}^{(m)}\right)\right]^\top \end{array} \right.$\\
with $m<<n$. The idea is to choose $Z$ and $\textbf{u}$ in order to explain the statistical relationship between $X$ and $\textbf{y}$ by the statistical relationship between $Z$ and $\textbf{u}$. Then, for the training and prediction of the sparse GP, the matrix to be inverted belongs to $\mathcal{M}_{mm}$, hence, achieving reduction in the computational complexity of GP. 

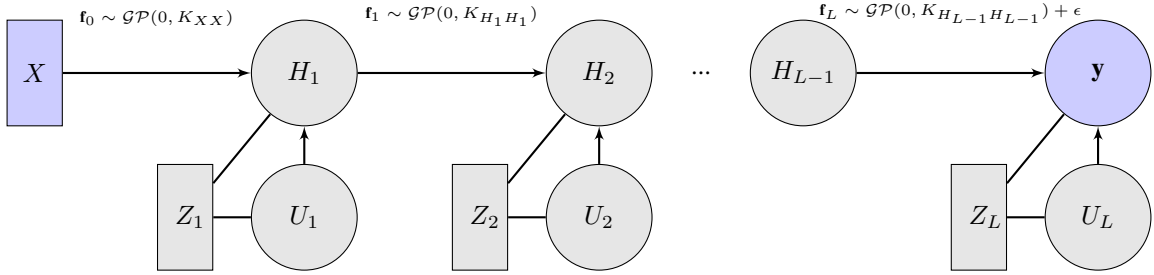
\begin{figure}[!h]

\center

\begin{tikzpicture}[node distance=2.5cm, auto,>=latex', thick]
\node[medium,rectangle,text width=0.5cm,align=center] (x) {${X}$ };

\path[->] node[medium,fill=gray!20, right =of x] (h1) {$H_1$}
                  (x) edge node [above=0.5cm] {\tiny $\textbf{f}_0\sim \mathcal{GP}(0,K_{XX})$} (h1);
\path[->] node[medium,fill=gray!20, below =0.5cm of h1] (u1) {$U_1$}
                  (u1) edge node [above=0.5cm] {} (h1);       
\node[medium,rectangle,fill=gray!20,text width=0.5cm,left =0.5cm of u1] (z1) { $\text{  }  {Z_1}$}
           (z1) edge node {} (u1) 
           (z1) edge node {} (h1);  
\path[->] node[medium,fill=gray!20, right =of h1] (h2) {$H_2$}
                  (h1) edge node [above=0.5cm] {\tiny $\textbf{f}_1\sim \mathcal{GP}(0,K_{H_1 H_1})$} (h2); 
\path[->] node[medium,fill=gray!20, below =0.5cm of h2] (u2) {$U_2$}
                  (u2) edge node [above=0.5cm] {} (h2);  
\node[medium,rectangle,fill=gray!20,text width=0.5cm,left =0.5cm of u2] (z2) { $\text{  }  {Z_2}$ }
           (z2) edge node {} (u2) 
           (z2) edge node {} (h2);                     
\path[->] node[textt,right = 0.2cm of h2] (tt) {...};
            
%\path[->] node[medium, above of=h2] (h1) {$\textbf{h}_1$}
%                  (x) edge node {$\textbf{v}_1$} (h1);
\path[->] node[medium,fill=gray!20, right =0.2cm of tt] (hN) {$H_{L-1}$};
\path[->] node[medium, right =of hN] (y) {$\textbf{y}$}

                  (hN) edge node [above=0.5cm]  {\tiny $\textbf{f}_L\sim \mathcal{GP}(0,K_{H_{L-1} H_{L-1}})+\epsilon$}  (y);
\path[->] node[medium,fill=gray!20, below =0.5cm of y] (ul) {$U_{L}$}
                  (ul) edge node [above=0.5cm,pos=-0.0005] {} (y); 
        \node[medium,rectangle,fill=gray!20,text width=0.5cm,left =0.5cm of ul] (zl) {$\text{ } {Z_L}$ }
           (zl) edge node [] {} (ul) 
           (zl) edge node {} (y);      
                  
\end{tikzpicture}
\caption{Representation of the introduction of the inducing variables in DGPs}
\label{sparse}
\end{figure}

In each layer of a DGP, a set of inducing variables is introduced  $Z_l =\left[\textbf{z}_l^{(1)},\ldots,\textbf{z}_l^{(m_l)}\right]^\top$, $\textbf{z}^{(i)} \in \mathbb{R}^{d_l},  \forall i \in \{ 1,\ldots,m_l \}$ and $U_l=\textbf{f}_l({Z_l})$ (Figure~\ref{sparse}) (notice here that since the intermediate layers are multi-output GPs, $U_l$ are matrices $\in \mathcal{M}_{nd_l}$ and not vectors, except in the last layer where $U_L$ corresponds to a one column matrix). Henceforth, for notation simplicity, the number of induced inputs in each layer are considered equal to $m$.

Now that the latent space has been augmented with the inducing variables, the posterior of the joint distribution of the latent variables $p\left(\{H_l, U_l\}|\textbf{y},X\right)$ is approximated by a variational distribution $q\left(\{H_l, U_l\}\right)$ with the assumption of independency between layers: 
\begin{equation}
q\left(\{H_l,U_l\}_1^L\right)=\prod_{l=1}^L q(H_l) q(U_l)
\end{equation}
Moreover, for the sake of simplicity the variational distributions are restrained to the exponential family. Then, by using Eq.(\ref{ELBO}) it holds: 
\begin{eqnarray}
\log p(\textbf{y}|X) &\geq&  \mathbb{E}_{q(\{H_l\}_1^L,\{U_l\}_1^L)}\left[\log\frac{p\left(\textbf{y},\{H_l\}_1^L,\{U_l\}_1^L|X,\{Z_l\}_1^L\right)}{q(\{H_l\}_1^L,\{U_l\}_1^L)}\right]=\mathcal{L}
\end{eqnarray}

After using some results from variational sparse GP \cite{titsias2009variational}, an analytical tractable bound is obtained for kernels that are feasibly convoluted with the Gaussian density such as the Automatic Relevance Determination (ARD) exponential kernel. The analytical optimal form of $q(U_l)$ as a function of $q(H_l)$ can be obtained \textit{via} the derivative of the variational lower bound $\mathcal{L}$ with respect to $q(U_l)$. Hence, collapsing $q(U_l)$ in the approximation by injecting its optimal form, allows to obtain a tighter lower bound depending on the following parameters: the kernel parameters $\{\boldsymbol{\Theta}_l\}_{l=1}^{l=L}$, the inducing inputs $\{Z_l\}_{l=1}^{l=L}$
and the variational distributions parameters $\{\bar{H}_l$, $S_l\}_{l=1}^{l=L}$ respectively the mean and covariance matrix defining the variational distributions $\{q(H_l)\sim \mathcal{N}(\bar{H}_l,S_l)\}_{l=1}^{l=L}$

Therefore, training a DGP model comes back to maximizing the evidence lower bound with respect to these parameters:

$$
\begin{array}{ll}

\text{Maximize:} &\mathcal{L}\\ 
\text{According to:} &\{\boldsymbol{\Theta}_l\}_{l=1}^{l=L},\{Z_l\}_{l=1}^{l=L},\{\bar{H}_l\}_{l=1}^{l=L},\{S_l\}_{l=1}^{l=L}\\

\end{array}
$$

The number of hyperparameters to optimize in the training of a DGP is more important than in the training of a regular GP where only the kernel hyperparameters are considered, and more important than variational sparse GP since the number of hyperparameters is considered for each layer. 

\subsubsection{The Doubly stochastic approach}
The Doubly Stochastic approach proposed by Salimbeni \textit{et al.} \cite{salimbeni2017doubly} drops the assumption of independence between layers and the special form of kernels. Indeed, the posterior approximation maintains the exact model conditioned on $U_l$: \begin{equation}
q\left(\{H_l,U_l\}_1^L\right)=\prod_{l=1}^L p(H_l|H_{l-1},U_l) q(U_l)
\end{equation}
However, the analytical tractability of the lower bound $\mathcal{L}$ is not maintained. The variational lower bound is then rewritten as follows (the mention of the dependence on $X$ and $Z_l$ is omitted for the sake of simplicity):
\begin{eqnarray}
\mathcal{L}&=&\mathbb{E}_{q(\{H_l,U_l\}_1^L)}\left[\log\frac{p\left(\textbf{y},\{H_l\}_1^L,\{U_l\}_1^L\right)}{q(\{H_l\}_1^L,\{U_l\}_1^L)}\right] \nonumber\\
&=&\mathbb{E}_{q(\{H_l,U_l\}_1^L)}\left[\log\frac{p\left(\textbf{y}|\{H_l\}_1^L,\{U_l\}_1^L\right)\prod_{l=1}^L p(H_l|H_{l-1},U_l) p(U_l)}{\prod_{l=1}^L p(H_l|H_{l-1},U_l) q(U_l)}\right] \nonumber\\
&=&\mathbb{E}_{q(\{H_l,U_l\}_1^L)}\left[\log\frac{\prod_{i=1}^N p(y^{(i)}|h_L^{(i)}) \prod_{l=1}^L p(U_l)}{\prod_{l=1}^L q(U_l)}\right] \nonumber\\
\mathcal{L}&=&\sum_{i=1}^N \mathbb{E}_{q(h_L^{(i)})} \left[\log p(y^{(i)}|h_L^{(i)})\right] - \sum_{l=1}^L KL\left[q(U_l)||p(U_l)\right]
\label{ds_elbo}
\end{eqnarray}

Keeping the $\{U_l\}_{l=1}^L$ in this formulation of the ELBO instead of collapsing then allows factorization over the data $X,\textbf{y}$ which enables parallelization. The computation of this bound is done by approximating the expectation with Monte Carlo sampling, which is straightforward using the propagation of each data-point $\textbf{x}^{(i)}$ through all the GPs:
\begin{equation}
q(h^{(i)}_L)=\int \prod_{l=1}^{L-1} q\left(\textbf{h}^{(i)}_l|U_l,\textbf{h}^{(i)}_{l-1},Z_{l-1}\right) \text{d}\textbf{h}_l^{(i)}
\end{equation} 
with $\textbf{h}_0^{(i)}=\textbf{x}^{(i)}$. The optimization of this formulation of the bound is done according to the kernel parameters $\{\boldsymbol{ \Theta_l}\}_{l=1}^{l=L}$, the inducing inputs $\{Z_l\}_{l=1}^{l=L}$ and the variational distribution hyperparameters of the inducing variables: $\{q(U_l)\sim \mathcal{N}(\bar{U}_l,{\Sigma}_l)\}_{l=1}^{l=L}$

\subsubsection{Other approaches}
Alternative methods for training a DGP have been proposed. Dai \textit{et al.} \cite{dai2015variational} improved the direct variational approach by instead of considering the parameters of the variational posteriors $q(H_l)$ as individual parameters, considered them as a transformation of observed data $\textbf{y}$ by multi-layers perceptron. Bui \textit{et al.} \cite{bui2016deep} proposed a deterministic approximation for DGP based on an approximated Expectation Propagation energy function, and a probabilistic back\text{-}propagation algorithm for learning. Havasi \textit{et al.} \cite{NIPS2018_7979} gave up the Gaussian approximation of the posterior distribution $\{q(U_l)\}_{l=1}^L$, and proposed the use of Stochastic Gradient Hamiltonian Monte Carlo for this approximation. Moreover, a Markov Chain Expectation Maximization algorithm is developed for the optimization of the hyper-parameters. 

\section{Bayesien Optimization using Deep Gaussian Processes}
\label{sec:2}
%A preliminary work has highlighted the interest and the challenges arising from the combination of DGPs and BO \cite{hebbal2018efficient}.
In this Section, a deep investigation is followed in order to develop the Deep Efficient Global Optimization algorithm (DEGO) a BO \& DGP algorithm. To lead to that, the choices needed to make this coupling possible are discussed. These choices concern the training approach for the DGP, the uncertainty model of the DGP prediction, the infill criteria, the induced variables in each layer and the configuration of the architecture of the DGP (number of layers, number of units, \textit{etc.}).\\

In this Section different functions are used to illustrate the analyses made. These functions are presented in Appendix~\ref{appendixA}.
%In this section, the modified Xiong function (\textit{c.f.} Eq.(\ref{xiong})) and the modified TNK function Eq.(\ref{tnk}) are used to illustrate the impact of the choices made.
\subsection{Training}

\begin{figure}
\begin{minipage}[c]{0.45\linewidth}
\includegraphics[width=0.95\linewidth]{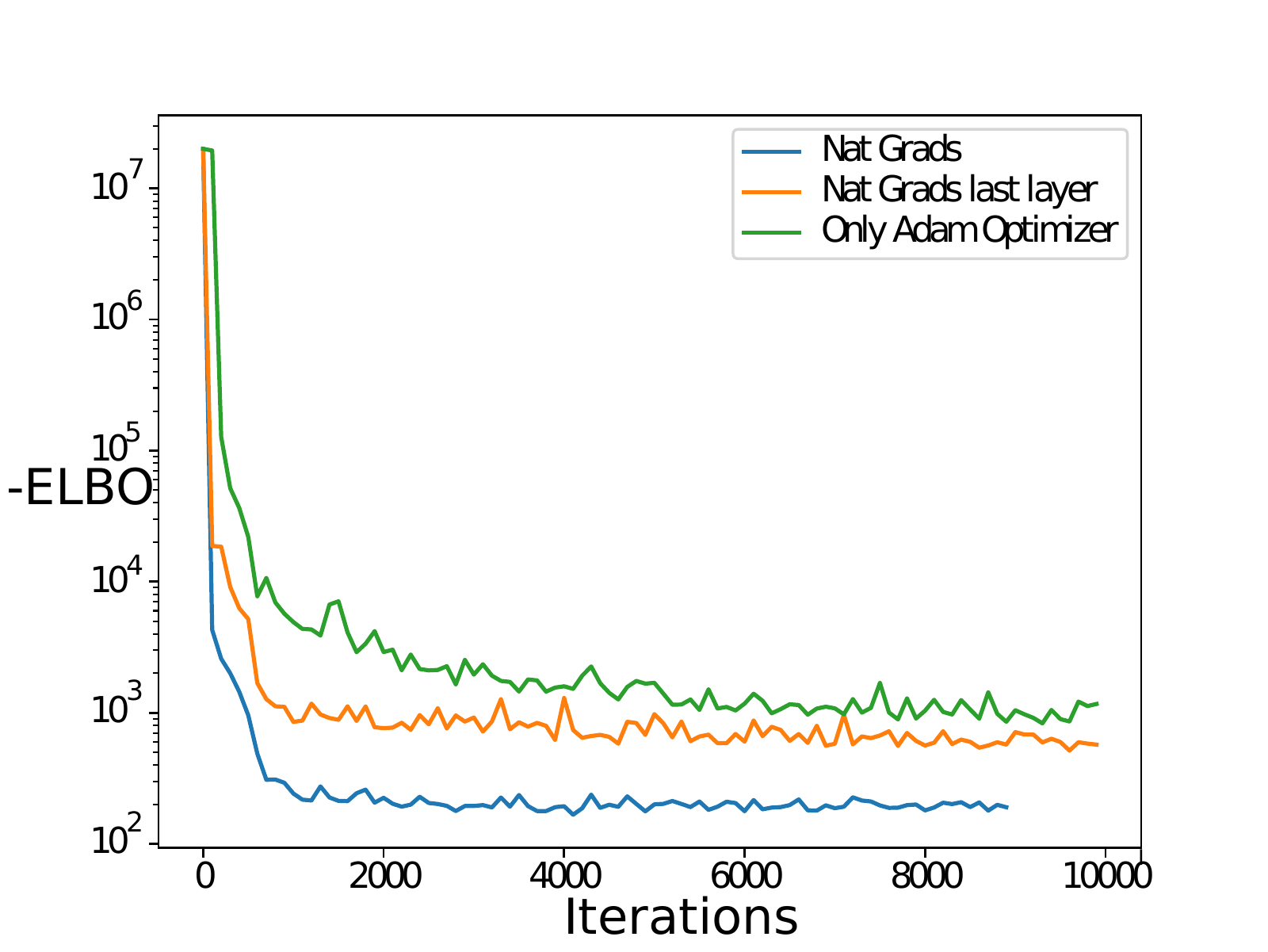}
\center \scriptsize (a) TNK constraint (d=2,n=20) $\gamma^{Adam}=10^{-2}, \gamma^{nat}_l=10^{-1},  \forall 0\leq l\leq2$
\end{minipage}
\hfill
\begin{minipage}[c]{0.45\linewidth}
\includegraphics[width=0.95\linewidth]{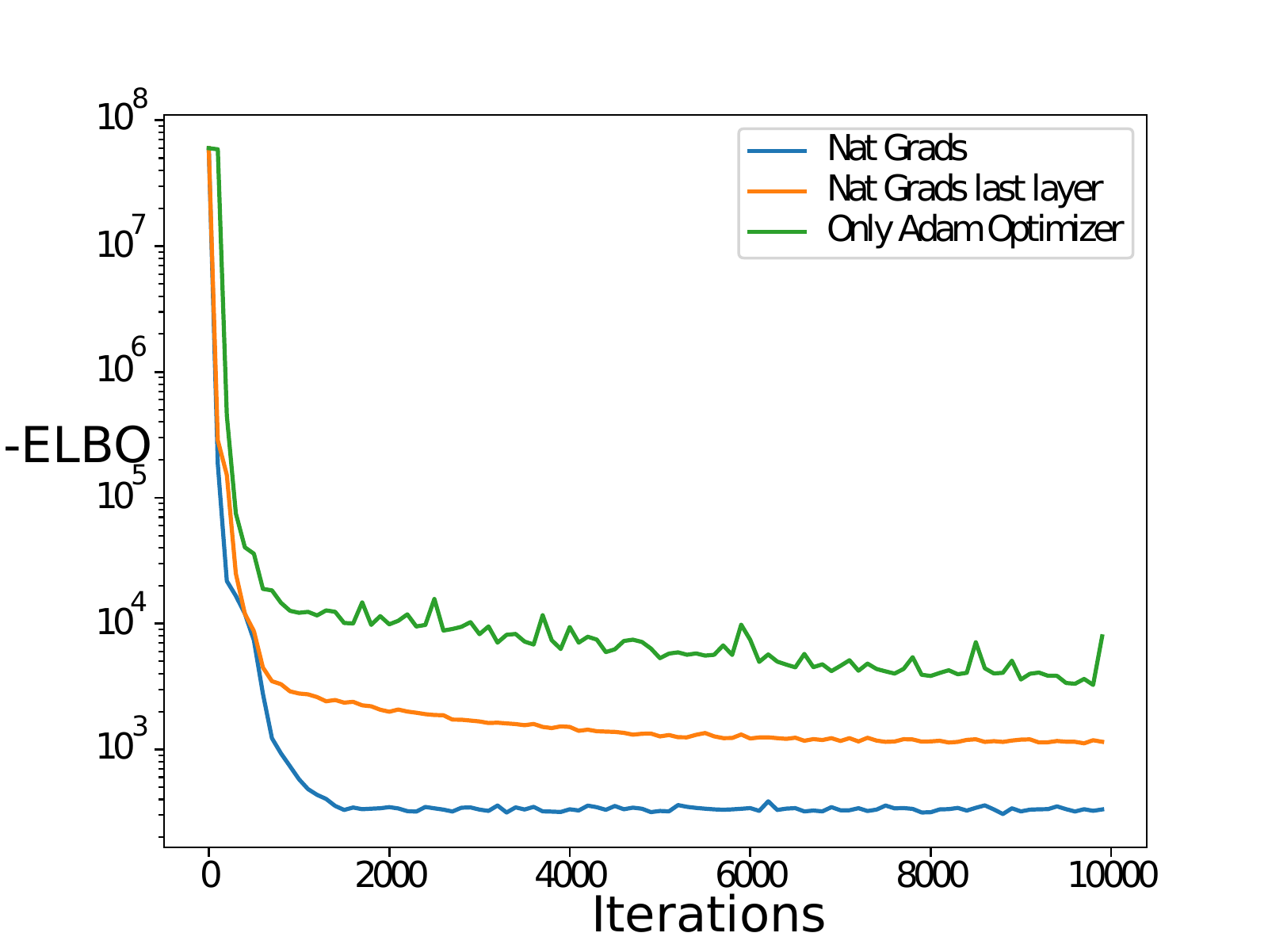}
\center \scriptsize (b) Hartmann 6d (d=6,n=60)\\ $\gamma^{Adam}=10^{-2}, \gamma^{nat}_2=10^{-1}, \gamma^{nat}_0=\gamma^{nat}_1=10^{-2} $
\end{minipage}
\begin{center}
\begin{minipage}[c]{0.45\linewidth}
\includegraphics[width=0.95\linewidth]{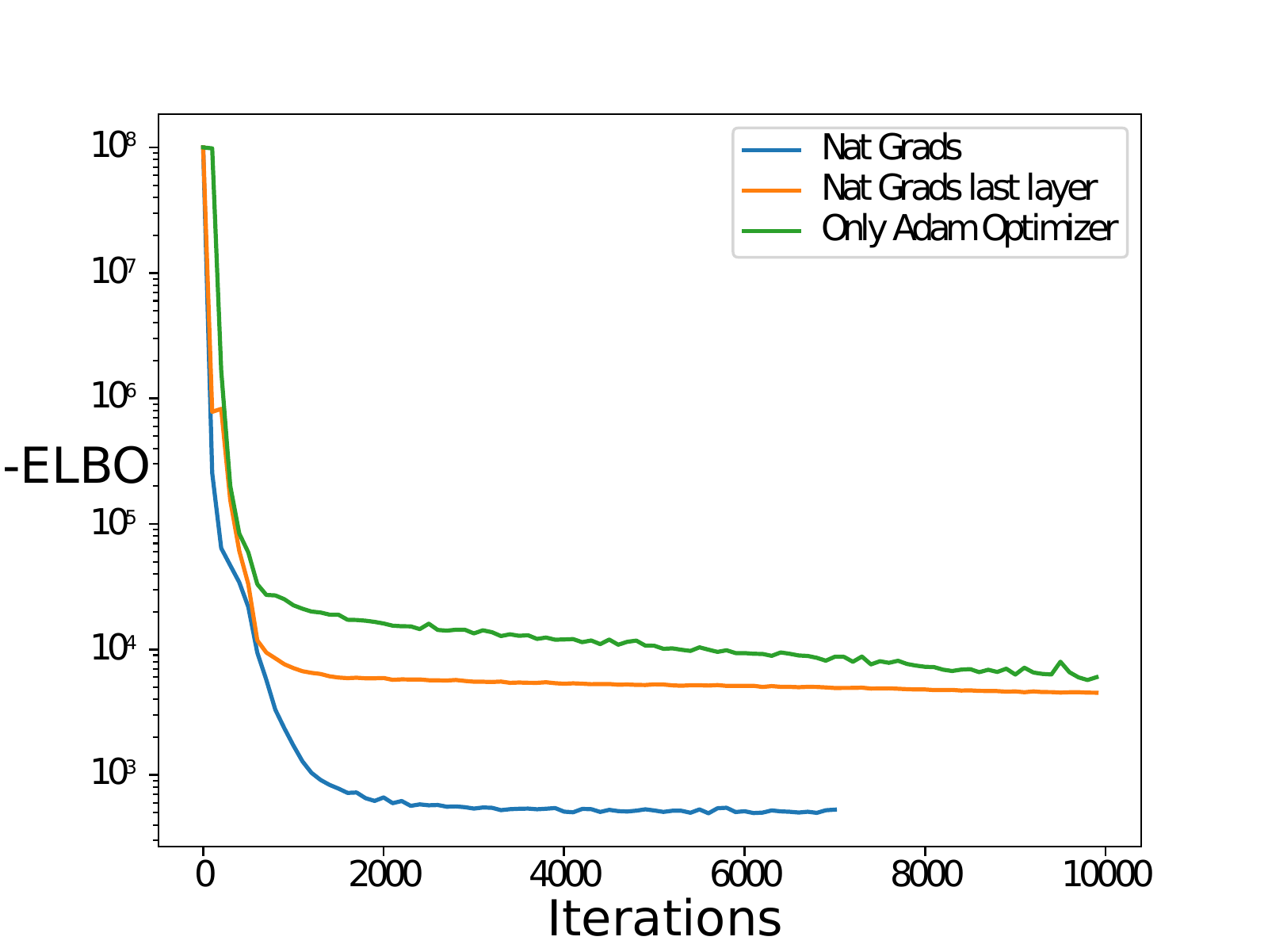}
\center \scriptsize (c) Trid  (d=10,n=100) \\$\gamma^{Adam}=10^{-2}, \gamma^{nat}_2=10^{-1}, \gamma^{nat}_0=\gamma^{nat}_1=10^{-2} $
\end{minipage}
\end{center}
\caption{Comparison of the evolution of the optimization of the ELBO on three different problems using three different optimizations: an alternate optimization using the natural gradient for all the variational parameters and the Adam optimizer for the hyperparameters, an alternate optimization using the natural gradient for the variational parameters of the last layer and the Adam optimizer for the other parameters, and an optimization using the Adam optimizer for all the parameters.  $\gamma^{adam}$ is the step size of the Adam optimizer and $\gamma^{nat}_l$ is the step size of the natural gradient for the variational parameters at layer $l$.}
\label{convergence_elbo}
\end{figure}

In \cite{hebbal2018efficient}, in the experimentations a DGP is used with the variational auto-encoded inference method \cite{dai2015variational}. This approach of training assumes that the variational distributions of the latent variables are factorizable over the layers. As mentioned in the previous Section, this assumption may not be realistic. The doubly stochastic inference approach proposed in \cite{salimbeni2017doubly} is preferred in the present study since it keeps the dependence between layers. The loss of analytical tractability may be compromising, since a Monte Carlo sampling approach is required. However, the form of the Evidence Lower Bound (ELBO) is fully factorizable over the dataset allowing important parallelization.\\
%Table~\ref{training_table} illustrates the Root-Mean-Square Error (RMSE) obtained in the approximation of the modified Xiong function with a DGP with the two training approaches. It is obvious in this experimentation that the doubly stochastic variational approach is more robust. Even if the minmum of the RMSE is better in the variational auto-encoded approach, the maximum RMSE observed is far worse. This patholigical behaviour of the variational auto-encoded approach can be observed in Figure~\ref{patholigical_Dai} where the approximated function under-fit severely the real function.
%Due to the fact that in BO the objective is to reduce time via surrogate models it would not be interesting to repeat the training multiple time until obtaining the better approximation. Moreover, there is not a possibility to evaluate how good the performance of the approximation is, since a test set would be very expensive to obtain, and a cross validation approach would be also time consuming and not pertinent with sparse training data-set. Due to this reasons the doubly stochastic approach will be used in this framework of BO and DGP.
The optimization of the ELBO (Eq.(\ref{ds_elbo})) is performed using a loop procedure consisting of an optimization step with the natural gradient to perform the optimization with respect to the parameters of the variational distributions $\{q(U_l)\}_{l=1}^L$ while fixing the other variables, then an optimization step using a stochastic gradient descent optimizer (Adam optimizer \cite{kingma2014adam}) to perform the optimization with respect to the hyperparameters. This optimization procedure has been done in the case of sparse variational GPs and has shown better results than using only a gradient descent optimizer for all the variables \cite{salimbeni2018natural}. However, using the natural gradient for all the distributions of the inner layers in the case of DGPs is tricky. Indeed, the optimization of the distribution of the variational parameters of the first layers is highly multi-modal. It is therefore more likely to take over large step size (inducing non-positive definite matrix in the natural space). One way to deal with this issue is to use different step sizes for each layer decreasing from the last layer to the first one. \\
The evolution of the ELBO using three different optimization approaches is presented in Figure~\ref{convergence_elbo} for three different problems. The optimization using a loop with a natural gradient step for all the variational distributions parameters and a step with the Adam optimizer for the hyperparameters gives the best results. However, for the Hartmann 6d and the Trid functions the size of the step of the natural gradient for the first layers is reduced compared to the step size of the last layer, in order to avoid overlarge step size.

\begin{figure}
\begin{minipage}[c]{0.45\linewidth}
\center
\includegraphics[width=0.95\linewidth]{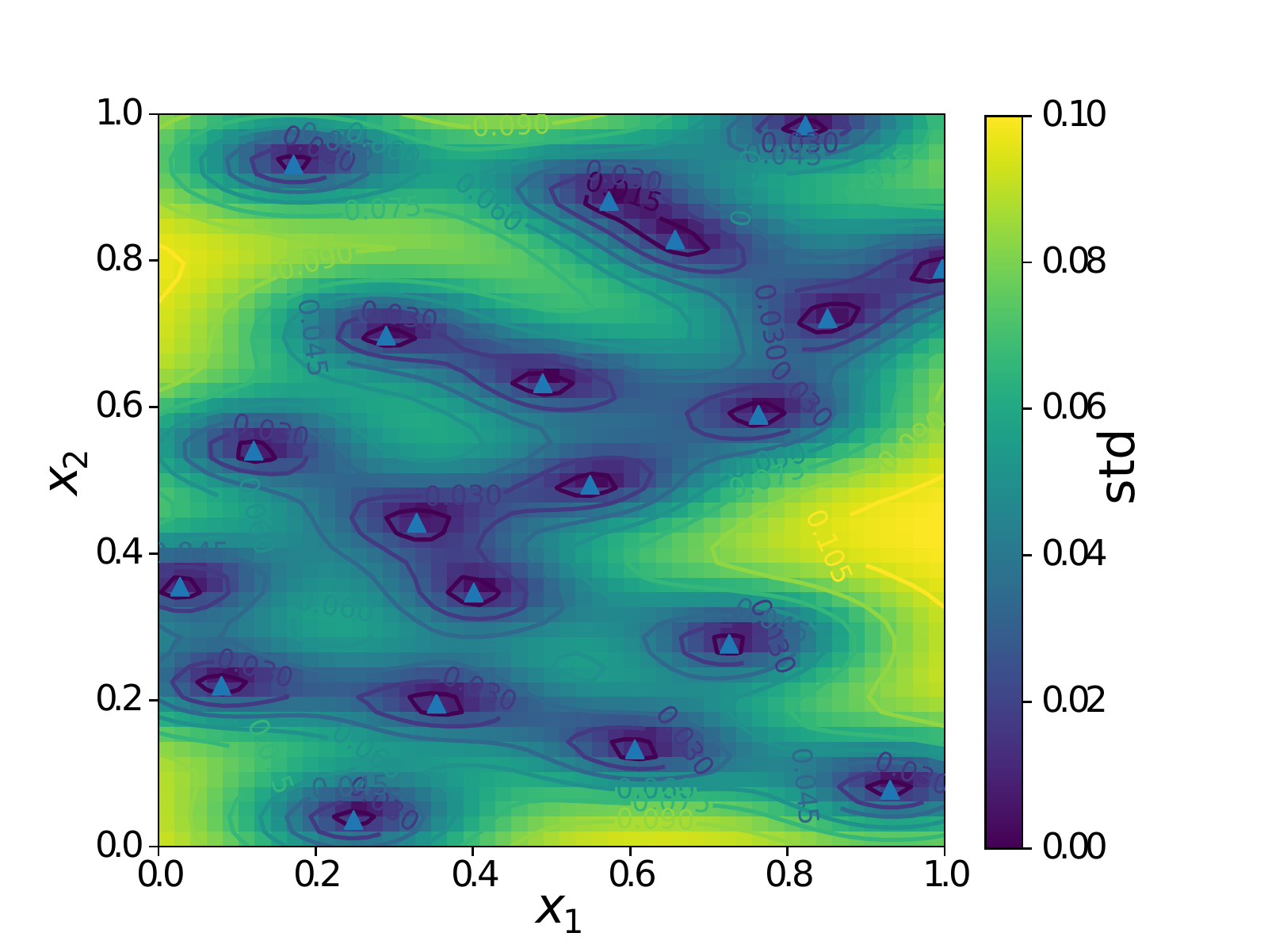}
\end{minipage}
\hfill
\begin{minipage}[c]{0.45\linewidth}
\center
\includegraphics[width=0.95\linewidth]{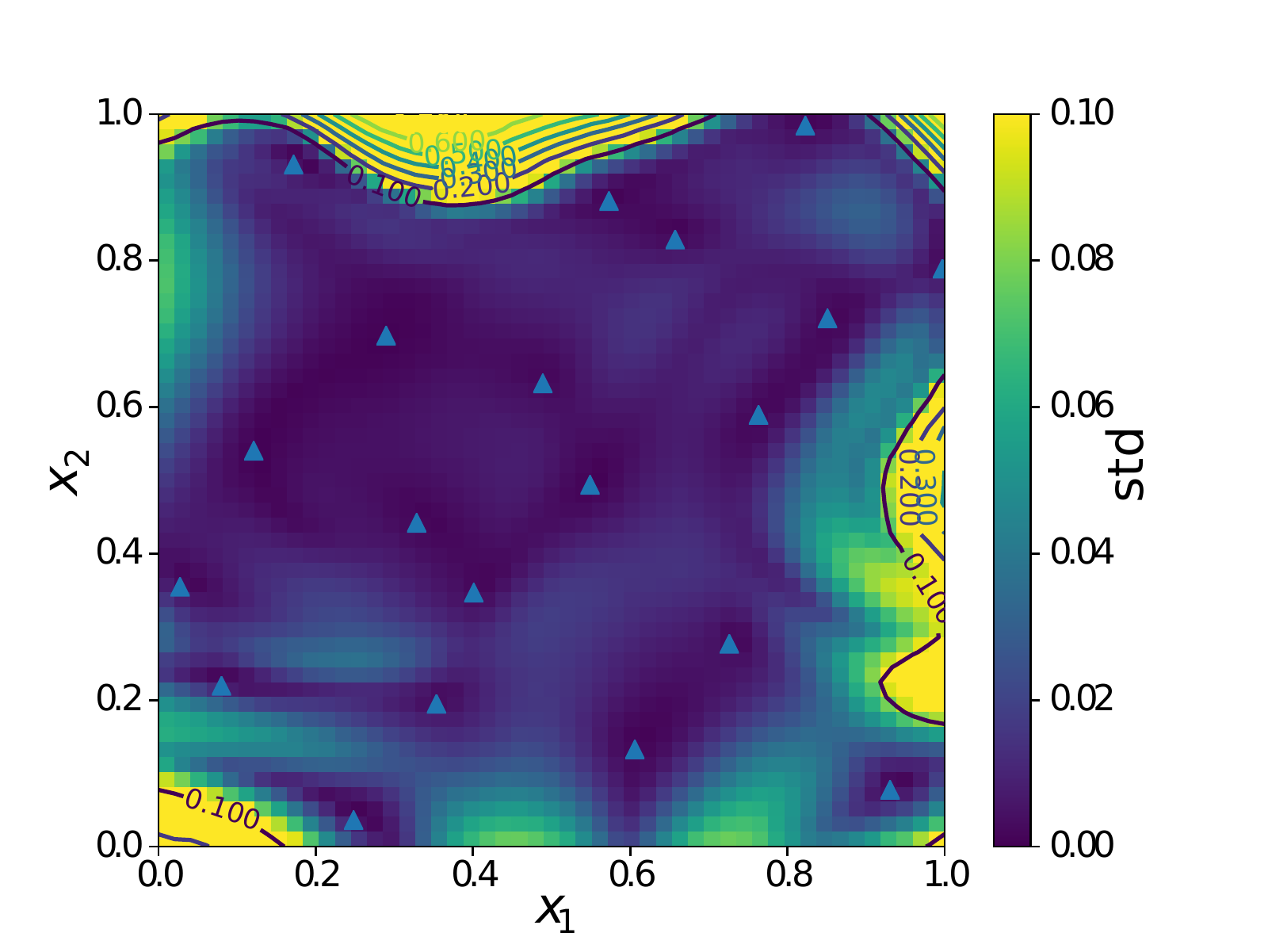}
\end{minipage}
\caption{Standard deviation on the prediction given by a 2 hidden layers DGP model on the TNK constraint function. (left) standard deviation given by a model optimized using natural gradient on all the variational parameters. (right) standard deviation given by a model optimized using natural gradient only on the variational parameters of the last layer. An underestimation of the uncertainty happens in the second approach.}
\label{uncertaintyplot}
\end{figure}

A test set to estimate the Root Mean Square Error (RMSE) and the test log-likelihood is used to assess the prediction and uncertainty performance of the models (Table~\ref{rmse_uncertainty}). The models optimized by the loop natural gradient for the variational distributions parameters of all layers and an Adam optimizer for the hyperparameters, give the best results. It is interesting to notice that the models optimized by the Adam optimizer on all the variables give comparable results on the prediction. However, it happen that they underestimate the uncertainty on the prediction (Figure~\ref{uncertaintyplot}). This explains the value obtained of the test log-likelihood in the case of the DGP model optimized by only an Adam Optimizer compared to the ones given by GP or DGP with natural gradient on all the variational parameters. In the context of BO this uncertainty measure is important for the construction of infill criteria. An underestimated uncertainty will make the BO algorithm sampling around the current minimum limiting thereby its exploration capabilities. Hence, a combination of the natural gradient on all the variational parameters and the Adam optimizer on the hyperparameters is used in DEGO for training the models.

\begin{table}
\center
\begin{tabular}{|K{1.7cm}|K{1.7cm}|K{1.7cm}|K{1.7cm}|K{1.7cm}|K{1.7cm}|}
  \hline Function & Approach & mean RMSE & std RMSE  & mean test log-likelihood & std test log-likelihood \\
  \hline

  TNK & GP & 0.18832 & 0.0305 & -8866.59  & 20166.95   \\
  \cline{2-6}
  constraint & DGP Adam & 0.17746 & 0.0482 & -467207 & 400777 \\
  \cline{2-6}
	& \textbf{DGP Nat} &\textbf{ 0.1659} &\textbf{ 0.013} & \textbf{-3671 }& \textbf{1766.48}  \\
	\hline
	Hartmann & GP & 0.3010 & 0.0311 & \textbf{-566.27} & \textbf{451.798}\\
	\cline{2-6}
	6d & DGP Adam & 0.3166 & 0.0252 & -2595.70 & 2393.99\\
	\cline{2-6}
	& DGP Nat & \textbf{0.2921} &  \textbf{0.0200} & -1386.12 & 1111.29\\
	\hline
	Trid & GP & 12934 & 965 & -10912 &  114\\
	\cline{2-6} & DGP Adam & 11978 & 496.71 & -12690 & 808\\
	\cline{2-6} & \textbf{DGP Nat} &  \textbf{11151} & \textbf{388} & \textbf{-10342} & \textbf{109}\\
	\hline
\end{tabular}

\caption{Comparison of the Root Mean Squared Error (RMSE) and the test log-likelihood and their standard deviations (std) on three different problems with a training data size of density 10 ($n=10 \times d$ where n is the data size and $d$ the input dimension of the function) on 50 repetitions. GP: Gaussian Process with an RBF kernel. DGP Adam: DGP with 2 hidden layers with all its parameters optimized by an Adam Optimizer. DGP Nat: DGP with 2 hidden layers with all the variational parameters optimized by a Natural gradient and the hyperparameters by an Adam Optimizer.}
\label{rmse_uncertainty}
\end{table}

\begin{figure}
\center
\includegraphics[width=0.75\linewidth]{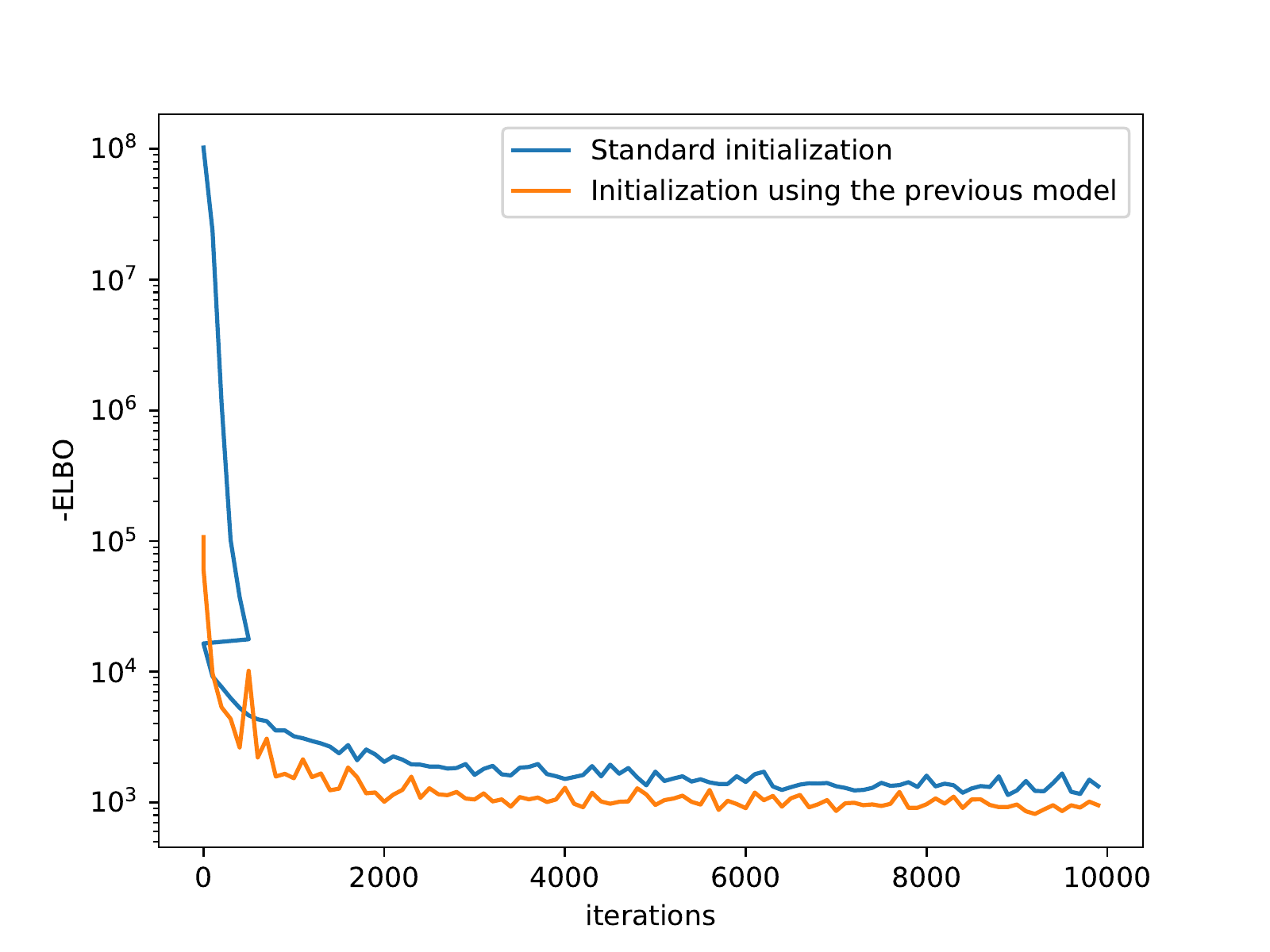}
\caption{Comparison of the evolution of the optimization of the ELBO in the case of using the standard initialization procedure and in the case of using the previous model optimal parameters as the initialization. Using the previous model allows better and faster convergence}
\label{convergenceinit}

\end{figure}

In the context of BO, this optimization procedure has to be repeated after each added point, which may be time consuming. To overcome this issue, the optimization is initialized using the optimal values of the previously trained DGP model. As shown in Figure~\ref{convergenceinit} this allows faster convergence. However, this can make the optimization tricked in bad local optima. Therefore, a complete training of the model is performed after a certain number of iterations depending on the problem at hand.   

The pseudo algorithm (Algorithm~\ref{algorithm1}) describes the training of the DGP model. The initialization can be done from a previous model optimal parameter values to allow faster optimization or can be done from scratch.

\begin{algorithm}[h]
\SetKwProg{try}{try}{:}{}
  \SetKwProg{catch}{catch}{:}{end}
 \textbf{Require:} $X, \textbf{y}$.\\
 \textbf{Require:} The number of induced variables, $m$.\\
 \textbf{Require:} The number of layers, $L$.\\
 \textbf{Require:} The number of iterations, $iter$.\\
 \textbf{Require:} The tolerance on the variation of the ELBO $\Delta_k$, $tol$.\\
 \textbf{Require:} The step sizes $\gamma^{Adam}$, $\gamma^{Nat}_l, \forall l=0,\ldots,L$\\
 \textbf{Require:} $\textbf{x}^{*var}, \textbf{x}^{*hyper} $ a previous model optimal variational parameters and hyperparameters to initialize from if available.\\
 
   \eIf{$\textbf{x}^{*var}, \textbf{x}^{*hyper} $ available}{
   \textbf{Initialize} parameters: \\
    $\textbf{x}^{hyper}_0 \leftarrow \textbf{x}^{*hyper} $\\
    $\textbf{x}^{var}_0 \leftarrow \textbf{x}^{*var}$ \
   }{
   \textbf{Initialize} using another procedure (random initialization, PCA, \textit{ect.})}\
\noindent $ELBO_0 \leftarrow X, \textbf{y}, \textbf{x}^{hyper}_0, \textbf{x}^{var}_0$\\
 $t\leftarrow 0$\\
 \While{$t<iter$ \textbf{and} $\Delta_k>tol$}{
$\textbf{x}^{hyper}_{t+1} \leftarrow \text{Adam optimizer step} (ELBO_t,\textbf{x}^{hyper}_{t},\gamma^{Adam})$\\
\If{oversized natural step $\boldsymbol{\gamma}^{nat}$}
{$\gamma_j^{Nat}\leftarrow \gamma_j^{Nat}/10, \forall j=1,\ldots l-1$ }
$\textbf{x}^{var}_{t+1} \leftarrow \text{Nat Grad Optimizer step} (ELBO_t,\textbf{x}^{var}_{t},\gamma_0^{Nat},\dots,\gamma_l^{Nat})$

$\Delta_k \leftarrow \min\{ELBO_j,j=0,\ldots,t-k\}-\min\{ELBO_j,j=0,\ldots,t\}$
 $t\leftarrow t+1$
  }
 \Return $model(X,\textbf{y},\textbf{x}^{var}_{t-1}, \textbf{x}^{hyper}_{t-1}) $
 
 \caption{DGP model optimization}
 \label{algorithm1}
\end{algorithm}
%\begin{table}[t]
%\centering
%\caption{Comparison between the doubly stochastic variational inference approach and the variational auto-encoded method: the Root-mean-square error (RMSE) metric of two DGPs with the same architecture (1 hidden layer of 2 units and 17 induced inputs in each layer) but each trained with a different inference method. The function to approximate is the modified Xiong function with 17 training points and 100 test data points. 10 repetitions are performed. }
%\begin{tabular}{|K{2cm}|K{2cm}|K{2cm}|K{2cm}|K{2cm}|}
%  \hline
%  \textbf{Inference approach} & \multirow{ 2}{*}{\textbf{RMSE mean}}&\multirow{ 2}{*}{\textbf{RMSE std} }& \multirow{ 2}{*}{\textbf{ RMSE min}}  &\multirow{ 2}{*}{ \textbf{ RMSE max}}  \\
%  \hline
%  Doubly stochastic & \multirow{ 2}{*}{0.0184}  & \multirow{ 2}{*}{0.0075} & \multirow{ 2}{*}{0.0077} & \multirow{ 2}{*}{0.032} \\
%  	\hline
%Variational auto-encoded  & \multirow{ 2}{*}{0.0312}  & \multirow{ 2}{*}{0.019} & \multirow{ 2}{*}{0.0065} & \multirow{ 2}{*}{0.0666} \\
%	\hline
%
%\end{tabular}
%
%\label{training_table}
%\end{table}
%\begin{figure}[t]
%\centering
%\includegraphics[width=0.8\linewidth]{Bad_approximation_DGP_Dai.pdf}
%\caption{Exemple of a bad approximation of the Xiong-modified function by a DGP trained using the variational auto-encoded approach.}
%\label{patholigical_Dai}
%\end{figure}

\subsection{Architecture of the DGP}
The architecture of the DGP is a key question when using a DGP in BO. The configuration of the architecture of the DGP includes the number of layers, the number of units in each layer and the number of induced variables in each layer.

Increasing these architecture parameters enables a more powerful ability of representation. However, these variables are directly related to the computational complexity of the algorithm. Indeed, the computational complexity of a BO with DGP is given by $\mathcal{O} (j\times s \times t \times n \times (m_1^2d_1+\ldots+m_l^2d_l+\ldots+m_L^2d_{L}))$ where $j$ is the number of added points, $s$ the number of propagated samples, $t$ the number of optimization steps in the training, $n$ the size of the data-set, $l$ the number of layers, $m_l$ the number of induced inputs at the layer $l$ and $d_l$ the number of hidden units at the layer $l$. Moreover, the number of optimization steps $l$ increases according to the number of variables. Hence, the configuration of the DGP has to be adapted to the budget of computational time available and the complexity of the problem at hand (see Section~\ref{sec:3} for computational times). Usually, in the early iterations of BO there is not enough information to use complex model, therefore a standard GP may be sufficient. Then, along the iterations the number of layers is increased. 

It is interesting to observe that the number of inducing variables is the preponderant term in the complexity of the BO with DGP. Induced inputs were first introduced in the framework of sparse GPs. By choosing a number of induced inputs $m$ with $m<<n$ and $n$ the number of data points, the complexity of the inference becomes $O(nm^2)$ instead of $O(n^3)$. Hence, completing computational speed ups in the construction of the model. In sparse GP, increasing the number of inducing inputs allows more precision until reaching $m=n$ when the the full GP model is recovered. 
In DGPs, the interpretation of the induced inputs is more complicated. Firstly, it is essential to use induced inputs to obtain the evidence lower bound for the inference in DGP. Secondly, the variables $H_l, l=1\ldots L$  are random variables and not deterministic as $X$. So, it is possible to gain more precision even if $m_l>n$. 

\begin{figure}[t]%
\begin{minipage}[c]{0.5\linewidth}

\includegraphics[width=0.95\linewidth]{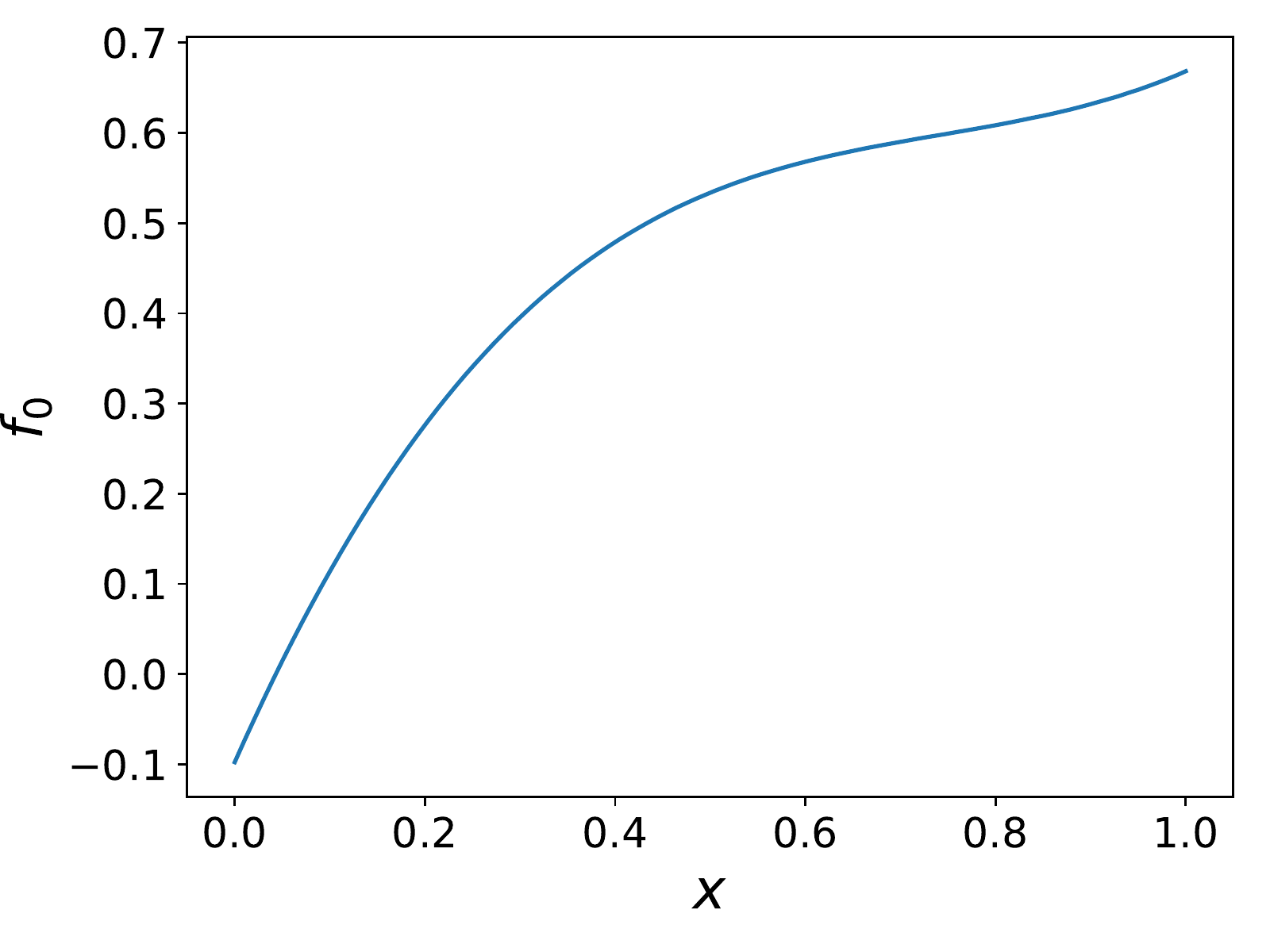}\\%\\
\center Layer 0

\end{minipage}
\hfill
\begin{minipage}[c]{0.5\linewidth}

\includegraphics[width=0.95\linewidth]{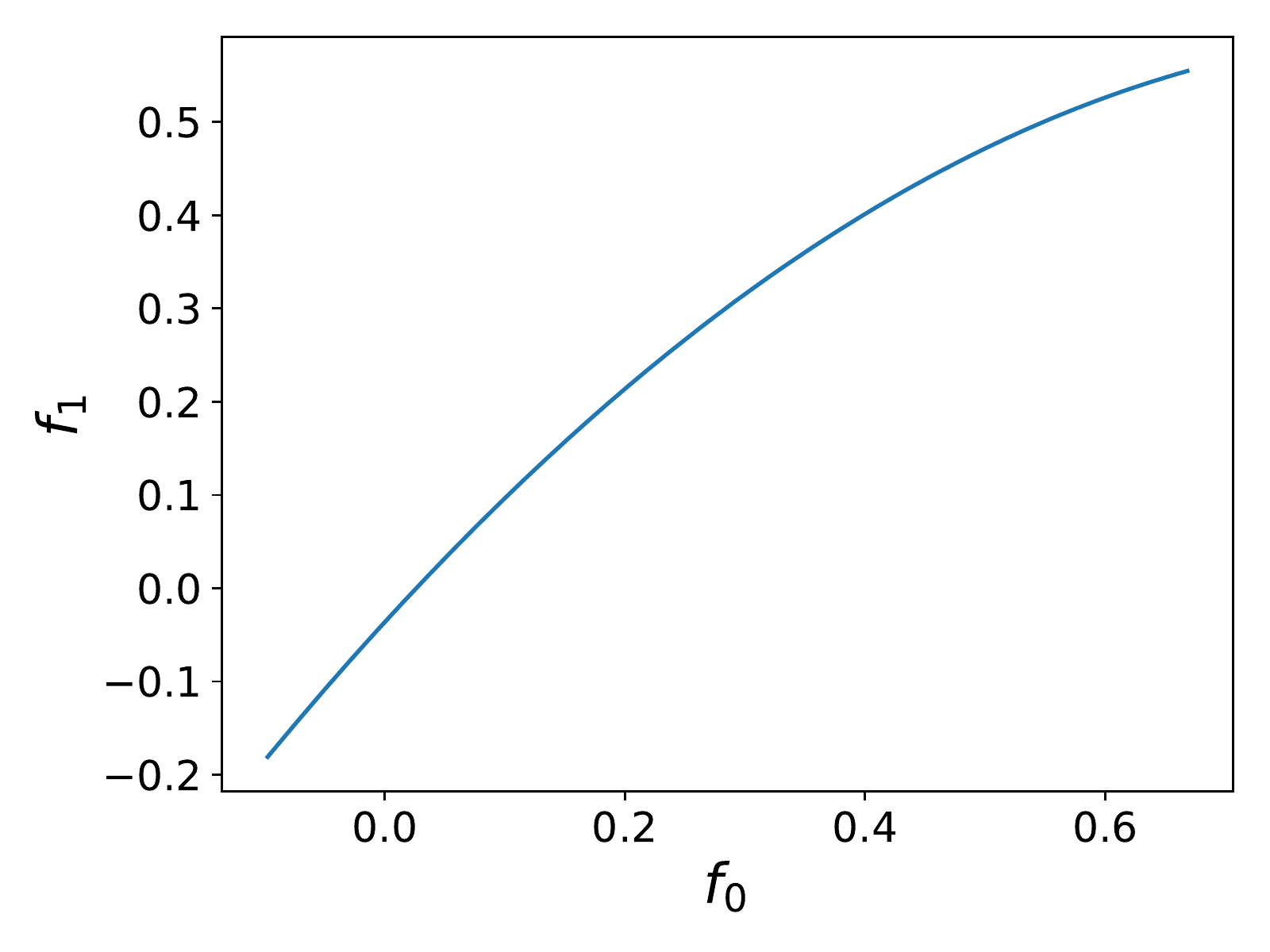}\\
\center Layer 1%
\end{minipage}
\begin{minipage}[c]{0.5\linewidth}

\includegraphics[width=0.95\linewidth]{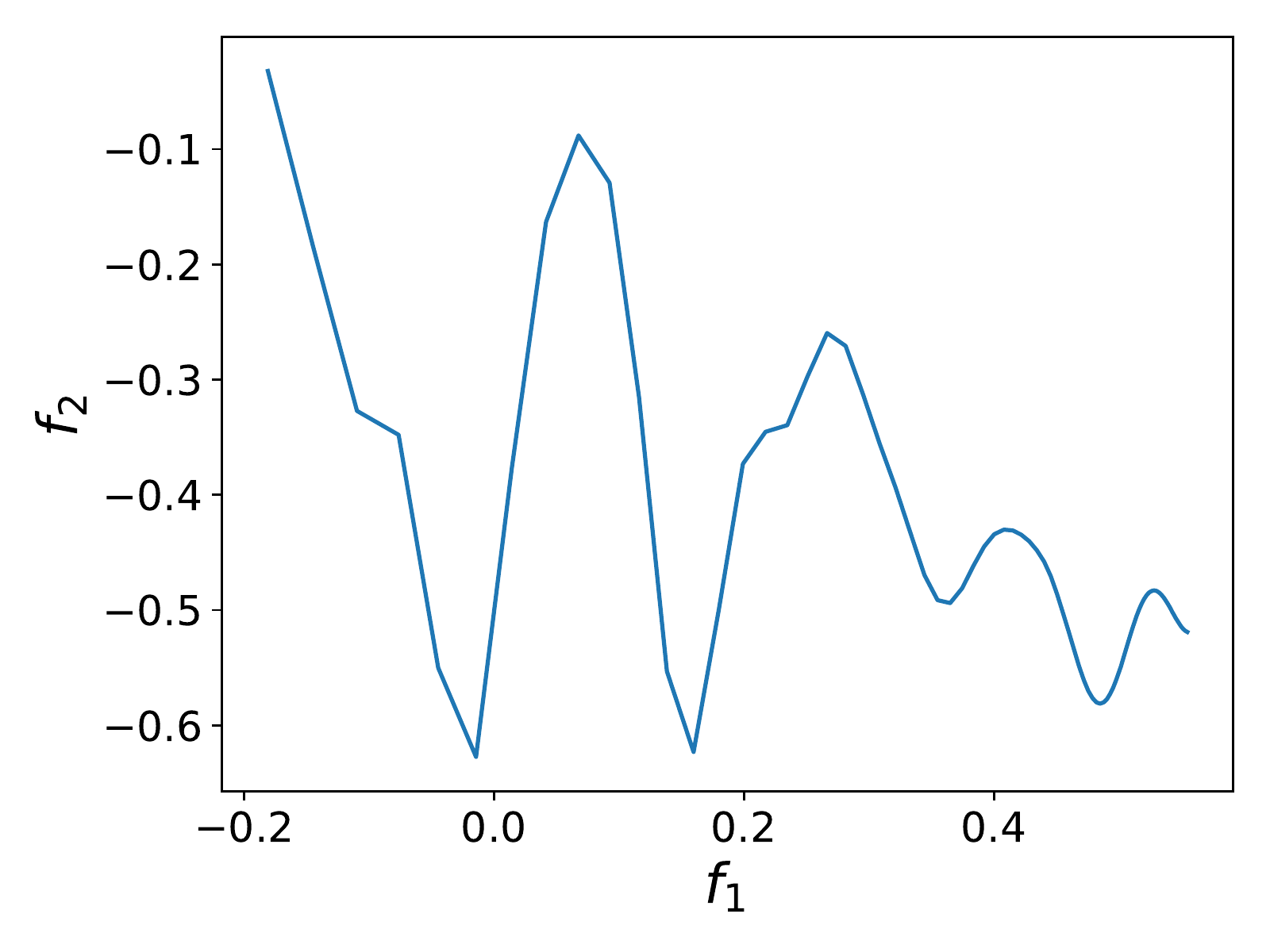}\\%\\
\center Layer 2

\end{minipage}
\hfill
\begin{minipage}[c]{0.5\linewidth}

\includegraphics[width=0.95\linewidth]{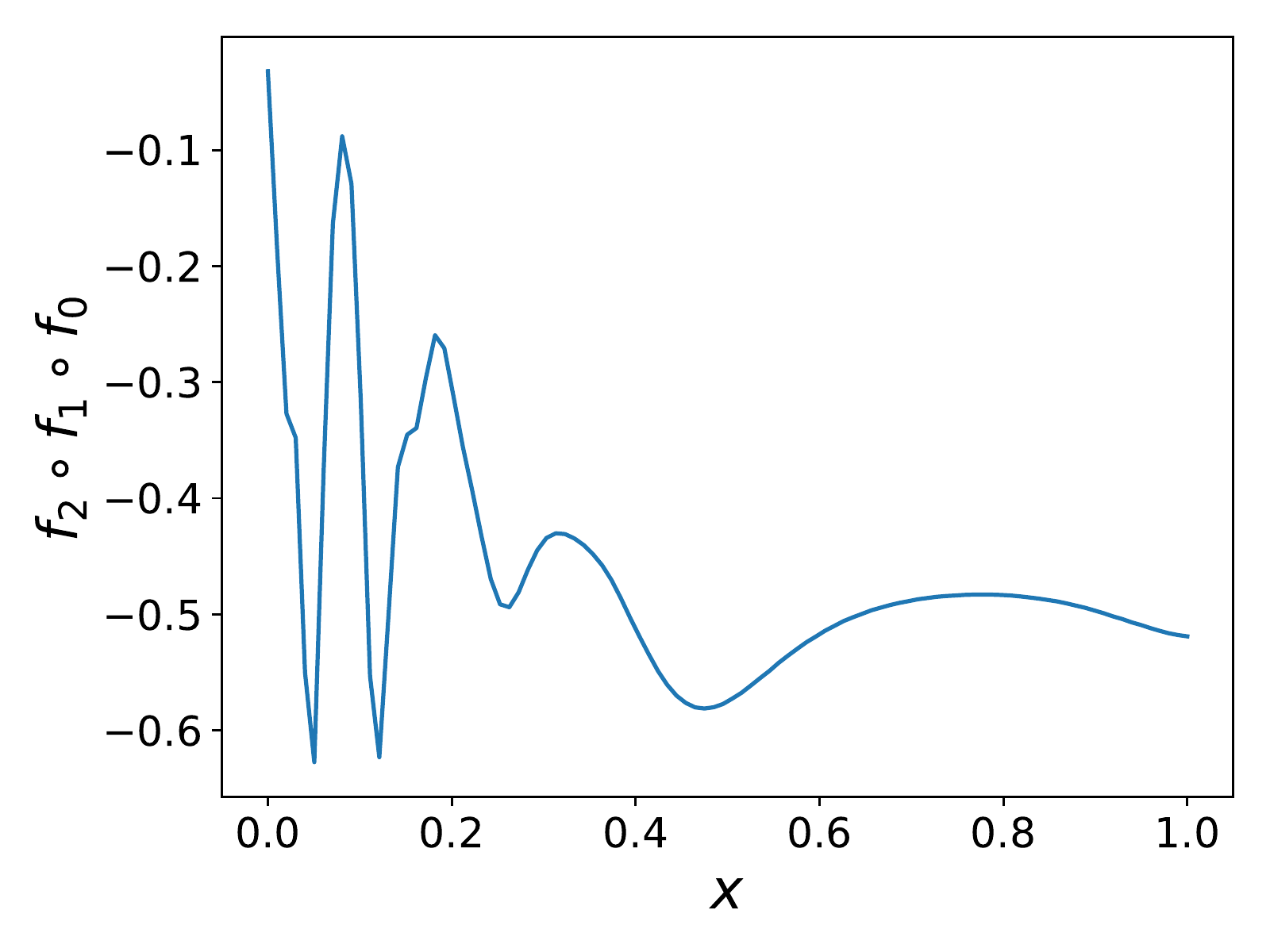}\\
\center All layers%
\end{minipage}

\caption{The input-output signal of each layer of a 2 hidden layers DGP used to approximate the modified Xiong function.}
\label{layers}
\end{figure}

However, the functional composition of GPs within a DGP makes each layer an approximation of a simpler function. In Figure~\ref{layers}, a 2-hidden layers DGP is used to approximate the modified Xiong function, the input-output of each layer is plotted. The intermediate layers try to deform the input space by stretching it, in order that the last layer approximates a stationary function, achieving an unparameterized mapping. Hence, the inner layers have a less complex behavior than the whole model. Thus, only a reduced number of induced inputs can capture the features of the hidden layers, hence, allowing computational speed ups. 

\subsection{Infill criteria}

\begin{figure}[t]%
\begin{minipage}[c]{0.5\linewidth}

\includegraphics[width=0.9\linewidth]{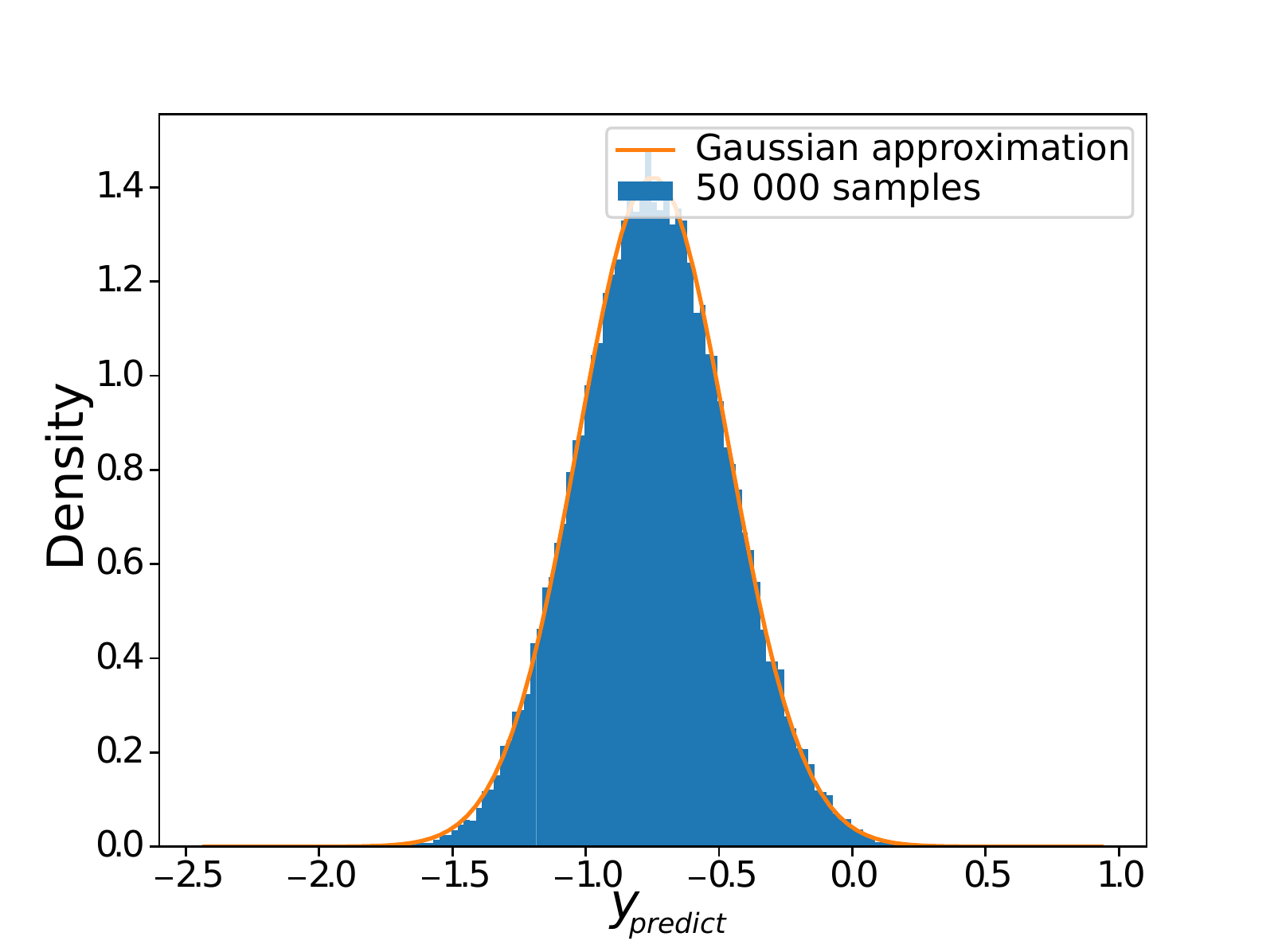}\\%\\

\end{minipage}
\hfill
\begin{minipage}[c]{0.5\linewidth}

\includegraphics[width=0.9\linewidth]{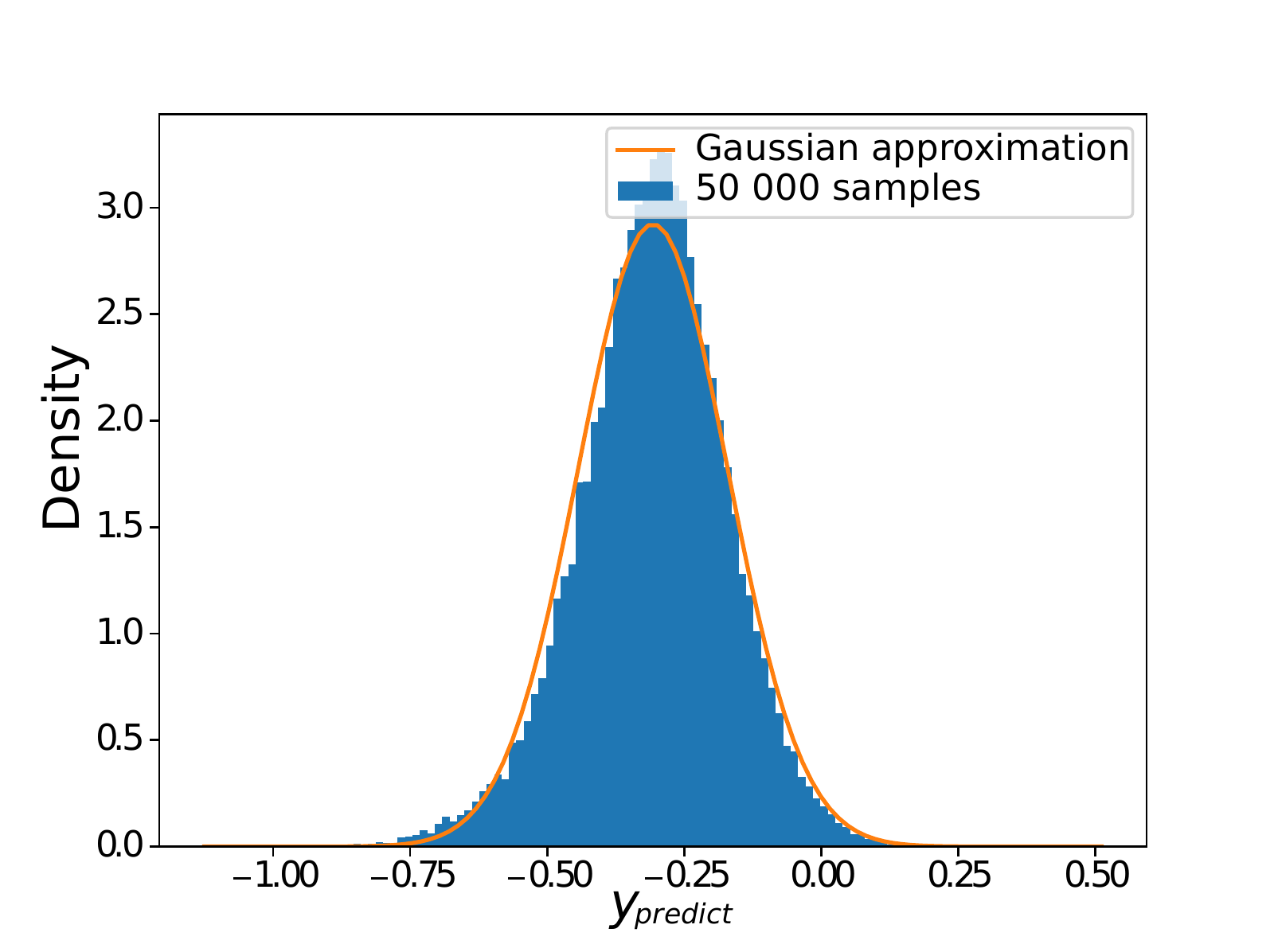}\\

\end{minipage}
\center
\begin{minipage}[c]{0.5\linewidth}

\includegraphics[width=0.9\linewidth]{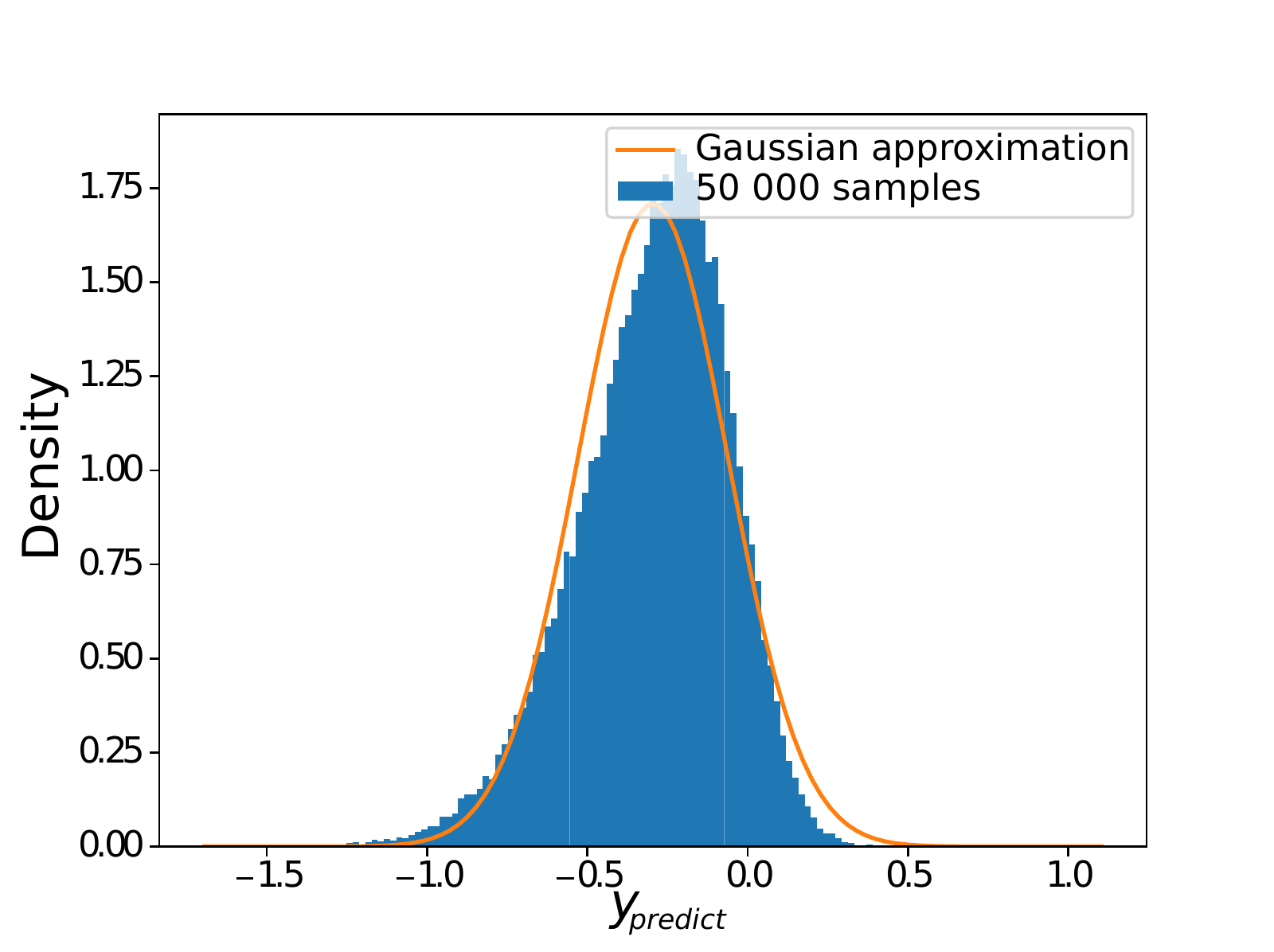}\\

\end{minipage}

\caption{Samples on the value of the prediction of a DGP model in three different locations. In the first two figures, the prediction is almost Gaussian. In the third figure the distribution of the prediction is slightly asymmetric, but it is still well approximated by a Gaussian distribution. }
\label{sampling}
\end{figure}

To use DGP in BO, it is essential to adapt the considered infill criteria to DGP. In fact, some infill criteria can not be used directly with DGP. For example, the EI formula (Eq.(\ref{eq:EI2})) is based on the fact that the prediction is Gaussian. However, in DGP the prediction does not \textit{a priori} follows a normal distribution. The expected improvement is the expected value of $I(\textbf{x})=max(0,y_{min}-f(\textbf{x}))$. Therefore, the direct approach is to use sampling techniques to approximate EI (Eq.(\ref{eq:EI1})). 
However, as observed in Figure~\ref{layers} the inner layers are often simple functions, almost linear, with a last layer that approximates a deformed stationary function. This allows the prediction from the composition of GPs to be reasonably considered as Gaussian (see Figure~\ref{sampling}). Hence, to predict using DGPs a Gaussian approximation can be made, in order to directly use the analytical formula of the infill criteria used for GPs. Hence, sampling directly on the value of the improvement is avoided.

Infill criteria such as EI are highly multi-modal, especially in high-dimensional problems. For this reason, an evolutionary algorithm such as the Differential Evolution algorithm \cite{4632146} is preferred. The DGP allows parallel prediction which makes it possible to evaluate the infill criteria for all the population simultaneously. The result obtained using the evolutionary algorithm can then be optimized by a local optimizer. This hybridization is preferred to the use of multiple local searches whose number increases exponentially with the dimension of the problem.

%Salimbeni \textit{et al.} approximated the prediction of a DGP as a Gaussian mixture by considering the mixture of the $S$ Gaussian distribution obtained at the final layer by propagating $S$ samples. 

\subsection{Summary of DEGO}
To summarize the proposed BO \& DGP algorithm, DEGO, Algorithm~\ref{algorithm2} describes the steps previously discussed. The Expected Improvement is used as the infill criterion, but other infill criteria may be used. If approximation of the DGP prediction by a Gaussian is not valid, sampling is used to approximate the infill criterion. Some empirical rules can be used to determine the number of initial points and the number of added points depending on the dimension of the problem $d$ (for the experimentations in Section~\ref{sec:3}, for all the problems an initial DoE of size $5 \times d$ is considered and $10 \times d$ points are added in the BO process.). The size of the induced variables is fixed along all the iterations to the total number of points at the end of the BO. This allows the models to keep the same number of parameters along the iterations, making it possible to initialize them from the previous models. Moreover, as discussed previously, setting the number of induced variables to a number larger than the number of points in DGP may allow a better representation. The model is updated from the previous model optimal parameter values for a certain number of consecutive iterations allowing speed ups in the optimization, and then initialized from scratch every $n_{update}$ iterations to avoid being tricked in some bad local minima.\\
In this algorithm the unconstrained case is considered. However, the generalization to the constrained case is straightforward, since it comes back to create DGP models also for the constraints and to use a constrained infill criterion as the Probability of Feasibility or the Expected Violation.

\begin{algorithm}[]
  \textbf{Require}: {Expensive black-box problem of dimension $d$ to optimize, $f^{exact}$}\\
  \textbf{Require}: Number of initial points $n$.\\
   \textbf{Require}: Number of total added points $n_{add}$.\\
  \textbf{Require}: Number of layers $l$. \\
  \textbf{Require}: Number of iterations in the optimization of the model $iter$.\\
  \textbf{Require}: Number of consecutive model updates using the previous model optimal values $n_{update}$.\\
  
 $X_0\leftarrow LHS(d,n)$ (or another design of experiments method)\\
 $\textbf{y}_0 \leftarrow f^{exact}(X_0)$ (\textbf{evaluate})\\
  $m \leftarrow n+n_{add}$ (set the number of induced variables to the final number of points) \\
 $t \leftarrow 0$ \\
 $model_0 \leftarrow \text{\textbf{DGP model optimization}}(X_0,\textbf{y}_0,m,l,iter)$ (optimize model from scratch)\\
 \While {$t\leq n_{add}$}{
  $t \leftarrow t+1$ \\
 $\textbf{x}^{(t)}\leftarrow argmax(EI_{model_{t-1}}(\textbf{x}))$(use sampling to estimate the $EI$)
 \\
 $y^{(t)}\leftarrow f^{exact}(\textbf{x}^{(t)})$ (\textbf{evaluate})\\
 $X_{t} \leftarrow \begin{bmatrix} X_{t-1}\\ \textbf{x}_t \end{bmatrix} $ (add a row to the matrix)\\
 $\textbf{y}_{t} \leftarrow \begin{bmatrix} \textbf{y}_{t-1}\\ y^{(t)} \end{bmatrix}$ (add an element to the vector)\\
  \eIf{$ t\%n_{update} \neq 0 $}{
   $model_t \leftarrow \text{\textbf{DGP model training}}(X_t,\textbf{y}_t,m,l,iter, model_{t-1})$ (optimize model using the optimal parameter values of the previous model as initialization)\
   }{
   $model_t \leftarrow \text{\textbf{DGP model training}}(X_t,\textbf{y}_t,m,l,iter)$ (optimize model from scratch)
  }
 
 }
  \Return{$X_t,\textbf{y}_t$}
 \caption{ DEGO algorithm}
 \label{algorithm2}
\end{algorithm}

\clearpage

\section{Experimentations}
\label{sec:3}
\subsection{Analytical test problems}
Experimentations on different analytical problems have been performed to assess the performance of DGPs in BO. Details on the experimental setup are presented in Appendix~\ref{appendixB}.
\subsubsection{2d constrained problem}

\begin{figure}[h]%
\center
\includegraphics[width=0.8\linewidth]{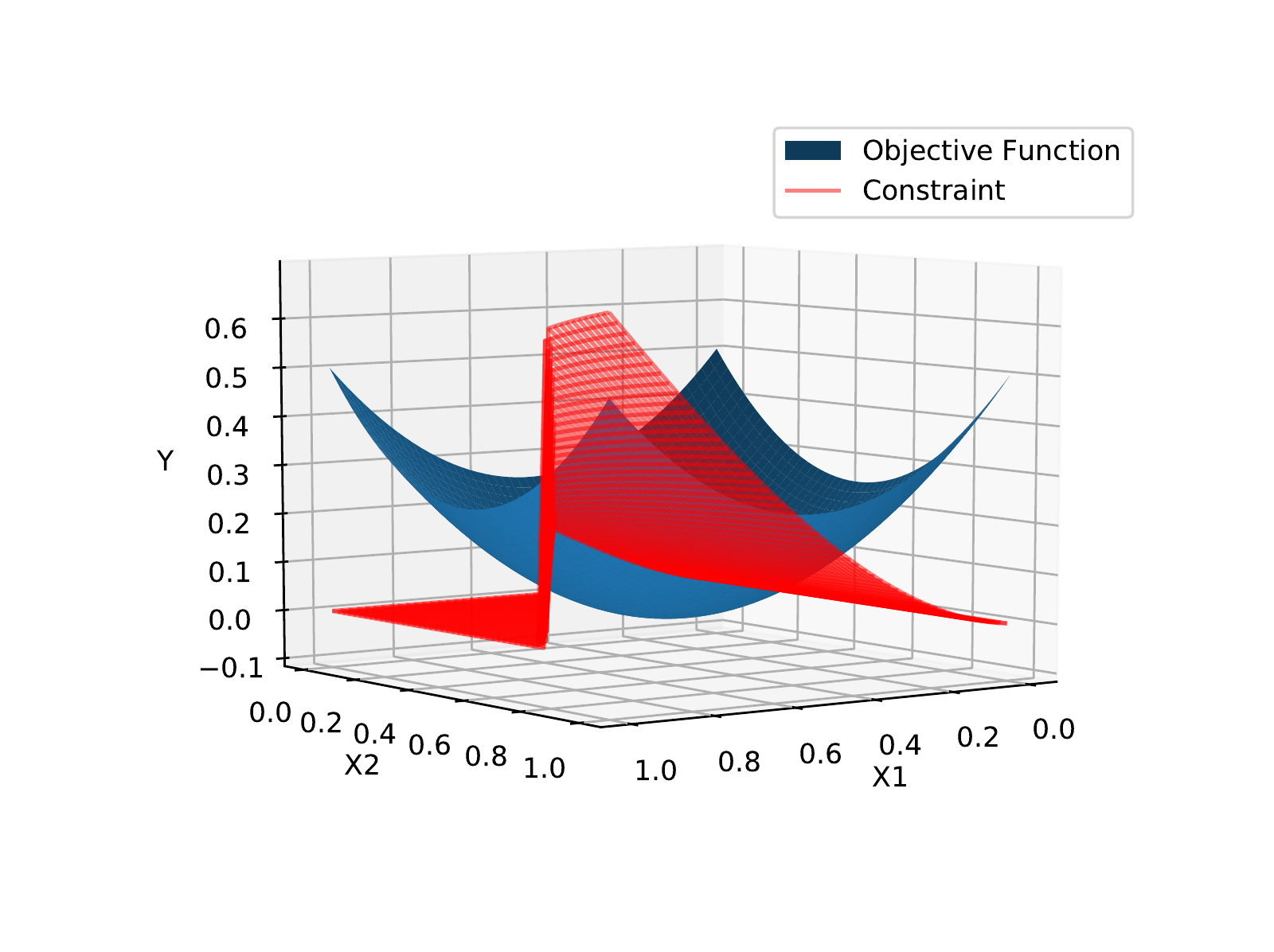}
\vspace{-1.8cm}
\caption{Objective and constraint functions 2d problem}
\label{exp2}
\end{figure}

The function to optimize is a simple two dimensional quadratic function. While the constraint is non-stationary and feasible when equal to zero. An important discontinuity between the feasible and non feasible regions breaks the smoothness of the constraint (Figure~\ref{exp2}). Therefore, the problem is challenging for standard GP, since the optimal region is exactly at the boundary of the discontinuity, requiring an accurate modeling of the non-stationarity.

A DoE of 10 initial data points is initialized using a Latin Hypercube Sampling. Then, 20 points are added using the Expected Violation criterion (EV) to handle the constraint. A standard Gaussian Process with an RBF kernel is used to approximate the objective function. The DGPs are used with an RBF kernel in each layer and are optimized using 5000 optimization steps. To assess the robustness of the algorithms, 50 repetitions are performed.

\begin{figure}
\center
\includegraphics[width=1\linewidth]{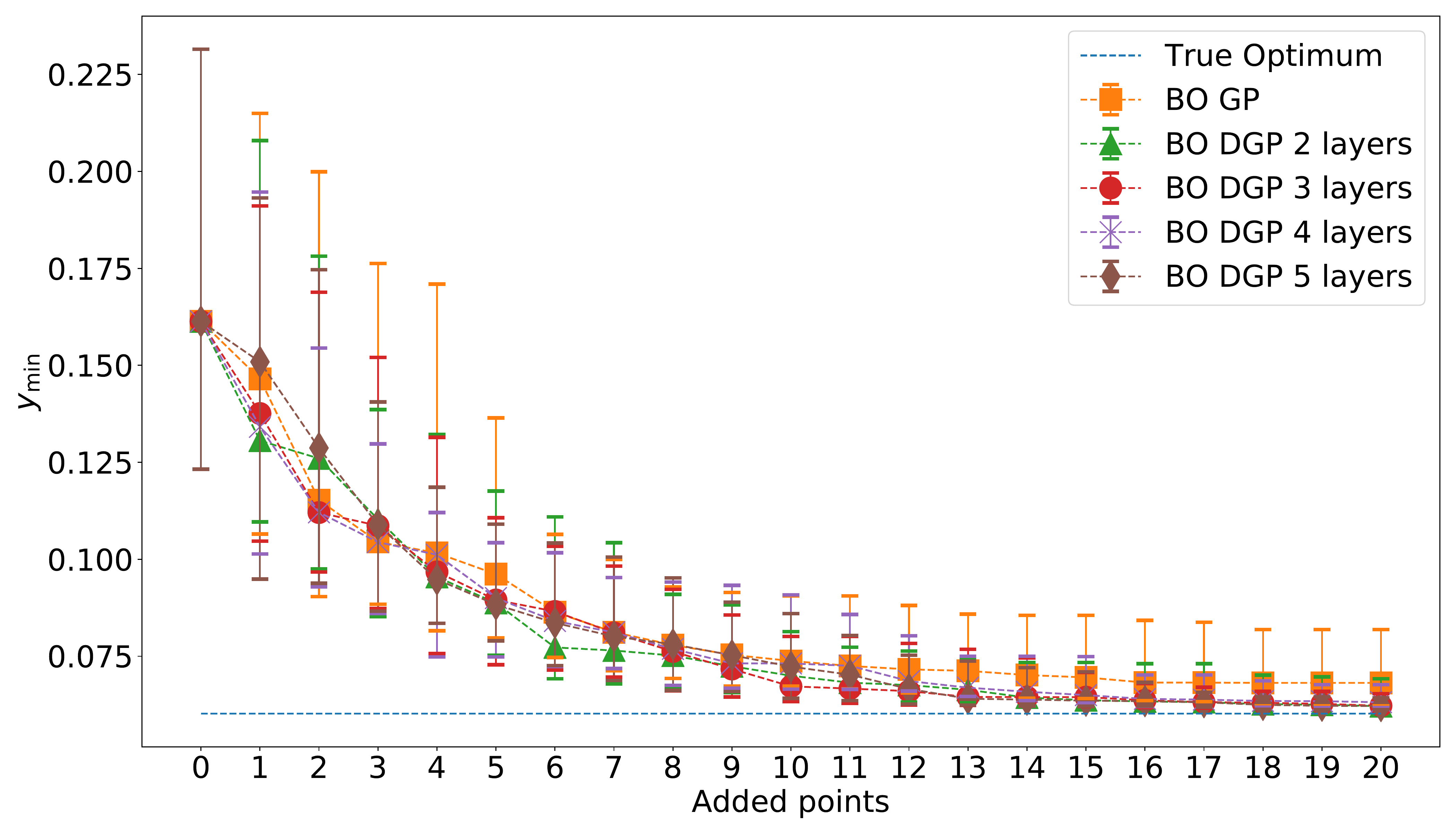}
\caption{Plot of convergence of BO using different architectures of DGPs with 5000 training steps for 50 different initial DoE and a standard GP. The markers indicate the median of the minimum obtained while the error-bars indicate the first and third quartiles.}
\label{conv_ns2D}
\end{figure}

\begin{figure}
\center
\includegraphics[width=0.9\linewidth]{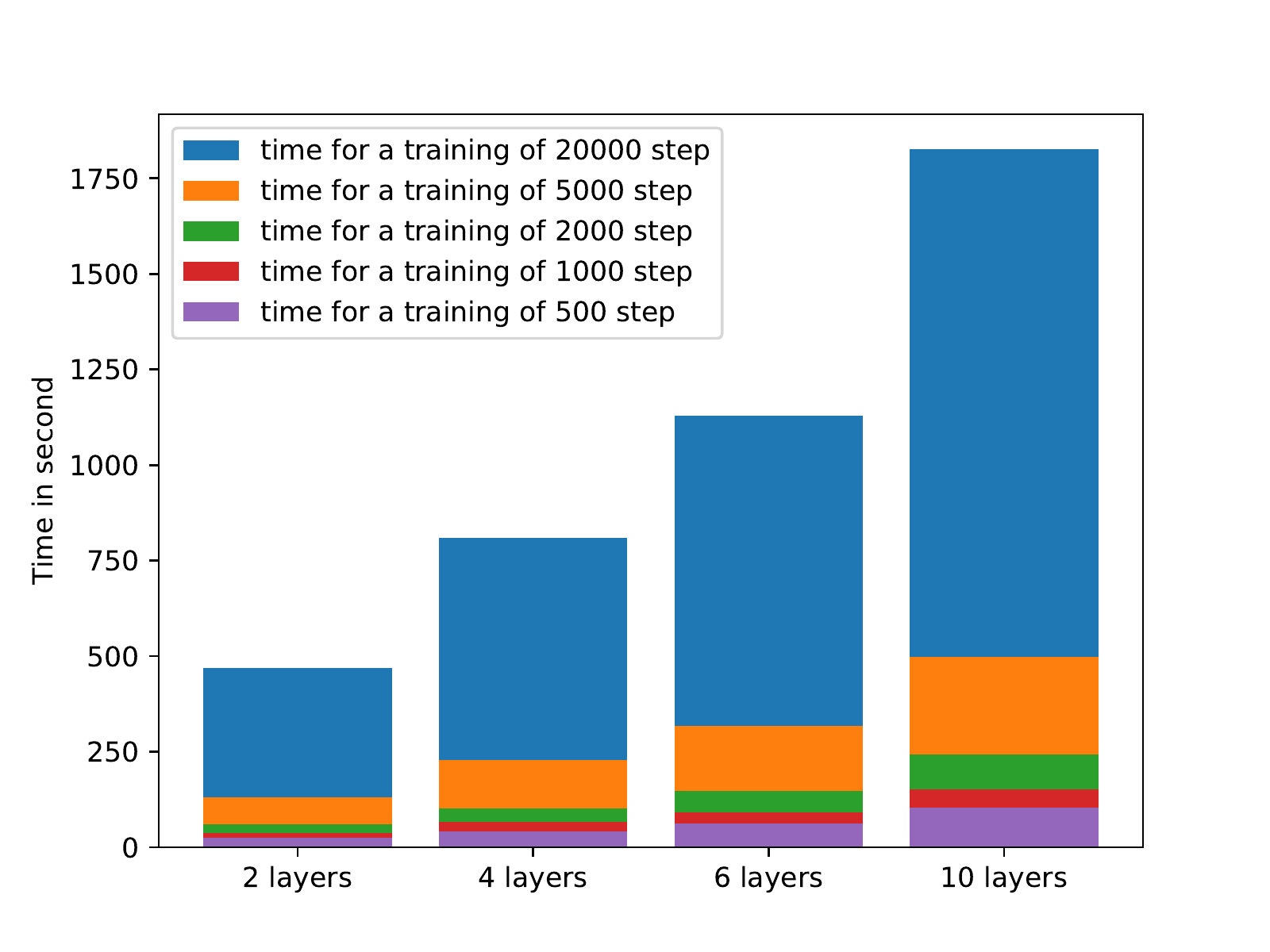}
\caption{Mean time in one iteration of BO according to the number of layers in DGP and the number of steps in the training of the model.}
\label{timens2D} 
\end{figure}

\begin{figure}
\begin{minipage}[c]{0.5\linewidth}
\includegraphics[width=1\linewidth]{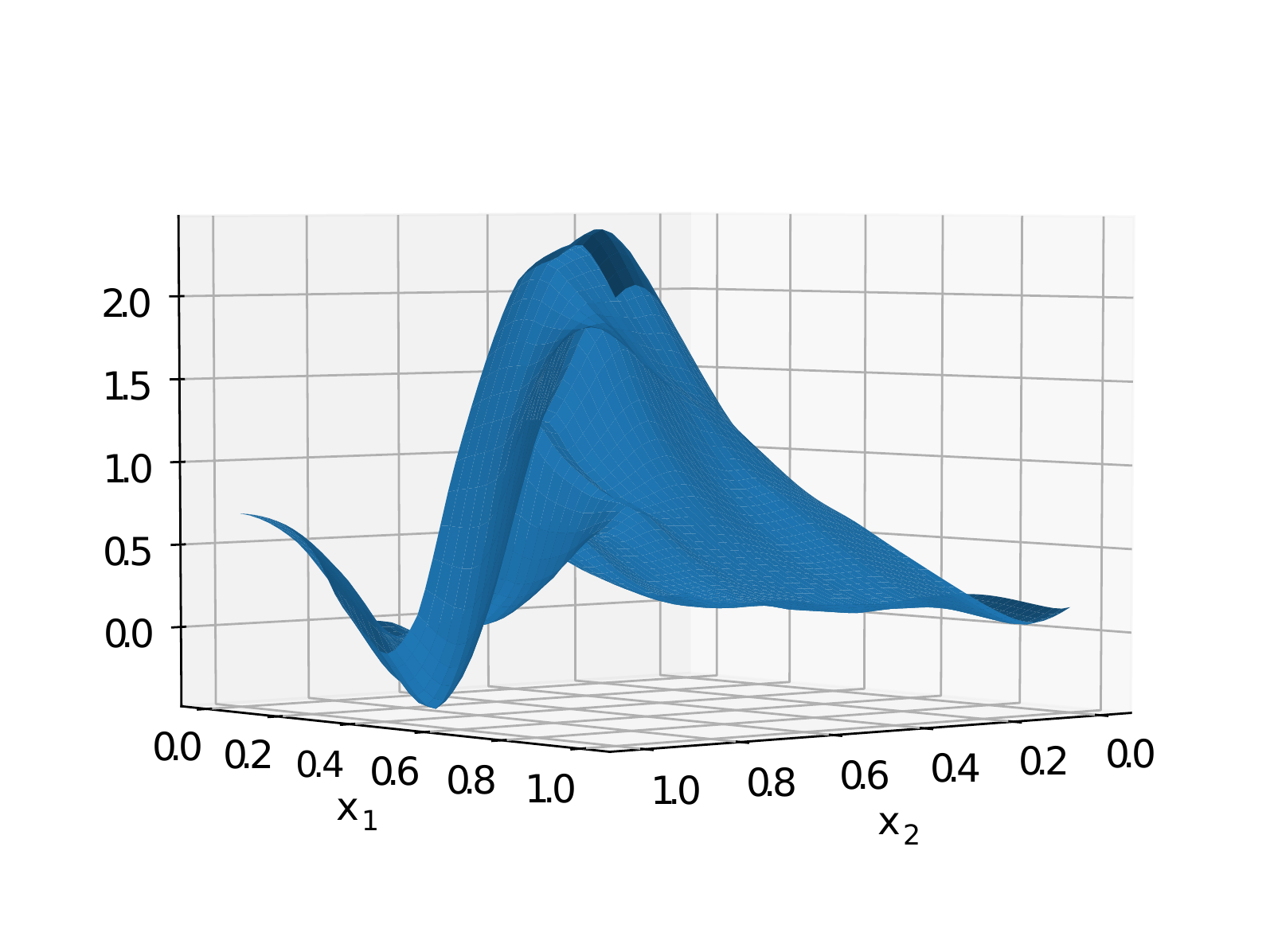}
\caption{GP approximation of the constraint with a DoE of 40 points.}
\label{GPapprox}
\end{minipage}
\hfill
\begin{minipage}[c]{0.5\linewidth}
\includegraphics[width=1\linewidth]{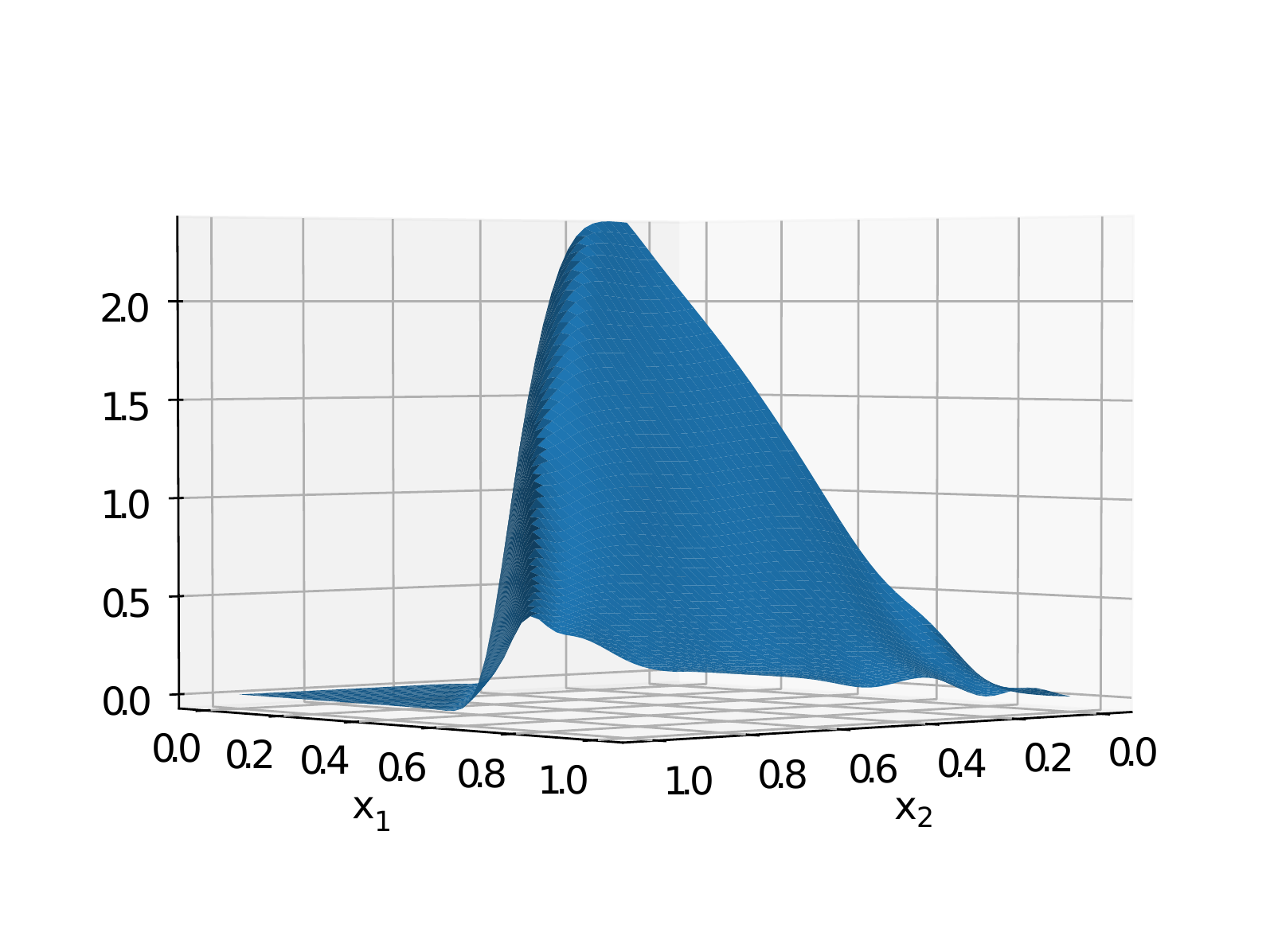}
\caption{DGP 2 layers approximation of the constraint with a DoE of 40 points.}
\label{DGPapprox}
\end{minipage}
\end{figure}

The convergence plots of the BO algorithms with GP, DGP 2, 3, 4 and 5 layers are displayed in Figure~\ref{conv_ns2D}. As expected, the BO \& GP is not well-suited for this problem, actually at the end of the algorithm, the median is still far from the actual minimum and there is an important variation. This is due to the fact that the GP can not capture the discontinuity and the feasible tray region of the constraint and considers an important region as unfeasible (Figure~\ref{GPapprox}). However, DEGO is able to capture the feasible region (Figure~\ref{DGPapprox}), which makes it able to give efficient results with a median at the end of the algorithms near to the actual minimum and better robustness. The interesting observation concerns the mean and standard deviation given in Table~\ref{compare_NS2C}, where the 3 layers DGP gives the best results. Increasing the number of layers deteriorates the quality of the results. This is explained by the fact that 5000 steps in the optimization for DGPs with more than three layers in this case is insufficient. Hence, the necessity to increase the number of optimization steps in the training of deeper models. However, increasing the number of layers and the number of steps induces additional computational time (Fig~\ref{timens2D}) which quickly becomes an important burden for high dimensional problems. For the remaining test cases only a DGP with two layers is considered.

\begin{table}
\center
\caption{Performance of BO (values of the minimum found) with standard GP and different DGP configurations on the constrained 2d problem. 50 repetitions are performed.}
\begin{tabular}{|>{\centering\arraybackslash} m{3cm}| >{\centering\arraybackslash} m{3cm}|>{\centering\arraybackslash} m{3cm}|>{\centering\arraybackslash} m{3cm}|}
  \hline
  \textbf{BO \&} & \textbf{mean minimum obtained}& {\textbf{standard deviation}} & \textbf{ \hspace{3cm}       gap between the mean minimum and the global minimum.}   \\[0.5cm]
  \hline
  GP & 0.09356 & 0.0605 & \hspace{3cm}{0.03336} \\[0.5cm]
   \hline
	DGP 2 L & 0.08468 & 0.059793 & \hspace{3cm}{0.02448}  \\[0.5cm]
	\hline
	\textbf{DGP 3 L }  & \textbf{0.073918} & \textbf{0.04293} & \hspace{3cm}{\textbf{0.01371}} \\[0.5cm]
	\hline
	DGP 4 L  & 0.08066 & 0.05073 & \hspace{3cm}{0.02046} \\[0.5cm]
	\hline
	DGP 5 L & 0.08204 & 0.05707 & \hspace{3cm}{0.02184}\\[0.5cm]
	\hline Global minimum & 0.0602 & - &  \multirow{ 2}{*}{-} \\[0.5cm]
\hline

\end{tabular}
\label{compare_NS2C}
\end{table}

\subsubsection{Trid function}

The Trid function is considered in 10 dimensions Eq.(\ref{trid_eq}). The range of variation of the $10d$ Trid function values is large. It varies from values of $10^5$ to its global minimum $f(\textbf{x}^*)=-210$ (Figure~\ref{Trid}). This large variation range makes it difficult for BO with stationary GP to find the global minimum.
This function was also used in \cite{toal2012non} to assess the performance of BO with non-stationary GP using the non-linear mapping and a mixture of the non-linear mapping and standard GP called Adaptive partial non-stationary kriging. 

The results of BO with a DGP of 2 hidden layers are compared to the results found in \cite{toal2012non} on 50 different DoE (Table ~\ref{compare_Trid}). The initial DoE is initialized with a Latin Hypercube Sampling with 50 initial points, and 100 points are added during the BO using the EI criterion.

%\begin{figure}[]%
%\includegraphics[width=0.9\linewidth]{trid.pdf}\\%\\
%\caption{Sectional $2d$ view of the Trid function showing where the global minimum lie}
%\label{Trid}
%%\hfill
%%\begin{minipage}[c]{0.45\linewidth}
%%\includegraphics[width=0.9\linewidth]{tridlog10.pdf}\\
%%\caption{Sectional $2d$ view of the log 10 Trid function showing its non-stationary behavior around the global minimum}
%%\label{logTrid}
%%\end{minipage}
%\end{figure}

\begin{figure}
\begin{minipage}[c]{0.5\linewidth}
\includegraphics[width=1\linewidth]{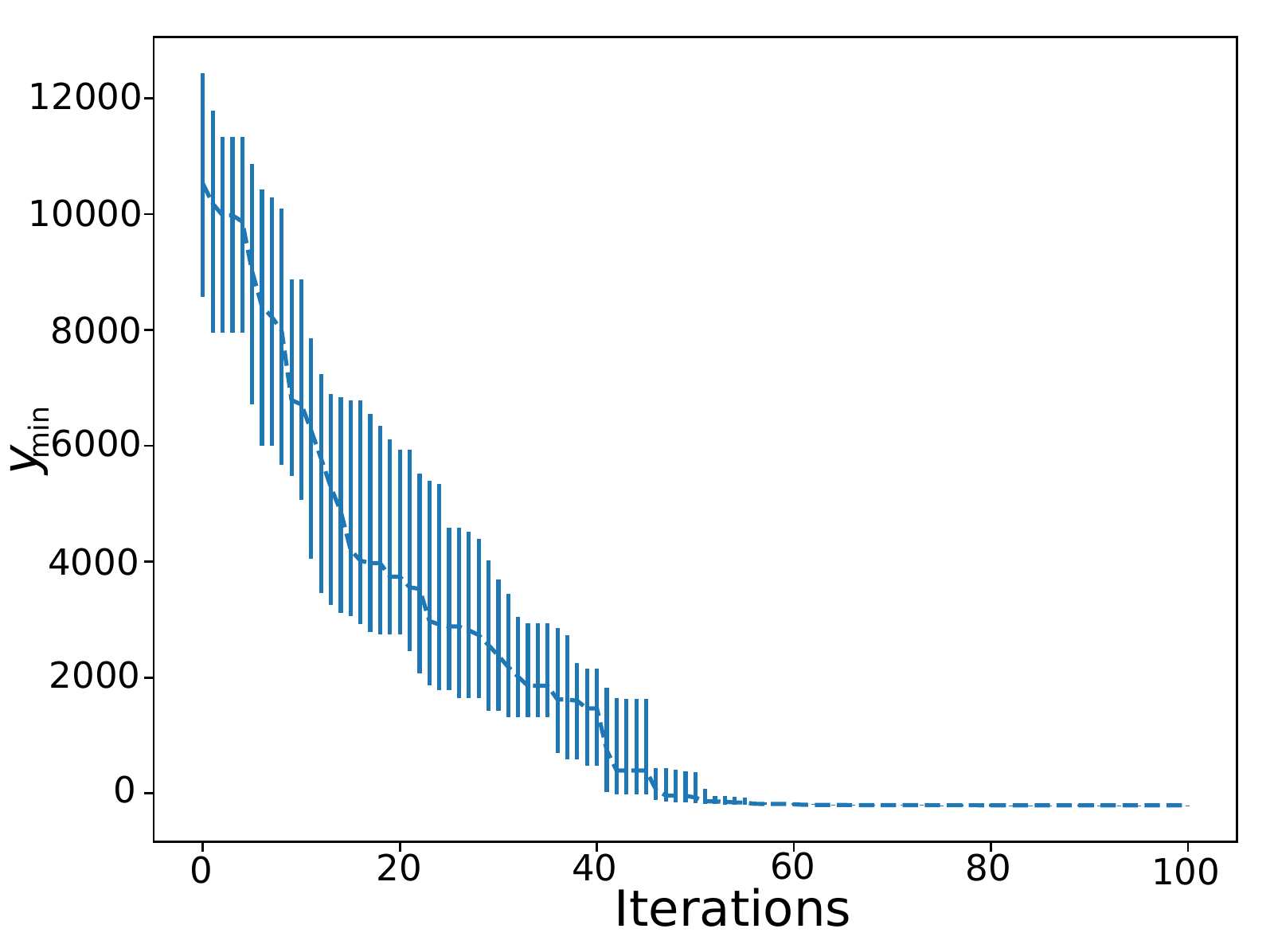}\\
\end{minipage}
\hfill
\begin{minipage}[c]{0.5\linewidth}
\includegraphics[width=1\linewidth]{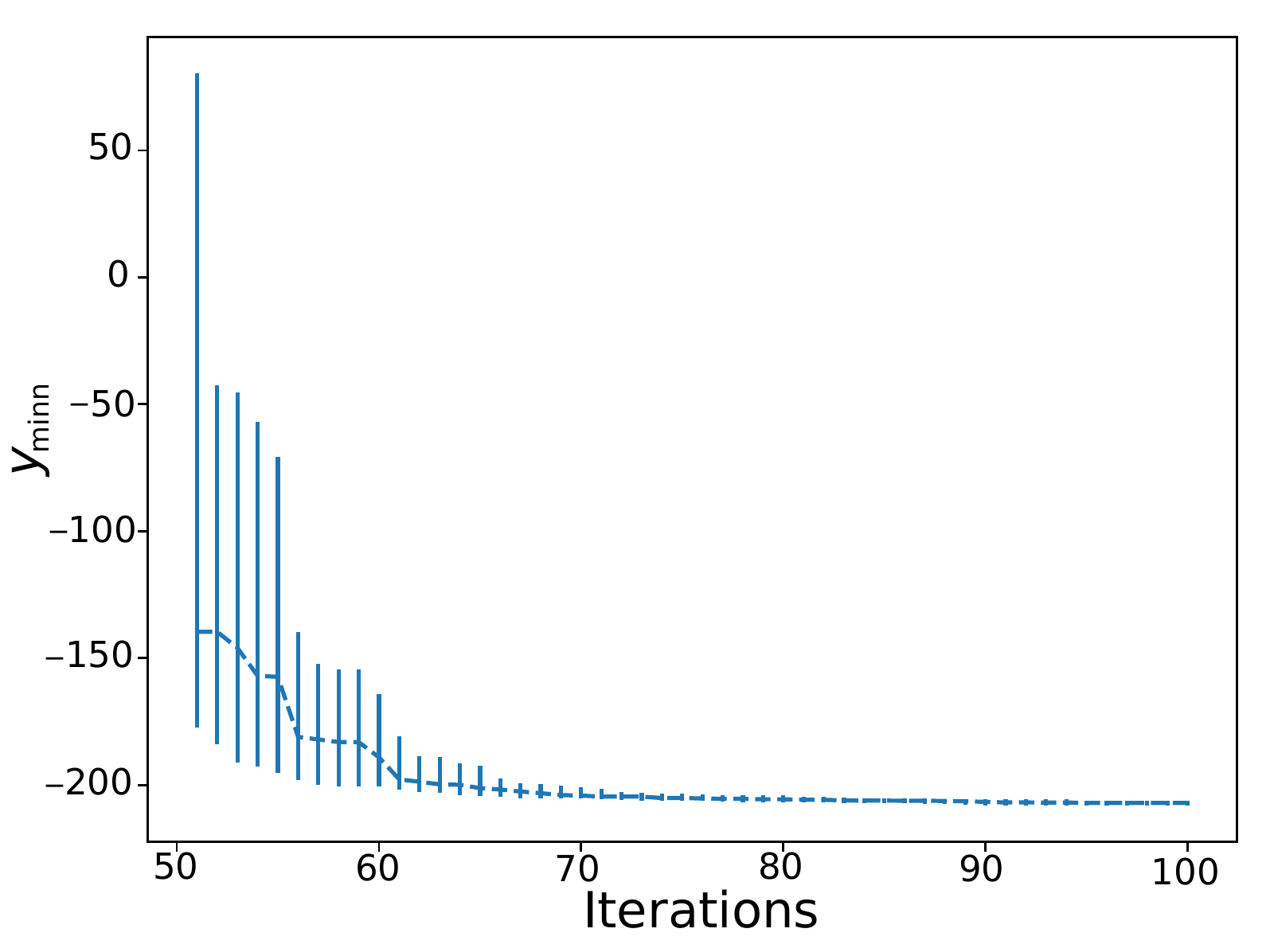}\\
\end{minipage}
\caption{DEGO plot of convergence of $y_{min}$ on the Trid function with 50 DoE. The curve represent the median of $y_{min}$ while the vertical bars represent the first and third quartiles. (left) Convergence plot on all the iterations. (right) Convergence plot on the 50 last iterations. }
\label{tnk_convergenceplot}
\end{figure}

\begin{table}
\centering	
\caption{Performance of BO (values of the minimum found) with standard GP, non-linear mapping with two knots (NLM), Adaptive partial non-stationary kriging (APNS), Deep Gaussian Processes with two hidden layers (DGP) on the Trid function.}

\begin{tabular}{|>{\centering\arraybackslash} m{3cm}| >{\centering\arraybackslash} m{3cm}|>{\centering\arraybackslash} m{3cm}|>{\centering\arraybackslash} m{3cm}|}
  \hline
  \textbf{BO \&} & \textbf{mean minimum obtained}& \textbf{standrad deviation}  & \textbf{ \hspace{3cm}       gap between the mean minimum and the global minimum.}    \\[0.5cm]
  \hline
  GP &{-20.730} & 75.654 & \hspace{3cm}{189.27}  \\[0.5cm]
   \hline
	NLM & -57.727 & 59.920 & \hspace{3cm}{152.273}  \\[0.5cm]
	\hline
	APNS  & -49.112& 62.746 & \hspace{3cm}{160.888} \\[0.5cm]
	\hline
	\textbf{DGP} (DEGO) &\textbf{-206.739} & {\textbf{1.5521}}& \hspace{3cm}{\textbf{3.261}} \\[0.5cm]
	\hline
	Global minimum & -210 & - & \multirow{ 2}{*}{-} \\[0.5cm]
\hline

\end{tabular}
\label{compare_Trid}
\end{table}
The minimum given by BO \& GP, NLM (Non-Linear Mapping) and APNS (Adaptive Partial Non-Stationary kriging) for this problem are not close to the global minimum. Moreover, there is a high variation in the obtained results, showing the difficulty of these approaches to handle this optimization problem. However, there is an important difference between DEGO (BO \& DGP) compared to the other approaches. The approximated mean minimum obtained $-206.739$ is very close to the actual global minimum $-210$ with a low standard deviation of $1.5521$, hence confirming the robustness of the approach. The convergence plot of DEGO (Figure~\ref{tnk_convergenceplot}) also shows the fast convergence of this approach. In fact, after only $65$ iterations the algorithm is stabilized around the global optimum.

\subsubsection{Hartmann-6d function}
The Hartmann-6d is a $6d$ function (Eq.(\ref{heart_eq})). Which is smooth and does not show non-stationary behavior (Figure~\ref{slice_hartmann}). The interest of this function is that BO coupled with some non-stationary approaches can not reach its global minima while BO \& classic GP presents good performance on it. Hence, using DEGO on this function allows to demonstrate the robustness of this non-stationary BO algorithm on stationary functions. This makes sense when applying BO to black-box functions when there is no information about the stationarity of the problem at hand. 
%$$f(\textbf{x})= \sum_{i=1}^4 \alpha_i \exp\left(-\sum_{j=1}^6 A_{ij} (x_j-Pij)^2 \right), \text{  } x_i\in[0,1], \forall i=1,\dots,6$$
%with: 
%$$\alpha=[1,1.2,3,3.2]^\top$$
%and
%$$P=10^{-4}\begin{bmatrix} 1312 & 1696 & 5569 & 124 & 8283 & 5886\\2329 & 4135 & 8307&3736&1004&9991\\2348&1451&3522&2883&3047&6650\\4047&8828&8732&5743&1091&381\end{bmatrix}$$ 
%and $$A= \begin{bmatrix} 10&3&17&3.5&1.7&8\\0.05&10&17&0.1&8&14\\3&3.5&1.7&10&17&8\\17&8&0.05&10&0.1&14 \end{bmatrix}$$

%\begin{figure}
%\center
%\includegraphics[width=0.9\linewidth]{hartmann_slice.pdf}\\
%\caption{Sectional $2d$ view of the Hartmann-6d function showing where the global minimum lie}
%\label{slice_hartmann}
%\end{figure}

The results of BO with a DGP of 2 hidden layers are compared to the results found in \cite{toal2012non} on 50 different DoE (Table~\ref{compare_heart}). The initial DoE are initialized using a Latin Hypercube Sampling with 30 initial points and 60 points are added during the BO process using the EI criterion. 

\begin{figure}
\center
\includegraphics[width=0.65\linewidth]{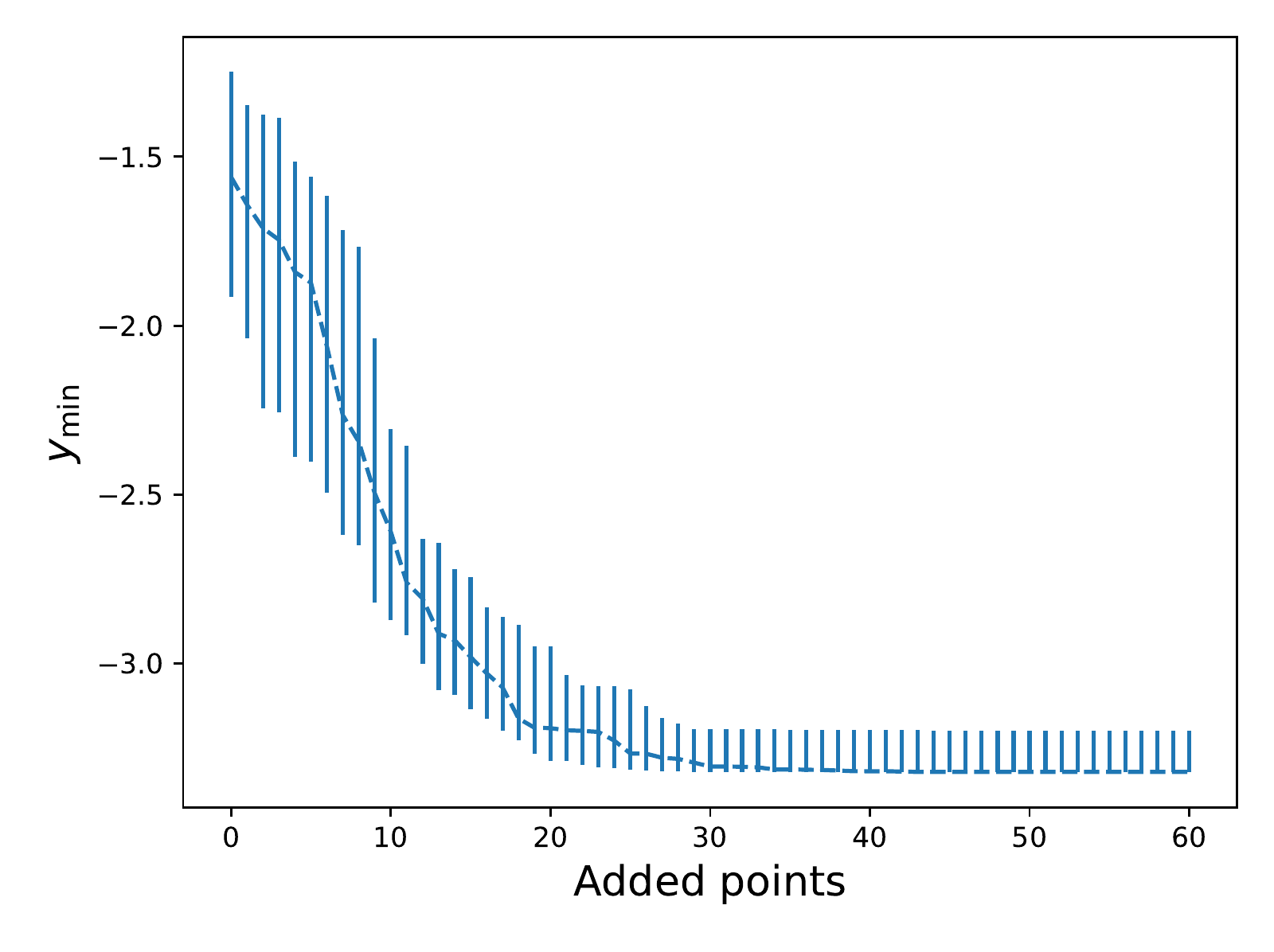}\\
\caption{DEGO plot of convergence of $y_{min}$ on the Hartmann-6d function with 50 DoE. The curve represent the median of $y_{min}$ while the vertical bars represent the first and third quartiles. }
\label{hart_conv}
\end{figure}

\begin{table}
\centering	
\caption{Performance of BO (values of the minimum found) with standard GP, non-linear mapping with two knots (NLM), Adaptive partial non-stationary kriging (APNS), and Deep Gaussian Processes with two hidden layers (DGP) on the Hartmann 6d function.}

\begin{tabular}{|>{\centering\arraybackslash} m{3cm}| >{\centering\arraybackslash} m{3cm}|>{\centering\arraybackslash} m{3cm}|>{\centering\arraybackslash} m{3cm}|}
  \hline
  \textbf{BO \&} & \textbf{mean minimum obtained}& \multirow{ 2}{*}{\textbf{standard deviation}} & \textbf{ \hspace{3cm}       gap between the mean minimum and the global minimum.}   \\[0.5cm]
  \hline
  GP &{-3.148} & {0.275} & \multirow{ 2}{*}{0.174} \\[0.5cm]
   \hline
	NLM & -2.818 & {0.570} & \multirow{ 2}{*}{0.504}  \\[0.5cm]
	\hline
	APNS  & -3.051 & {0.415} & \multirow{ 2}{*}{0.271} \\[0.5cm]
	\hline
	\textbf{DGP} (DEGO) &\textbf{-3.250} & {\textbf{0.098}} & \multirow{ 2}{*}{\textbf{0.072}} \\[0.5cm]
	\hline
	Global minimum & -3.322 & - & \multirow{ 2}{*}{-} \\[0.5cm]
\hline

\end{tabular}
\label{compare_heart}
\end{table}
The results obtained by BO \& NLM and APNS are far from the global optimum and show important variation. The stationary GP gives better and more robust results, since it is adapted to the stationary behavior of the Hartmann function. However, the minimum obtained by DEGO is closer to the global optimum and the optimization is more robust to the initial DoE than standard GP even if the function is stationary. This shows the interest of using the DEGO even for functions without any information on their stationary behavior unlike BO \& NLM and APNS that are well-suited for stationary functions. Moreover, the convergence curve of the DEGO (Figure~\ref{hart_conv}) highlights its speed of convergence, since after only 30 added points the results given are better and more robust than the other algorithms at the end of the BO process. 

\subsection{Application to industrial test case: design of aerospace vehicle}
To confirm the interest of the BO \& DGP approach, DEGO, a real world aerospace vehicle design optimization problem is considered. It consists of the maximization of the change in velocity ($\Delta V$) of a solid-propellant booster engine. It is a representative physical problem for solid booster design with simulation models fast enough to provide the real minimum to compare and illustrate the efficiency of the proposed algorithm. 
\subsubsection{Description of the problem}

\begin{figure}[!h]
\center
\includegraphics[width=0.85\linewidth]{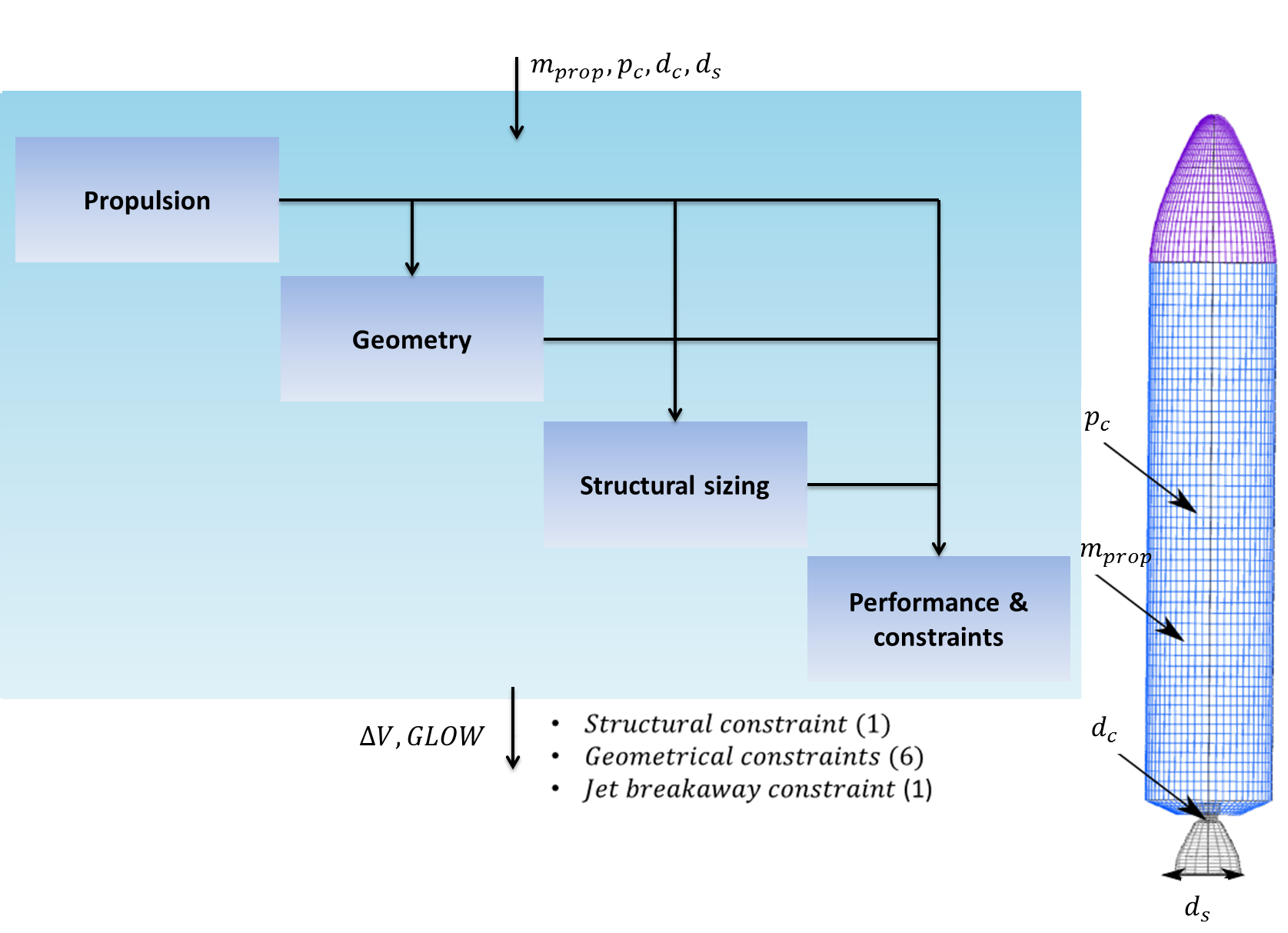}\\%
\caption { Optimization problem of a solid-propellant booster engine.  }%
\label{MDO}
\end{figure}

The optimization of $\Delta V$ for a solid propellant booster is considered (Figure~\ref{MDO}). Four design variables are considered:
\begin{itemize}
\item Propellant mass: $5 \text{ t}<m_{prop}<15 \text{ t}$
\item Combustion chamber pressure: $5 \text{ bar}<p_{c}<100 \text{ bar}$
\item Throat nozzle diameter: $0.2 \text{ m}<d_{c}<1 \text{ m}$
\item Nozzle exit diameter: $0.5 \text{ m}<d_{s}<1.2 \text{ m}$
\end{itemize}
Different constraints are also considered including a structural one limiting the combustion pressure according to the motor case, $6$ geometrical constraints on the internal vehicle layout for the propellant and the nozzle, a jet breakaway constraint concerning the nozzle throat diameter and the nozzle exit diameter, and a constraint on the maximum Gross Lift-Off Weight (GLOW) allowed.
\begin{center}
$
\begin{array}{ll}

\text{Minimize:} &- \Delta V(\textbf{x})\\ 
\text{w.r.t:} &\textbf{X}=[{m}_{prop},{p}_{c},_{c},_{s}]\\
\text{s.t:} & \left\{\begin{array}{l} \text{1 structural constraint}\\
\text{6 geometrical constraints}\\
\text{1 jet breakaway constraints}\\
\text{maximum GLOW allowed}
\end{array} \right.
\end{array}
$
\end{center}

\begin{figure}
\begin{minipage}[c]{0.5\linewidth}
\center
\includegraphics[width=1\linewidth]{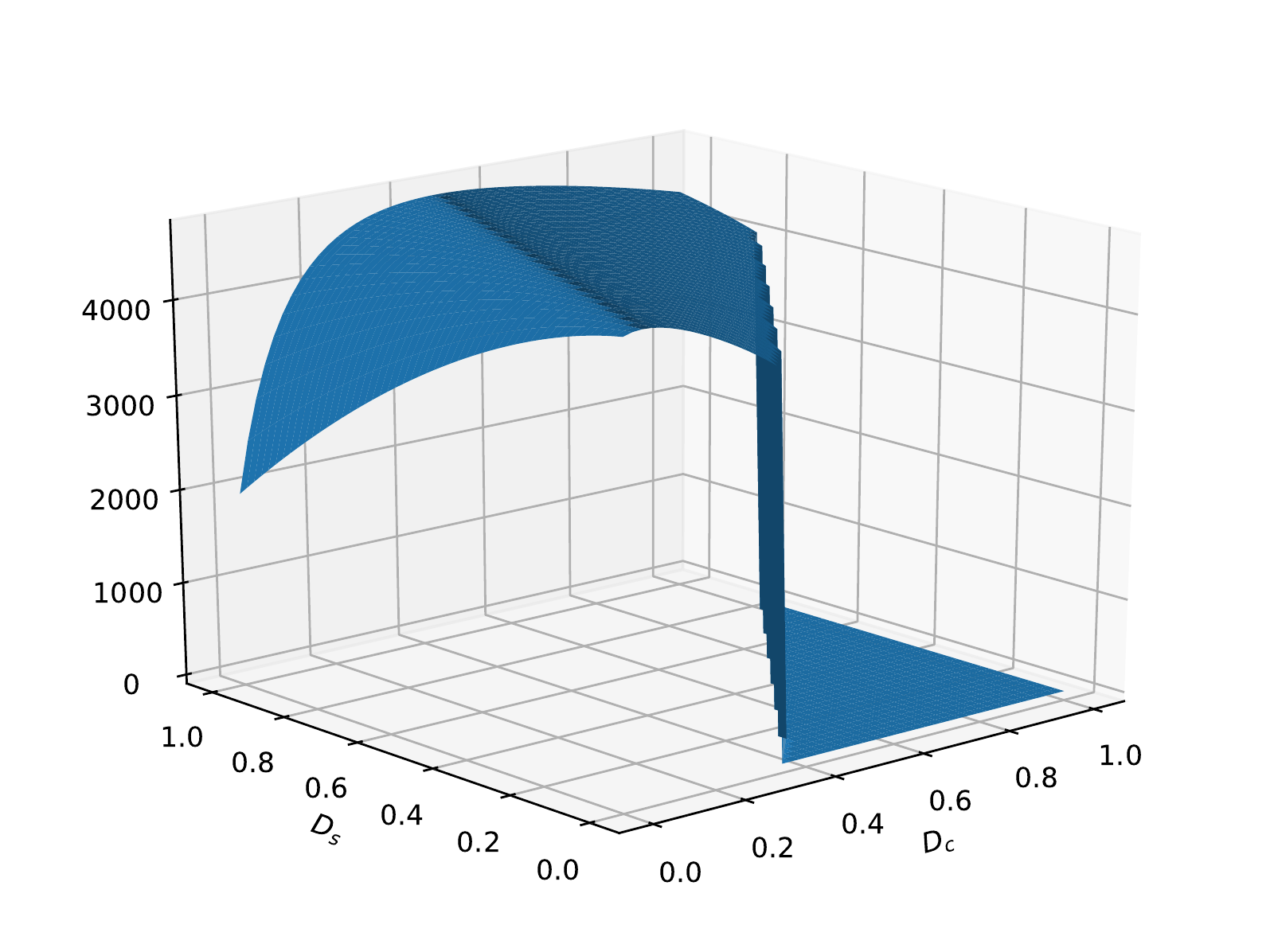}\\
\scriptsize A sectional view of the change in velocity according to the diameters of the nozzle. 
\end{minipage}
\hfill
\begin{minipage}[c]{0.5\linewidth}
\center
\includegraphics[width=1\linewidth]{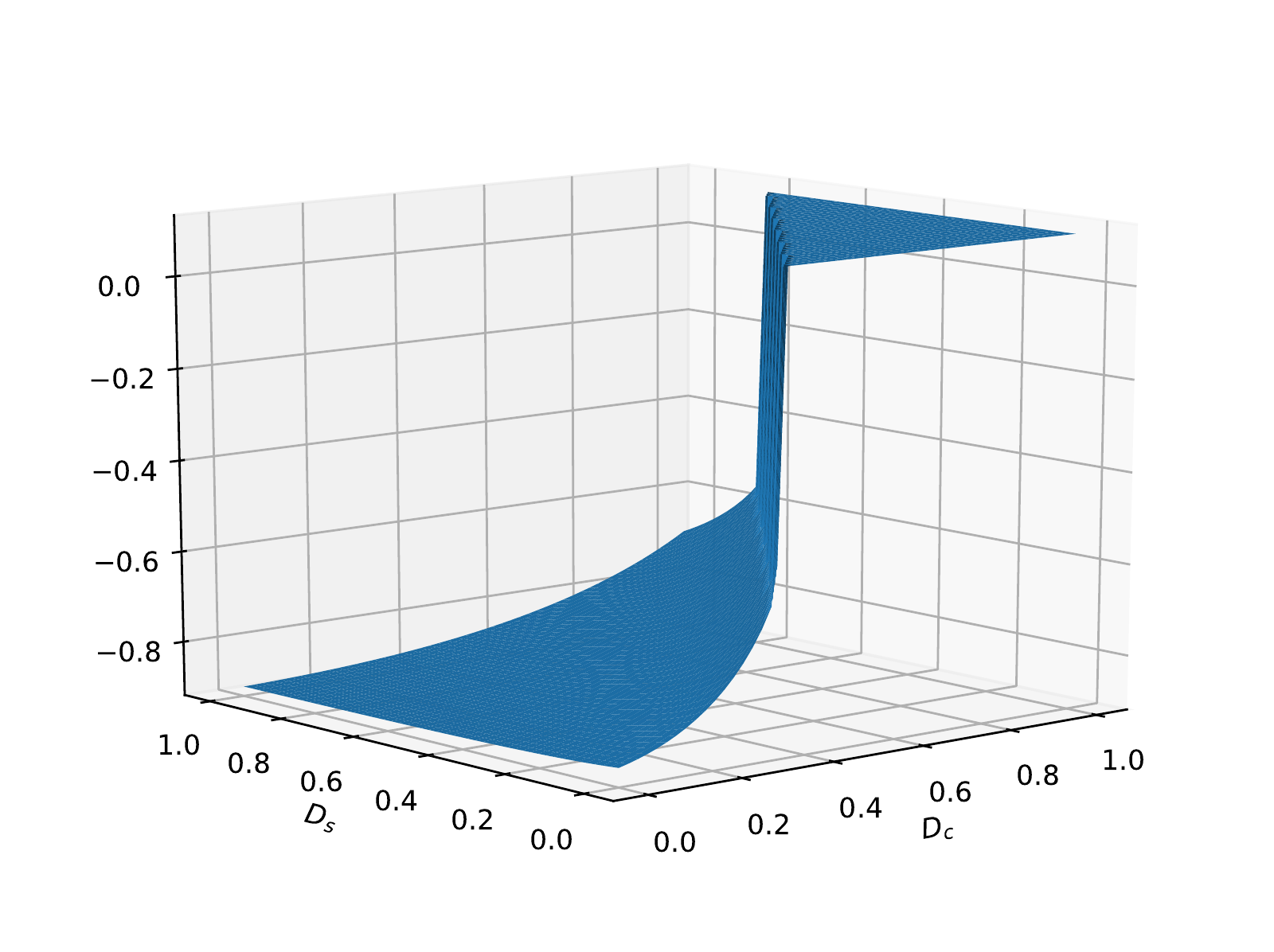}\\
\scriptsize A sectional view of a constraint according to the diameters of the nozzle.
\end{minipage}
\center
\begin{minipage}[c]{0.5\linewidth}
\center
\includegraphics[width=1\linewidth]{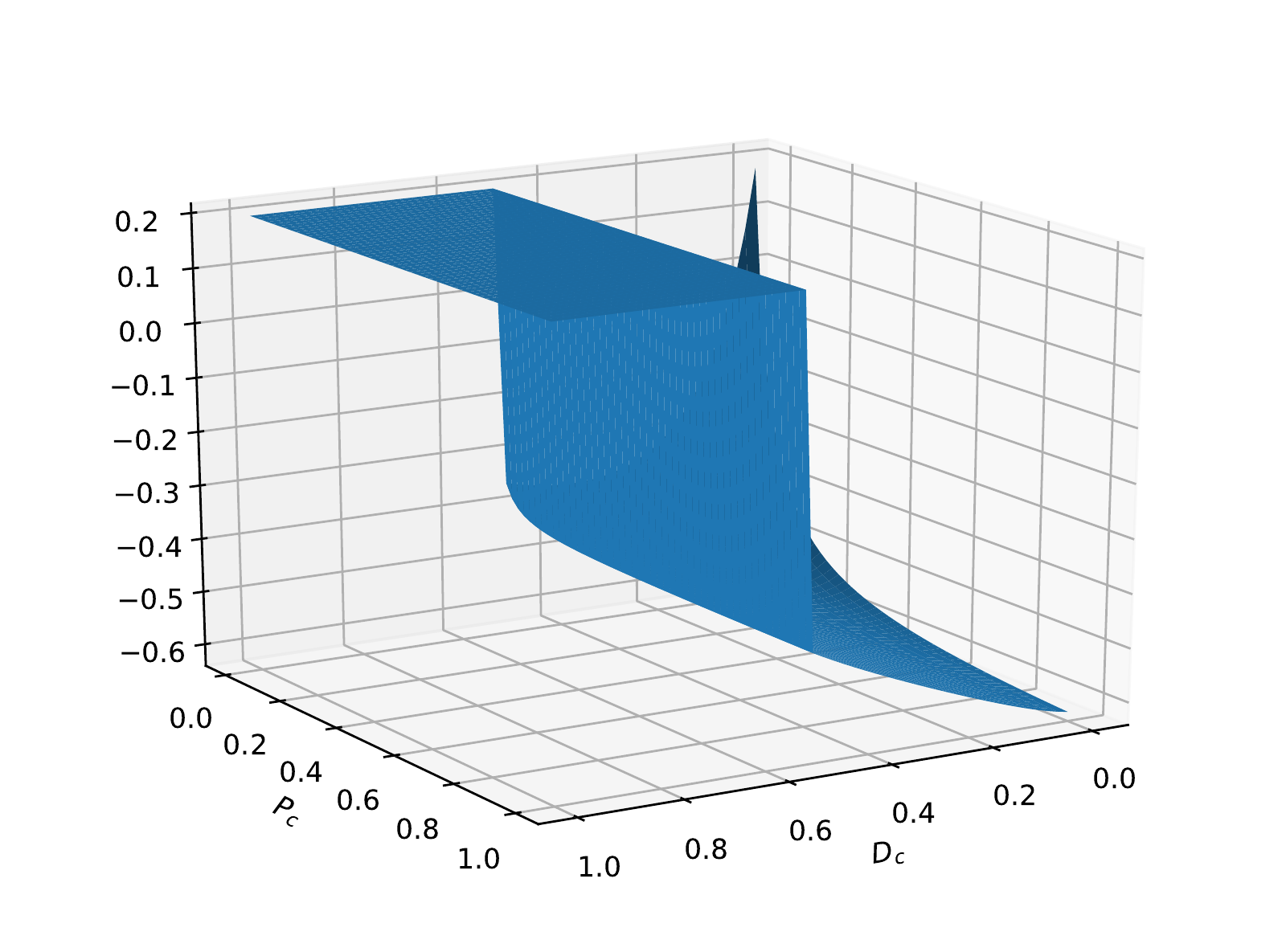}\\
\scriptsize A sectional view of a constraint according to the throat nozzle diameter and the combustion chamber pressure.
\end{minipage}
\caption{Sectional view of the non-stationary behaviors of some functions involved in the booster problem}
\label{plot_ns}
\end{figure}

This problem is expected to have non-stationarity behaviors due to some constraints. In fact, the constraints may have a different behavior in the feasible and unfeasible regions. Moreover the objective function which is the change in velocity may have a tray region when it is equal to zero, due to an insufficient propellant mass (Figure~\ref{plot_ns}). 

\subsubsection{Experimental results}
The initial DoE are set using a Latin Hypercube Sampling of 30 points and 50 points are added with BO using EI for the objective function and EV for the constraints. To assess the robustness of the results, $10$ repetitions are performed.

\begin{figure}
\includegraphics[width=0.9\linewidth]{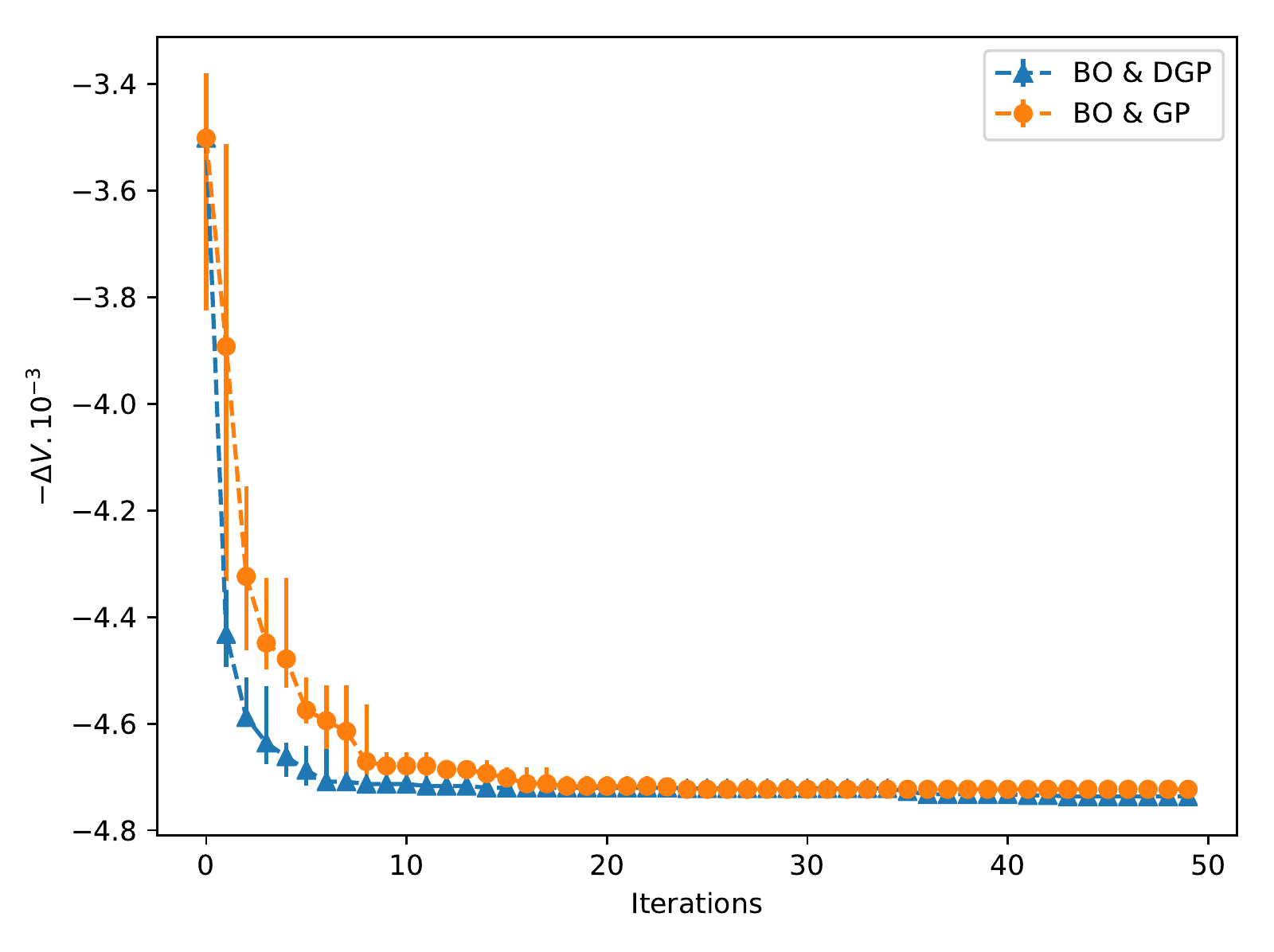}\\
\caption{Convergence curve of $-\Delta V$ using DEGO and BO \& GP.}
\label{convergence}
\end{figure}

The plots of convergence of the BO algorithms are displayed in Figure~\ref{convergence}. After adding 50 points, both BO \& DGP (DEGO) and standard GP reach the global minimum. However, DEGO is faster to converge. DEGO shows robust results near the global optimum $4738 m/s$ after only 12 iterations, while the BO \& GP is not stabilized until 24 iterations (Table~\ref{table_booster}).  

\begin{table}[!h]
\centering	
\caption{Performance of the algorithms after 12 added points, after 24 added points and after 50 added points.}

\begin{tabular}{|c|c|c|c|c|c|c|}
  \hline  & \multicolumn{2}{c|}{After 12 added points} & \multicolumn{2}{c|}{After 24 added points} & \multicolumn{2}{c|}{After 50 added points}\\
  \hline
  \textbf{Algorithm} & \textbf{ Mean }&\textbf{Std} & \textbf{Mean} & \textbf{ Std}&\textbf{Mean} & \textbf{Std} \\
   \hline
   BO \& GP & 4656 & 97.12 & 4709  & 41.33 & 4725 & 10.63 \\
  \hline
	DEGO & 4709 & 27.95 & 4718 & 22.53 & 4736 & 7.49    \\
	\hline
	
\end{tabular}
\label{table_booster}
\end{table}
The speed of convergence is important in case of expensive black-box functions. Indeed, one evaluation of the objective function or the constraints can cost multiple hours, even multiple days. Hence, DEGO is interesting even for problems where BO \& GP can reach the global minimum, due to its speed of convergence which can reduce drastically the number of evaluations needed to converge. 
\clearpage
\section{Conclusion and future works}
A coupling between Bayesian Optimization (BO) and Deep Gaussian Processes (DGP) has been proposed in this paper. This coupling induces some adaptations of the handling of DGPs (training approach, uncertainty on the prediction, architecture of the DGP) and also on BO (the iterative structure of BO, infill criteria). The main propositions are the use of natural gradient on all the variational parameters of the DGP in the training which enables a better convergence of the Evidence Lower Bound, and a better uncertainty quantification on the prediction. Also, to take advantage of the iterative structure of BO, the optimal parameter values of the previous model are used as initialization for the next one to speed up the training of the model. In the considered problems, a DGP with 2 hidden layers proved to give a well-balanced compromise between the time complexity in the training and its power of representation. To use the classic infill criteria considering that the prediction of the DGP model is not necessarily Gaussian, a sampling procedure to approximate infill criterion such as the Expected Improvement was proposed. The algorithm DEGO obtained following these propositions was assessed on analytical test optimization problems. The experimentation showed its better efficiency and robustness compared with standard BO \& GP and approaches using non-linear mapping to handle non-stationarity. Finally, this algorithm was applied to a real-world aerospace engineering design problem, showing its improved speed of convergence compared to standard BO \& GP.

The goal of this paper was to propose a BO \& DGP algorithm and to illustrate its tangible interest over the state-of-the-art approaches. However, it is necessary to explore more this coupling.  For example, the handling of the step size of the Natural Gradient used in each layer when training a DGP model can be improved. Also, infill criteria such as Thompson Sampling or criteria using information theory may be more adapted to DGP than the EI. More parallelism can also be integrated at different levels of DEGO.

\section*{Acknowledgments}
This work is co-funded by ONERA-The French Aerospace Lab and Universit\'e de Lille, in the context of a joint PhD thesis.\\
Discussions with Hugh Salimbeni and Zhenwen Dai were very helpful for this work, special thanks to them.\\
The Experiments presented in this paper were carried out using the Grid'5000 testbed, supported by a scientific interest group hosted by Inria and including CNRS, RENATER and several Universities as well as other organizations (see https://www.grid5000.fr).

\newpage
\begin{appendices}
\section{Functions}
\label{appendixA}
Modified Xiong function:
\begin{equation}
f(x)=-0.5\left(\sin\left(40(x-0.85)^4\right)\cos\left(2.5(x-0.95)\right)+0.5(x-0.9)+1\right), \text{  } x \in[0,1]
\label{xiong}
\end{equation}
Modified TNK constraint function:
\begin{equation}
f(\textbf{x})=1.6(x_0-0.6)^2+1.6(x_1-0.6)^2-0.2\cos\left(20\arctan\left(\frac{0.3x_0}{(x_1+10^{-8})}\right)\right)-0.4, \text{   } \textbf{x}\in[0,1]\times[0,1]
\label{TNK}
\end{equation}

\begin{figure}[h]
\begin{minipage}[c]{0.5\linewidth}
\includegraphics[width=1\linewidth]{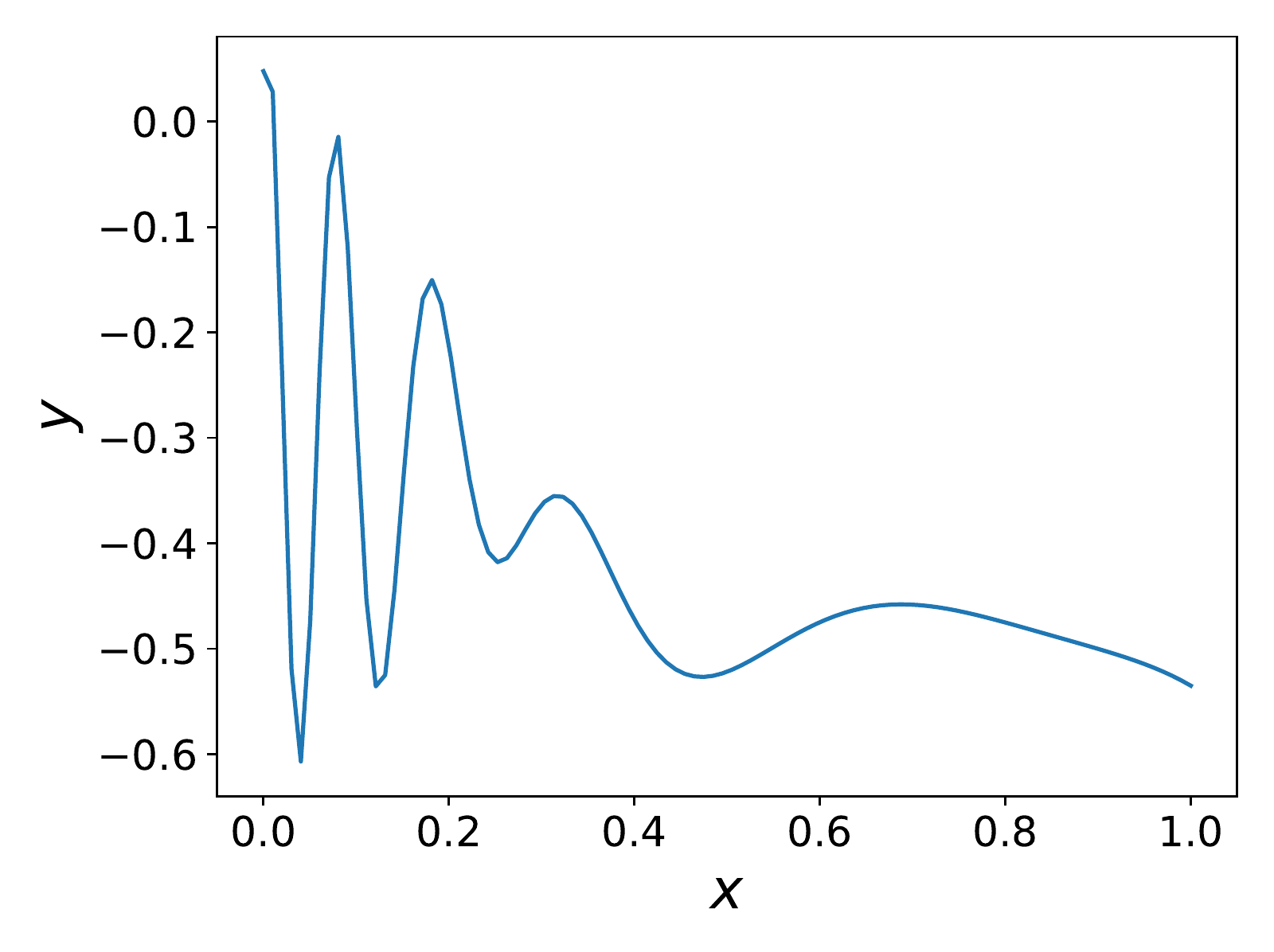}
\caption{Modified Xiong function}
\label{xiong_plot}
\end{minipage}
\hfill
\begin{minipage}[c]{0.5\linewidth}
\includegraphics[width=1\linewidth]{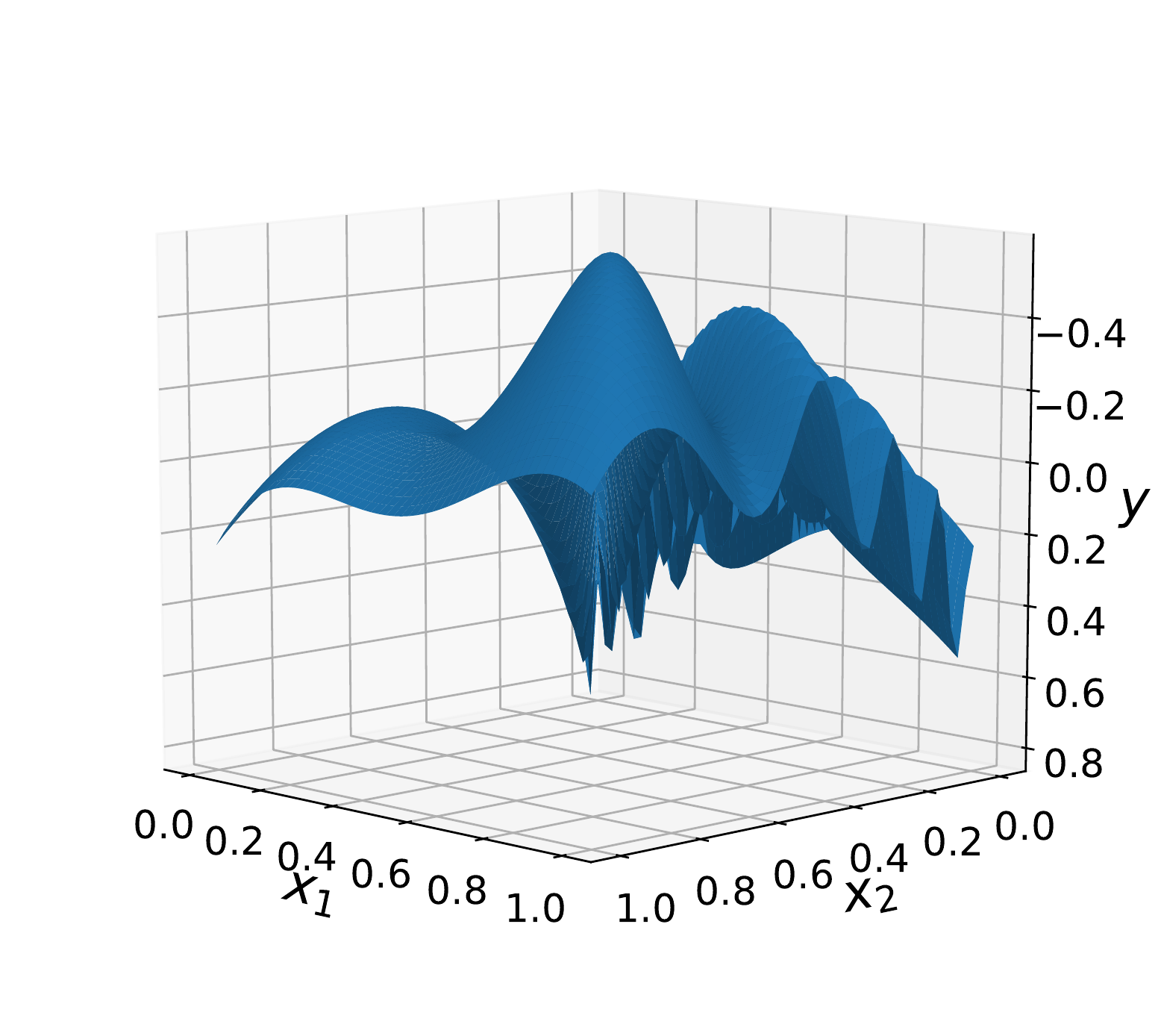}
\caption{Modified TNK constraint}
\label{xiong_plot}
\end{minipage}
\end{figure}
\noindent 10d Trid function:
\begin{equation}
f(\textbf{x})=\sum_{i=1}^{10} (x_i-1)^2-\sum_{i=2}^{10}x_ix_{i-1}, \text{  } x_i \in [-100,100], \forall i=1,\dots,10
\label{trid_eq}
\end{equation} 
Hartmann-6d function:
\begin{equation} f(\textbf{x})= \sum_{i=1}^4 \alpha_i \exp\left(-\sum_{j=1}^6 A_{ij} (x_j-Pij)^2 \right), \text{  } x_i\in[0,1], \forall i=1,\dots,6
\label{heart_eq}
\end{equation}
with: 
$$\alpha=[1,1.2,3,3.2]^\top$$
and
$$P=10^{-4}\begin{bmatrix} 1312 & 1696 & 5569 & 124 & 8283 & 5886\\2329 & 4135 & 8307&3736&1004&9991\\2348&1451&3522&2883&3047&6650\\4047&8828&8732&5743&1091&381\end{bmatrix}$$ 
and $$A= \begin{bmatrix} 10&3&17&3.5&1.7&8\\0.05&10&17&0.1&8&14\\3&3.5&1.7&10&17&8\\17&8&0.05&10&0.1&14 \end{bmatrix}$$
\begin{figure}[h]
\begin{minipage}[c]{0.5\linewidth}
\includegraphics[width=1\linewidth]{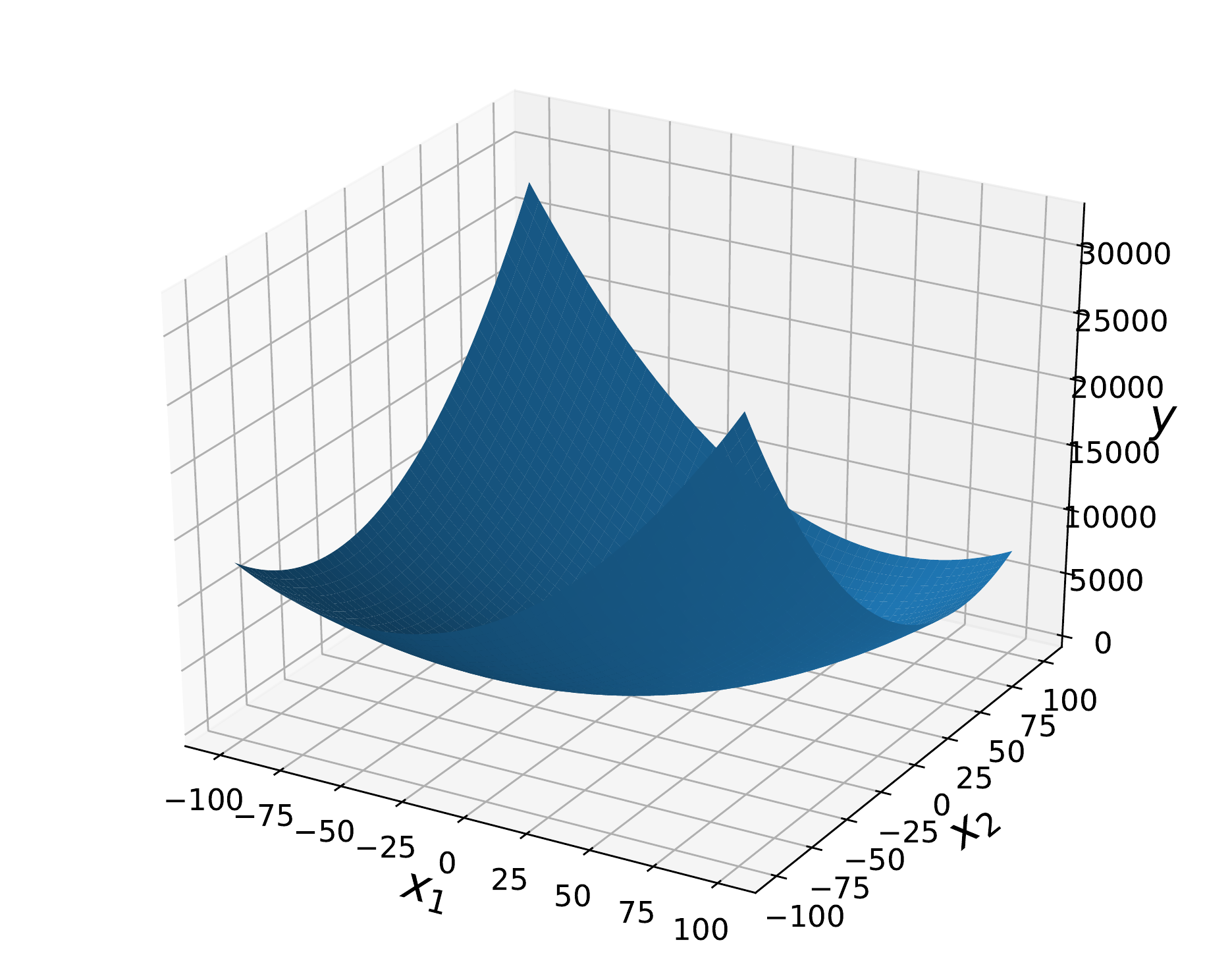}
\caption{Sectional $2d$ view of the Trid function showing where the global minimum lie}
\label{Trid}
\end{minipage}
\hfill
\begin{minipage}[c]{0.5\linewidth}
\includegraphics[width=1\linewidth]{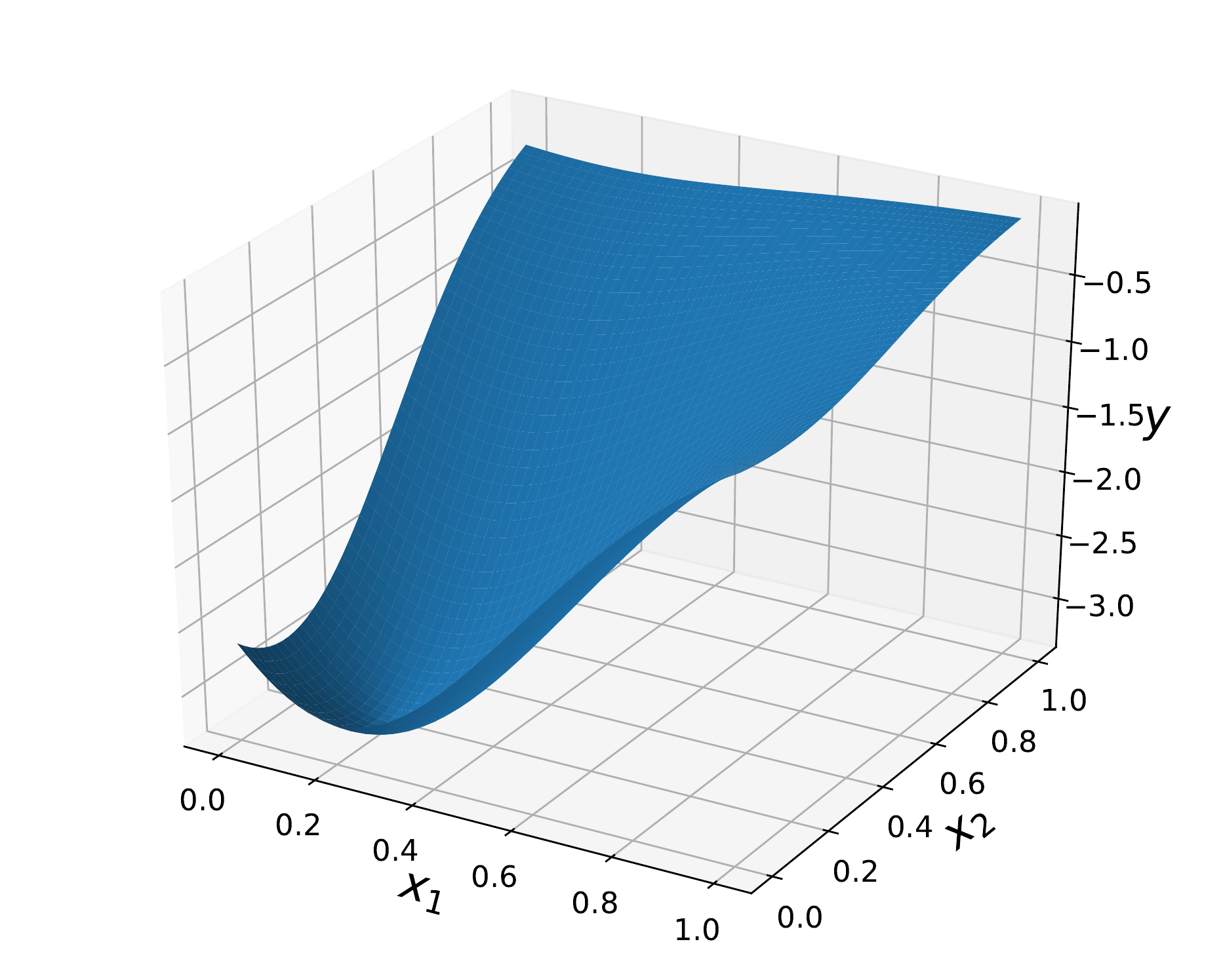}
\caption{Sectional $2d$ view of the Hartmann-6d function showing where the global minimum lie}
\label{slice_hartmann}
\end{minipage}
\end{figure}
\section{Experimental setup}
\label{appendixB}
\begin{itemize}
\item All experiments were executed on Grid'5000 using a Tesla P100 GPU. The code is based on GPflow \cite{GPflow2017} and Doubly-Stochastic-DGP \cite{salimbeni2017doubly}.
\item For all DGPs, RBF kernels are used with a length-scale and variance initialized to 1 if it does not get an initialization from a previous DGP. The data is scaled to have a zero mean and a variance equal to 1.
\item The Adam optimizer is set with $\beta_1=0.8$ and $\beta_2=0.9$ and a step size $\gamma^{adam}=0.01$.
\item The natural gradient step size is initialized for all layers at $\gamma^{nat}=0.1$
\item For DEGO the number of successive updates before optimizing from scratch is 5.
\item The infill criteria are optimized using a parallel differential evolution algorithm with a population of 400 and 100 generations.
\item A Github repository featuring DEGO algorithm will be available after the publication of the paper.  
\end{itemize}

\end{appendices}

\bibliographystyle{unsrt}
\bibliography{Biblio_DEGO}

\end{document}